\documentclass[preprint,12pt,3p]{elsarticle}
\usepackage[utf8]{inputenc}
\usepackage{multirow}
\usepackage{array}
\usepackage[lofdepth,lotdepth]{subfig}
\usepackage[ruled]{algorithm2e}
\usepackage{times,amsmath,amssymb}
\usepackage[T1]{fontenc}
\usepackage[polutonikogreek,english]{babel}
\usepackage{float}
\usepackage{amssymb}
 \usepackage{appendix}

\usepackage{amssymb}

\journal{pattern recognition}

\begin{document}

\begin{frontmatter}

\title{Multimodal Subspace Support Vector Data Description}

\address[label1]{Faculty of Information Technology and Communication Sciences, Tampere University, FI-33720 Tampere, Finland}
\address[label3]{Programme for Environmental Information, Finnish Environment Institute, FI-40500 Jyväskylä, Finland}
\address[label2]{Department of Engineering, Electrical and Computer Engineering, Aarhus University, DK-8200 Aarhus, Denmark\fnref{label4}}

\author[label1]{Fahad Sohrab\corref{cor1}}
\cortext[cor1]{corresponding author}
\ead{fahad.sohrab@tuni.fi}
\author[label1,label3]{Jenni Raitoharju}
\ead{jenni.raitoharju@tuni.fi}
\author[label2]{Alexandros Iosifidis}
\ead{alexandros.iosifidis@eng.au.dk}
\author[label1]{Moncef Gabbouj}
\ead{moncef.gabbouj@tuni.fi}

\begin{abstract}
In this paper, we propose a novel method for projecting data from multiple modalities to a new subspace optimized for one-class classification. The proposed method iteratively transforms the data from the original feature space of each modality to a new common feature space along with finding a joint compact description of data coming from all the modalities. For data in each modality, we define a separate transformation to map the data from the corresponding feature space to the new optimized subspace by exploiting the available information from the class of interest only. We also propose different regularization strategies for the proposed method and provide both linear and non-linear formulations. The proposed Multimodal Subspace Support Vector Data Description outperforms all the competing methods using data from a single modality or fusing data from all modalities in four out of five datasets. 
\end{abstract}


\end{frontmatter}



\section{Introduction}
In our surroundings, on a daily basis, we are exposed to information from many different sources. Different sensors are used to gather information about similar objects. Our brains usually perform well in combining the information from different sources to make a concise analysis of that particular entity. In order to analyze an entity, even a single source of information might be enough, but to make some critical decisions it is important to combine information from different sources in a systematic way. For example, if a person is walking in a crowd, the main information to not hit anything comes from visual cues, but people can warn each other also by voice or even by touch, and this extra information helps in understanding the environment in a better way. The smell could help to avoid unpleasant spots, too. As another example, while watching a movie, only visual information of the scenes may not be enough to understand the whole scenario, but the audio and/or captions combined together with the visuals information will provide the full information. 

In machine learning techniques for predictive data modeling, training data are used to form a model that can accurately classify future instances into a predefined number of classes. In many cases, data comes from sensors and can be further processed to extract different features. The term \textit{multimodal} is used to describe the data coming from different sensors (also referred to as mode or modality), however, it is also used as a synonym to \textit{multi-view} when different features are extracted from the same sensor or when there are multiple similar sensors, e.g., cameras. The aim of multimodal machine learning algorithms is to build models that can process and relate information from more than one modality (or view).

The examples of multimodal representations are prevalent in different application areas. In \cite{qu2017active}, an active multimodal sensor system for target recognition and tracking is studied where information from three different sensors (visual, infrared, and hyperspectral) is used. In \cite{zhang2018vehicle}, a framework for vehicle tracking with multimodal data (velocity and images) is proposed where the outcome of velocity modality estimated by using a Kalman filter on the data obtained from motion sensors is fused with features learned from image modality by the color-faster R-CNN method. In \cite{kye2017multimodal}, a multimodal data collection framework for mental stress monitoring is studied. In the proposed framework, physiological and motion sensor data of people under stress are collected.

The data in multimodal applications come from different modalities, where each modality has its own statistical properties and contains specific information. The different modalities usually share high-level concepts and semantic information, and all together contain more information than any single-modal data \cite{gu2017learning}. If we build a model separately for each modality, the relationship between the modalities cannot be exploited efficiently. In multimodal subspace learning, the goal is to infer a shared latent representation, that can accurately model data from each original modality and exploit the relationship between the modalities.

In traditional multiclass machine learning, an adequate amount of data are available for all the categories during training and, hence, the algorithm takes advantage of all available training data from all classes to train a model \cite{iosifidis2016mcsvm}. However, it is possible that during the training, data are highly imbalanced, or the only data available is from a single class. In such cases, one-class classification techniques are used. It is useful in many different cases, such as outlier detection, predicting specific events, or, in general, predicting a specific target class. While much effort has been put on solving one-class classification tasks for data of a single modality \cite{khan2014one}, much less effort has been put on solving one-class multimodal challenges in general, and we are not aware of any prior work in the field of multimodal learning for one-class classification. In one-class multimodal tasks, it is assumed that the only data available is from a single class in many different modalities.

In this paper, we propose a novel method for solving multimodal one-class classification tasks. The proposed method, Multimodal Subspace Support Vector Data Description (MS-SVDD), finds a transformation for each modality along with defining a common model for all modalities in a lower-dimensional subspace optimized for one-class classification. The rest of the paper is organized as follows. In Section \ref{relatedwork}, an overview of related work is presented. In Section \ref{svddexplain}, the newly proposed MS-SVDD is derived and discussed. In Section \ref{experiments}, we present the experimental setup and results, and finally, in Section \ref{conclusion}, conclusions are drawn.

\section{Background and related work}\label{relatedwork}
In this section, we briefly discuss the principles of multimodal learning, along with subspace learning. We also provide an overview of traditional methods used for multiclass multimodal data description and one-class unimodal data description. 

\subsection{Multimodal learning}
The availability of many different modalities can be bliss if it increases the performance of the machine learning model. However, if the data description algorithm fails to make a strong connection between the different available modalities, the performance can be degraded. To ensure better performance of the model by combining data from different modalities, mainly two principles should be ensured, i.e., consensus and complementary principles \cite{xu2013survey}:
\begin{itemize}
    \item\textbf{Consensus principle} aims at minimizing the disagreement between data available from different modes. Maximizing the agreement will reduce the error rate, and better modeling of data is achieved while combining data from different modalities.   
    \item\textbf{Complementary principle} in the context of multimodal learning means that data from each modality may contain some knowledge not contained by the other ones. So it is necessary to exploit information from all the available modes to make an accurate description of data.
\end{itemize}

The multimodal machine learning techniques can be described by three main properties: two-view vs. multi-view, linear vs. non-linear, and unsupervised vs. supervised  \cite{cao2018generalized}. As the name indicates, in two-view learning, the number of views is limited to two. In multi-view learning, the number of views is not limited. The difference between supervised and unsupervised learning is that, in supervised learning, the information on output labels of the training data is taken into account when training the model, while in unsupervised methods, the labels are not used to model the underlying structure or distribution of the data \cite{khante2017learning}. Linear techniques for multimodal subspace learning may be too simple to provide a representative model. Hence, kernel methods are proposed to capture non-linear patterns in data. 

The multimodal learning techniques have been mainly applied on four applications domains \cite{baltruvsaitis2018multimodal}: i.e., audio-visual speech recognition \cite{heckmann2018audio}, multimedia content indexing and retrieval \cite{cao2017multi}, understanding human multimodal behaviors \cite{chen2017detecting}, and language and vision media description \cite{venugopalan2017captioning}. Recently, there has been a rising trend in applying multimodal machine learning algorithms also to other applications. For example, in \cite{maimaitijiang2020soybean}, a multimodal data fusion technique is used for the prediction of soybean yield from an unmanned aerial vehicle.

In multimodal learning, the main goal is to develop a process of fusing information from various modalities. In \cite{baltruvsaitis2019multimodal}, the fusion strategies are divided into two different categories as model-agnostic and model-based approaches. In model-agnostic approaches, the fusion is either late, early, or hybrid. In early fusion, the data or extracted features are fused together at the very initial phase of modeling. A new feature vector is usually formed by concatenating all the available data from different modes, and the model is trained with the new feature vector. In late fusion, multiple models are trained, and the fusion is done for scores generated by each model for the corresponding modality. The score generated by each model can be a threshold or some probability used in decision making. Hybrid fusion exploits the advantage of both early fusion and late fusion. Model-based approaches for fusion explicitly fuses data during their construction, such as kernel-based approaches, graphical models, and neural networks. In this work, we present a model-based approach for data fusion.
\subsection{Subspace learning}
In the current era of data science, where high-dimensional multimodal big data are generated every minute in different industries, there is a need to get the essential insights and mine knowledge in this high-dimensional data. Subspace learning aims at representing data in a lower-dimensional space by keeping intact all the information available in the original higher-dimensional space. 

Algorithms developed for linear subspace learning find a projection matrix for labeled training data (represented by vectors) satisfying some optimality criteria. Principal Component Analysis (PCA) is one of the first subspace learning methods mentioned in literature. In PCA, a subspace is learned by orthogonally projecting data to a subspace so that the variance of data is maximized. PCA works only with a single mode of data, i.e., all data should be in the same dimension. Another traditional subspace learning method is Linear Discriminant Analysis (LDA), which finds a linear transformation by exploiting the class information.

Analogous to PCA, but used for two-view learning, is canonical-correlation analysis (CCA) \cite{hotelling1936relations}. CCA is a classic and conventional method for subspace learning, which aims at relating two sets of data by finding out the pairs of directions which provide a maximum correlation between the two sets. It has recently become one of the popular methods for unsupervised subspace learning because of its generalization capability and has been used extensively for multimodal data fusion and cross-media retrieval \cite{benesty2018canonical}. In subspace learning, state-of-the-art results are achieved by methods which have embraced some stimulus from conventional subspace learning methods \cite{xu2018cross}. 

As an extension of methods for linear transformation, kernel methods are introduced to describe nonlinear function or decision boundaries. In kernel methods, the data are mapped to a typically higher-dimensional kernel-space using a kernel function where it exhibits linear patterns \cite{scholkopf2001learning,sadooghi2018improving}. For example, in \cite{scholkopf1997kernel}, kernel-PCA performing a nonlinear form of PCA is proposed.

\subsection{One-class classification}
In one-class classification, the parameters of the model are estimated using data from the positive class only because data from the other classes are either not available at all or it is too diverse in nature to be modeled statistically \cite{kefi2019novel}. The positive class is also called the target class, and the data from the other classes, which are not available during training, is called negative, or an outlier class. For example, a unimodal biometric system uses a single biometric trait for verification or identification \cite{raghavendra2018improved}.

Support Vector Data Description (SVDD) \cite{tax2004support} is among the most widely used one-class classification methods used for anomaly detection and other related applications. SVDD obtains a spherical boundary around target data which can be made flexible by using the kernel trick. The obtained boundary is used to detect outliers during the test, i.e., anything inside the closed boundary is classified as a target class and otherwise as an outlier. The Lagrangian of SVDD is given as follows
\begin{eqnarray}\label{origSVDD}
L = \sum_{i=1}^{N} \alpha _{i} \mathbf{x}_{i}^T  \mathbf{x}_{i}- \sum_{i=1}^{N}\sum_{j=1}^{N} \alpha _{i} \mathbf{x}_{i}^T \mathbf{x}_{j} \alpha _{j},
\end{eqnarray}
where $\mathbf{x}_i$ is the input target training instance and maximizing \eqref{origSVDD} gives a set of $\alpha_i$ corresponding to each instance. The instances with $\alpha_i \ge 0$ define the data description. Other common one-class classification method is One-Class Support Vector Machine (OC-SVM) \cite{scholkopfu1999sv}. 

Techniques for enhancing the performance of one-class classification methods, mainly extensions of SVDD, can be categorized into four main categories: methods based on data structure, kernel issue, boundary shape, and non-stationary data \cite{sadeghi2018automatic}. As the name indicates, in the data structure category, the main focus is on the structure of data. For example, in \cite{el2014support}, a confidence coefficient is associated with each training sample to deal with the uncertainty of data. In kernel issue extensions, the main focus is on reducing the complexity or proposing new kernels for one-class classification. For example, in \cite{jeong2015support}, a new kernel is proposed to improve the accuracy of SVDD for time series classification. Proposing changes in the boundary for enclosing the target data comes under the third category for improving one-class classification accuracy. For example, in \cite{forghani2011support}, the ellipse shape is used for encapsulating target data instead of the traditional sphere used in SVDD.
In \cite{mygdalis2016graph}, it is shown that both SVDD and OC-SVM lead to the same solution when exploiting the elliptical shape of the class. The last category of algorithms for improving one-class classifier performance attempts to handle non-stationary data. For example in \cite{tax2003online}, Incremental-SVDD (I-SVDD) is proposed to handle non-stationary or increasing data. Recently, in \cite{hamidzadeh2018belief}, an algorithm developed for reducing the effect of uncertain data around the hypersphere of SVDD achieved the state of the art result on many UCI \cite{Dua:2017} datasets. In this paper, we consider baseline SVDD combined with multimodal subspace learning. However, in the future, the method can be further extended using similar ideas.

In the area of multimodal one-class classification, researchers have mainly focused on fusing the output labels of multiple models trained for each type of feature independently, i.e., without taking into account information from other feature types for one model \cite{tran2013user}. 

\section{Multimodal subspace support vector data description}\label{svddexplain}
 MS-SVDD maps data from high-dimensional feature spaces to a low-dimensional feature space optimized for one-class classification. The optimized subspace is shared by data coming from all modalities. MS-SVDD is an extension of Subspace Support Vector Data Description (S-SVDD), which was proposed for unimodal data in \cite{sohrab2018subspace}. The main novelty of MS-SVDD is using the multimodal approach for one-class classification. Here, we first derive the linear MS-SVDD. Then we derive two non-linear versions using the kernel trick \cite{scholkopf2001learning} and the Nonlinear Projection Trick (NPT) \cite{kwak2013nonlinear}, respectively.
 \begin{figure*}[t]
	\centering
	\includegraphics[width=\textwidth]{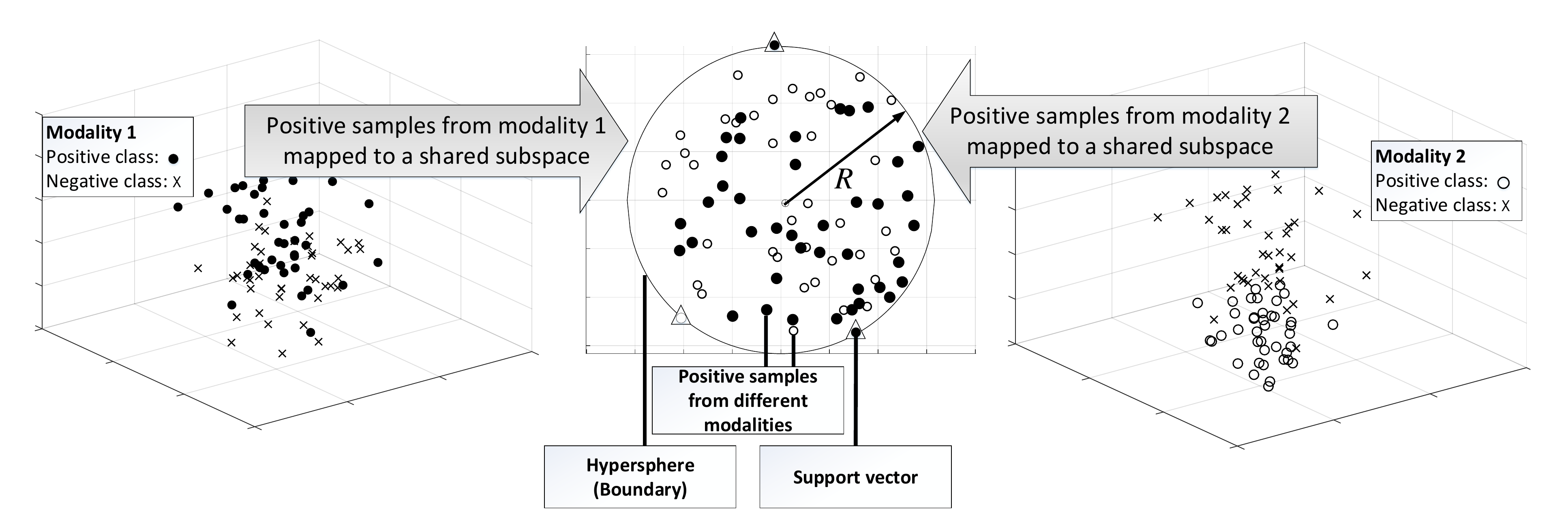}
	\caption{Depiction of proposed MS-SVDD: Data from two modalities in their corresponding feature space are mapped to a common subspace, where positive class instances are enclosed inside a (hyper)sphere.}
	\label{mssvdddemo}
\end{figure*}

\subsection{Linear MS-SVDD}
Let us assume that the items to be modelled are represented by M different modalities. The instances in each modality $m$, $m=1,\dots,M$, are represented by $\mathbf{X}_m=[\mathbf{x}_{m,1},\mathbf{x}_{m,2},\dots\mathbf{x}_{m,N}]$, $\mathbf{x}_{m,i} \in \mathbb{R}^{D_m}$, where $N$ is the total number of instances and $D_m$ is the dimensionality of the feature space in modality $m$. MS-SVDD tries to find a projection matrix $\mathbf{Q}_m$ $\in$  $\mathbb{R}^{d \times {D}_m}$ for each modality, which will project the corresponding instances to a lower ($d$)-dimensional optimized subspace shared by all modalities. Thus, a feature vector $\mathbf{x}_{m,i}$ is projected to a $d$-dimensional vector $\mathbf{y}_{m,i}$ as
\begin{equation}\label{eq:Y_i}
\mathbf{y}_{m,i} = \mathbf{Q}_m \mathbf{x}_{m,i},\forall m \in \{1,\dots,M\} \:\:, \forall i \in \{1,\dots,N\}.
\end{equation}
To obtain a common description of all the data transformed from their corresponding modalities to the new common subspace, we exploit Support Vector Data Description (SVDD) \cite{tax2004support} to form a closed boundary around the target class data in the new subspace. The center and radius of the hypersphere are denoted by  $\mathbf{a} \in \mathbb{R}^d$ and $R$, respectively. Figure \ref{mssvdddemo} depicts the basic idea of the proposed method. 

In order to find a compact hypersphere which encloses all the target data from all the modalities in the new subspace, we minimize 

\begin{equation}\label{erfunc}
F(R,\textbf{a}) = R^2 \nonumber
\end{equation}  
s.t.
\begin{eqnarray}\label{constraints0}
\| {\mathbf{Q}_m\mathbf{x}_{m,i}} - \mathbf{a} \|^2_{2} \le R^2, \forall m \in \{1,\dots,M\}, \forall i \in \{1,\dots,N\}.
\end{eqnarray}

By introducing slack variables $\xi_{m,i}$, such that most of the training data from all the modalities in the new common space should lie inside the hypersphere, the above criterion becomes
\begin{equation}\label{errorfunc}
F(R,\textbf{a}) = R^2 + C\sum_{m=1}^{M}\sum_{i=1}^{N} \xi_{m,i}\nonumber
\end{equation}  
s.t.
\begin{eqnarray}
\|\mathbf{Q}_m\mathbf{x}_{m,i} - \mathbf{a}\|_2^2 \le R^2 + \xi _{m,i}, \nonumber\\
\xi_{m,i} \ge 0,\nonumber\\ \forall m \in \{1,\dots,M\}, \forall i \in \{1,\dots,N\}. \label{constraints2}
\end{eqnarray}  
The Lagrange function corresponding to \eqref{constraints2} can be given as
\begin{eqnarray}
L= {R^2} + C\sum_{m=1}^{M}\sum_{i=1}^{N} \xi_{m,i}
- \sum_{m=1}^{M}\sum_{i=1}^{N} \gamma_{m,i} \xi_{m,i} - \sum_{m=1}^{M}\sum_{i=1}^{N} \alpha_{m,i} \Big( R^2 + \xi _{m,i}- \nonumber\\ \mathbf{x}_{m,i}^T \mathbf{Q}_m^T \mathbf{Q}_m \mathbf{x}_{m,i} + 2\textbf{a}^T \textbf{Q}_m \textbf{x}_{m,i}- \textbf{a}^T\textbf{a} \Big)   \label{lang}
\end{eqnarray}
The Lagrangian function should be maximized with respect to  $\alpha_{m,i}\geq0$, and $\gamma_{m,i} \geq 0$ and minimized with respect to  $R$,  $\textbf{a}$,  ${\xi_{m,i}}$, and $\mathbf{Q}_m$.
By setting the partial derivative to zero, we get
\begin{eqnarray}
\frac{\partial L}{\partial R}=0 &\Rightarrow& \sum_{m=1}^{M}\sum_{i=1}^{N} \alpha_{m,i} = 1 \label{der1} \\
\frac{\partial L}{\partial \mathbf{a}}=0 &\Rightarrow& \mathbf{a} =  \sum_{m=1}^{M}\sum_{i=1}^{N} \alpha_{m,i} \mathbf{Q}_m\mathbf{x}_{m,i} \label{der2} \\
\frac{\partial L}{\partial \xi _{m,i}}=0 &\Rightarrow& C- \alpha _{m,i} - \gamma _{m,i}  = 0 \label{der3} \\
\frac{\partial L}{\partial \mathbf{Q}_m}=0 &\Rightarrow& 
 \mathbf{Q}_m = \Big(\textbf{a}  \sum_{i=1}^{N} \alpha_{m,i} \textbf{x}_{m,i}^T\Big)\Big( \sum_{i=1}^{N} \alpha_{m,i}   \mathbf{x}_{m,i} \mathbf{x}_{m,i}^T \Big)^{-1} \label{der4}
\end{eqnarray} 

It is clear from \eqref{der1}-\eqref{der4} that parameters $\mathbf{\alpha}$ and $\mathbf{Q}$ are interrelated and cannot be jointly optimized. Hence we apply a two step iterative optimization process where, in each step, we fix one parameter and optimize the other. Substituting (\ref{eq:Y_i}), (\ref{der1}), (\ref{der2}) and (\ref{der3}) in the Lagrangian function (\ref{lang}), we get
\begin{eqnarray}\label{Lang2}
L = \sum_{m=1}^{M}\sum_{i=1}^{N} \alpha _{m,i} \mathbf{y}_{m,i}^T  \mathbf{y}_{m,i}-\sum_{m=1}^{M}\sum_{i=1}^{N}\sum_{n=1}^{M}\sum_{j=1}^{N} \alpha _{m,i} \mathbf{y}_{m,i}^T \mathbf{y}_{n,j} \alpha _{n,j}.
\end{eqnarray}
We see that optimizing \eqref{Lang2} for $\alpha$ corresponds to the traditional SVDD applied in the subspace. Maximizing \eqref{Lang2} for a particular set of data will give us $\alpha_{m,i}$ corresponding each sample. The value of $\alpha_{m,i}$ for corresponding sample defines its position with respect to the hypersphere:
\begin{itemize}
  \item Samples with $0 < \alpha_{m,i} < C$ define the data description and lie on the boundary of hypersphere, they are refered to as support vectors. 
  \item Samples with $\alpha_{m,i}=C$ are outside the boundary.
  \item  Samples with $\alpha_{m,i}=0$ lie inside the boundary.
\end{itemize}

In the second step, we fix $\alpha$ and update $\mathbf{Q}_m$ for each modality. For this step, we add a regularization term $\omega$:
\begin{eqnarray}\label{lang3}
L = \sum_{m=1}^{M}\sum_{i=1}^{N} \alpha _{m,i} \mathbf{x}_{m,i}^T \mathbf{Q}_{m}^T  \mathbf{Q}_{m} \mathbf{x}_{m,i} -\sum_{m=1}^{M}\sum_{i=1}^{N}\sum_{n=1}^{M}\sum_{j=1}^{N} \alpha _{m,i}\mathbf{x}_{m,i}^T \mathbf{Q}_{m}^T  \mathbf{Q}_{n} \mathbf{x}_{n,j} \alpha _{n,j} + \beta\omega.
\end{eqnarray}
The regularization term $\omega$ expresses the covariance of data from different modalities in the new low-dimensional space, and $\beta$ is a regularization parameter for controlling the significance of $\omega$. We propose different settings for $\omega$ as 
\begin{eqnarray} 
\omega_0 &=& 0, \label{generalconstraint0} \\
\omega_1 &=& \sum_{m=1}^{M} \text{tr}(\mathbf{Q}_m\mathbf{X}_m  \mathbf{X}^T_m\mathbf{Q}^T_m), \label{generalconstraint1} \\
\omega_2 &=& \sum_{m=1}^{M}  \text{tr}(\mathbf{Q}_m\mathbf{X}_m \boldsymbol{\alpha}_m \boldsymbol{\alpha}^T_m \mathbf{X}^T_m\mathbf{Q}^T_m), \label{generalconstraint2} \\
\omega_3 &=& \sum_{m=1}^{M}  \text{tr}(\mathbf{Q}_m\mathbf{X}_m \boldsymbol{\lambda}_m \boldsymbol{\lambda}^T_m \mathbf{X}^T_m\mathbf{Q}^T_m), \label{generalconstraint3} \\
\omega_4 &=& \sum_{m=1}^{M} \sum_{n=1}^{M} \text{tr}(\mathbf{Q}_m\mathbf{X}_m\mathbf{X}^T_n\mathbf{Q}^T_n), \label{generalconstraint4} \\
\omega_5 &=&\sum_{m=1}^{M} \sum_{n=1}^{M}  \text{tr}(\mathbf{Q}_m\mathbf{X}_m \boldsymbol{\alpha}_m \boldsymbol{\alpha}^T_n \mathbf{X}^T_n\mathbf{Q}^T_n), \label{generalconstraint5} \\
\omega_6 &=&\sum_{m=1}^{M} \sum_{n=1}^{M}  \text{tr}(\mathbf{Q}_m\mathbf{X}_m \boldsymbol{\lambda}_m \boldsymbol{\lambda}^T_n \mathbf{X}^T_n\mathbf{Q}^T_n), \label{generalconstraint6} 
\end{eqnarray} 
where $\boldsymbol{\alpha}_m \in \mathbb{R}^{N}$ in \eqref{generalconstraint2} and \eqref{generalconstraint5} is a vector having the elements $\alpha_{m,1},...,\alpha_{m,N}$. Thus, $\boldsymbol{\alpha}_m$ has non-zero values for support vectors and outliers. $\boldsymbol{\lambda}_m \in \mathbb{R}^{N}$ in \eqref{generalconstraint3} and \eqref{generalconstraint6} is a vector having the elements of $\boldsymbol{\alpha}_m$ that are smaller than $C$. Values of $\boldsymbol{\alpha}_m$ corresponding to the outliers (i.e., $\alpha_{m,i} = C$) are replaced with zeros in $\boldsymbol{\lambda}_m$. Thus, $\boldsymbol{\lambda}_m$ has non-zero values only for the support vectors. For $\omega_0$, the regularization term becomes obsolete and it is not used in the optimization process. In $\omega_1$, the regularization term only uses representations coming from the respective modality and no representations from the other modalities are used to describe the variance of the positive class. In $\omega_2$, all support vectors, i.e., representations at the hypersphere boundary, and outliers are used to describe the class variance for the update of the corresponding $\mathbf{Q}_m$. In $\omega_3$, only support vectors of the respective modality are used to describe the variance of the class to be modelled. In $\omega_4$, data from all the modalities are used to describe the covariance and regularize the update of $\mathbf{Q}_m$. In $\omega_5$, the instances belonging to the hypersphere boundary and outliers from all modalities are used to describe the covariance. In $\omega_6$, only the support vectors belonging to class boundary from all modalities are used to update $\mathbf{Q}_m$ and describe the covariance of the positive class. 

Note that the MS-SVDD formulation reduces to S-SVDD \cite{sohrab2018subspace} if data from only one modality $(M=1)$ are taken into account for data description. In S-SVDD, a single projection matrix $\mathbf{Q}$ is determined for mapping the data $\mathbf{X}$ from higher-dimensional space to a lower-dimensional space. A regularization term $\psi$, which expresses the class variance in the low-dimensional space, is added to the Lagrangian function of S-SVDD:
\begin{equation} \label{psi} 
\psi = \text{tr}(\mathbf{Q}\mathbf{X} \boldsymbol{\lambda}\boldsymbol{\lambda}^T \mathbf{X}^T\mathbf{Q}^T),
\end{equation}
where $\boldsymbol{\lambda}$ can take different forms as described in \cite{sohrab2018subspace}. The regularization terms, $\omega_0$, $\omega_1$, $\omega_2$, and $\omega_3$ for MS-SVDD become equivalent to the regularization terms proposed for S-SVDD when $M=1$. Hence, MS-SVDD is a more generalized form of S-SVDD, which can form a data description by considering data from multiple modalities.

We update $\mathbf{Q}_m$ by using the gradient of $L$ in \eqref{lang3} with respect to $\mathbf{Q}_m$,
\begin{eqnarray}\label{GD1}
\mathbf{Q}_m \leftarrow \mathbf{Q}_m - \eta \Delta L,
\end{eqnarray}
where $\eta$ is the learning rate parameter and the gradient of $L$ is calculated as
\begin{eqnarray}\label{eqforQ1}
\frac{\partial L}{\partial \mathbf{Q}_m} = 2\sum_{i=1}^{N} \alpha_{m,i} \mathbf{Q}_m \mathbf{x}_{m,i} \mathbf{x}_{m,i}^T
- 2\sum_{i=1}^{N}\sum_{j=1}^{N}\sum_{n=1}^{M} \mathbf{Q}_n \mathbf{x}_{n,j} \mathbf{x}_{m,i}^T \alpha_{m,i} \alpha_{n,j} + \beta\Delta \omega,
\end{eqnarray}
where $\Delta \omega$ is the derivative of the regularization term with respect to $\mathbf{Q}_m$
\begin{eqnarray} 
\Delta\omega_0 &=& 0, \label{deltageneralconstraint0} \\
\Delta\omega_1 &=& 2\mathbf{Q}_m\mathbf{X}_m  \mathbf{X}^T_m, \label{deltageneralconstraint1} \\
\Delta\omega_2 &=& 2\mathbf{Q}_m\mathbf{X}_m \boldsymbol{\alpha}_m \boldsymbol{\alpha}^T_m  \mathbf{X}^T_m, \label{deltageneralconstraint2} \\
\Delta\omega_3 &=& 2\mathbf{Q}_m\mathbf{X}_m \boldsymbol{\lambda}_m \boldsymbol{\lambda}^T_m  \mathbf{X}^T_m, \label{deltageneralconstraint3} \\
\Delta\omega_4 &=& 2\sum_{n=1}^{M} (\mathbf{Q}_n\mathbf{X}_n  \mathbf{X}^T_m), \label{deltageneralconstraint4} \\
\Delta\omega_5 &=& 2\sum_{n=1}^{M} (\mathbf{Q}_n\mathbf{X}_n \boldsymbol{\alpha}_n \boldsymbol{\alpha}^T_m  \mathbf{X}^T_m), \label{deltageneralconstraint5} \\
\Delta\omega_6 &=& 2\sum_{n=1}^{M} (\mathbf{Q}_n\mathbf{X}_n \boldsymbol{\lambda}_n \boldsymbol{\lambda}^T_m  \mathbf{X}^T_m). \label{deltageneralconstraint6} 
\end{eqnarray} 

We initialize the $\mathbf{Q}_m$ using PCA. At every iteration, the projection matrix is orthogonalized and normalized so that
\begin{eqnarray}\label{eqforQm1}
 \mathbf{Q}_m \mathbf{Q}_m^T= \mathbf{I},
\end{eqnarray}
where $\mathbf{I}$ is an identity matrix. We use QR decomposition for orthogonalizing and normalizing the projection matrix $\mathbf{Q}_m$. Algorithm \ref{algo} describes the overall MS-SVDD algorithm.

\begin{algorithm}[H]
  \caption{MS-SVDD optimization}\label{algo}
\SetAlgoLined
\SetKwInOut{Input}{Inputs}
\SetKwInOut{Output}{Outputs}
\Input{$\mathbf{Z}_m$ for each $m=1,...,M$, // Input data from all modalities \\$\beta$, // Regularization parameter for controlling significance of $\omega$ \\$\eta$, // Learning rate parameter \\$d$, // Dimensionality of joint subspace \\$C$, // Regularization  parameter in SVDD \\$M$ // Total number of modalities} 
 \Output{$\textbf{S}_m$ for each $m=1,...,M$, // Projection matrices for different modalities\\$R$, // Radius of hypersphere \\$\mbox{\boldmath$\alpha$}$ // Defines the data description }  
  \vspace{3mm}
 $\textbf{Z}_m=\textbf{X}_m$ for linear and NPT case ($\textbf{K}_m$ for kernel case)\\
 $\textbf{S}_m=\textbf{Q}_m$ for linear and NPT case ($\textbf{W}_m$ for kernel case)\\
\vspace{3mm} 
\For{m=1:M}{
   \vspace{2mm}
   Initialize $\mathbf{S}_m$ via linear-PCA (kernel-PCA)\;
 }
   \vspace{2mm}
   \For{$iter=1:max\_iter$}{
       \vspace{2mm}
    For each $m$, map $\mathbf{Z}_m$ to $\mathbf{Y}_m$ using Eq. (\ref{eq:Y_i}) (Eq. \eqref{kerneleq2})\;
          Form Y by combining all $\mathbf{Y}_m$'s\;
         \vspace{1mm}
        Solve SVDD in the subspace to obtain \boldmath$\alpha$ in Eq. \eqref{Lang2}\;
   \vspace{2mm}
  \For{m=1:M}{
   \vspace{2mm}
     Calculate $\Delta L$ using Eq. \eqref{eqforQ1} (Eq. \eqref{eqforW1}) \;
    Update $\mathbf{S}_m \leftarrow \mathbf{S}_m - \eta \Delta L$\;
        \vspace{2mm}
   Orthogonalize and normalize $\mathbf{S}_m$ using QR decomposition (eigendecomposition);\\
  }
}
   For each $m$, compute $\mathbf{Y}_m$ using Eq. (\ref{eq:Y_i}) (Eq. \eqref{kerneleq2})\;
   Form $\mathbf{Y}$ by combining all $\mathbf{Y}_m$'s\;
   Solve SVDD to obtain the final data description\;
\end{algorithm}

\subsection{Non-linear MS-SVDD}\label{kernelandopt}
For non-linear mapping from the original feature spaces to a new shared feature space, we use two approaches. The first approach is based on the standard kernel trick \cite{scholkopf2001learning} and the second on the Nonlinear Projection Trick (NPT) \cite{kwak2013nonlinear}, which is used as a computationally lighter alternative to the kernel trick.

\subsubsection{Non-linear MS-SVDD with standard kernel trick}\label{standardkerneltrick}
In the non-linear data description, the original data are mapped to a kernel space $\mathcal{F}$ using a non-linear function $\phi(\cdot)$ such that $\mathbf{x}_{m,i} \in \mathbb{R}^{D_m} \rightarrow \phi(\mathbf{x}_{m,i}) \in \mathcal{F}$. The kernel space dimensionality can possibly be infinite. Then the data are projected from the kernel space to $\mathbb{R}^{d}$ as
\begin{equation}\label{eq:Y_i_kernel}
\mathbf{y}_{m,i} = \mathbf{{Q}}_m \phi(\mathbf{x}_{m,i}), \:\:\forall i \in \{1,\dots,N\}.
\end{equation}
 In order to calculate $\mathbf{y}_{m,i}$, we use the so-called kernel trick by expressing the projection matrix $\mathbf{Q}_m$ as a linear combination of the training data representations of the respective modality in the kernel space $\mathcal{F}$, leading to
\begin{eqnarray}\label{kerneleq2}
\mathbf{y}_{m,i} = \mathbf{W}_m \mathbf{\Phi}_m^T \phi(\mathbf{x}_{m,i}) = \mathbf{W}_m \mathbf{k}_{m,i}, \:\forall i \in \{1,\dots,N\},
\end{eqnarray}
where $\mathbf{\Phi}_m \in \mathbb{R}^{|\mathcal{F}| \times N}$ is a matrix formed in $\mathcal{F}$ containing the training data representations of modality $m$, $\mathbf{W}_m \in \mathbb{R}^{d \times N}$ is a matrix containing the weights for $\mathbf{\Phi}_m$ needed to form $\mathbf{Q}_m$, and $\mathbf{k}_{m,i}$ is the $i$-th column of the Gramian matrix, also called as the kernel matrix, $\mathbf{K}_m \in \mathbb{R}^{N \times N}$, having elements equal to $\mathbf{K}_{m,ij} = \phi(\mathbf{x}_{m,i})^T \phi(\mathbf{x}_{m,j})$. In our experiments, we use the Radial Basis Function (RBF) kernel, given by
\begin{equation}\label{RBFkernel}
\mathbf{K}_{m,ij} = \exp  \left( \frac{ -\| \mathbf{x}_{m,i} - \mathbf{x}_{m,j}\|_2^2 }{ 2\sigma^2 } \right),
\end{equation}  
where  $\sigma>0$  is a hyperparameter and determines the width of the kernel. 

 The augmented version of the Lagrangian function now takes the following form:
\begin{eqnarray}\label{kernelLang}
L = \sum_{m=1}^{M}\sum_{i=1}^{N} \alpha _{m,i} \mathbf{k}_{m,i}^T \mathbf{W}_{m}^T  \mathbf{W}_{m} \mathbf{k}_{m,i} -
\sum_{m=1}^{M}\sum_{i=1}^{N}\sum_{n=1}^{M}\sum_{j=1}^{N} \alpha _{m,i}\mathbf{k}_{m,i}^T \mathbf{W}_{m}^T  \mathbf{W}_{n} \mathbf{k}_{n,j} \alpha _{n,j} + \beta\omega.
\end{eqnarray}
The $\alpha$'s are calculated optimizing \eqref{Lang2} with $\mathbf{W}_m$'s fixed, i.e., applying SVDD in the subspace. In the second step, the $\alpha$'s are fixed and $\mathbf{W}_m$'s are updated with the gradient descent:
\begin{eqnarray}\label{GD2}
\mathbf{W}_m \leftarrow \mathbf{W}_m - \eta \Delta L,
\end{eqnarray}
where the gradient is calculated as
\begin{eqnarray}\label{eqforW1}
\frac{\partial L}{\partial \mathbf{W}_m} = 2\sum_{i=1}^{N} \alpha_{m,i} \mathbf{W}_m \mathbf{k}_{m,i} \mathbf{k}_{m,i}^T 
- 2\sum_{i=1}^{N}\sum_{j=1}^{N}\sum_{n=1}^{M} \mathbf{W}_n \mathbf{k}_{n,j} \mathbf{k}_{m,i}^T \alpha_{m,i} \alpha_{n,j} 
+ \beta\Delta \omega.
\end{eqnarray}
The gradient of the regularization term, $\Delta\omega$, now takes the following forms:
 \begin{eqnarray} 
\Delta\omega_0 &=& 0, \label{wkdeltageneralconstraint0} \\
\Delta\omega_1 &=& 2\mathbf{W}_m\mathbf{K}_m  \mathbf{K}^T_m, \label{wkdeltageneralconstraint1} \\
\Delta\omega_2 &=& 2\mathbf{W}_m\mathbf{K}_m \boldsymbol{\alpha}_m \boldsymbol{\alpha}^T_m  \mathbf{K}^T_m, \label{wkdeltageneralconstraint2} \\
\Delta\omega_3 &=& 2\mathbf{W}_m\mathbf{K}_m \boldsymbol{\lambda}_m \boldsymbol{\lambda}^T_m  \mathbf{K}^T_m, \label{wkdeltageneralconstraint3} \\
\Delta\omega_4 &=& 2\sum_{n=1}^{M} (\mathbf{W}_n\mathbf{K}_n  \mathbf{K}^T_m), \label{wkdeltageneralconstraint4} \\
\Delta\omega_5 &=& 2\sum_{n=1}^{M} (\mathbf{W}_n\mathbf{K}_n \boldsymbol{\alpha}_n \boldsymbol{\alpha}^T_m  \mathbf{K}^T_m), \label{wkdeltageneralconstraint5} \\
\Delta\omega_6 &=& 2\sum_{n=1}^{M} (\mathbf{W}_n\mathbf{K}_n \boldsymbol{\lambda}_n \boldsymbol{\lambda}^T_m  \mathbf{K}^T_m). \label{wkdeltageneralconstraint6} 
\end{eqnarray}

We initialize the matrix $\textbf{W}_m$ for each mode using kernel-PCA. We orthogonalize and normalize $\mathbf{W}_m$ at every iteration so that
\begin{eqnarray}\label{eqforWm}
 \mathbf{W}_m \mathbf{\Phi}_m^T \mathbf{\Phi}_m \mathbf{W}_m^T= \mathbf{I}.
\end{eqnarray}
We decompose \eqref{eqforWm} using eigendecomposition as \begin{eqnarray}\label{eqforWm2}
 \mathbf{W}_m \mathbf{\Phi}_m^T \mathbf{\Phi}_m \mathbf{W}_m^T =\mathbf{V}_m\mathbf{\Lambda}_m \mathbf{V}_m^{T},
\end{eqnarray}
where $\mathbf{\Phi}_m^T \mathbf{\Phi}_m$ is $\mathbf{K}_m$, $\mathbf{\Lambda}_m$ is a diagonal matrix containing the eigenvalues of  $\mathbf{W}_m \mathbf{\Phi}_m^T \mathbf{\Phi}_m \mathbf{W}_m^T$ and $\mathbf{V}_m$ contains the corresponding eigenvectors. After further simplification, the normalized projection matrix $\mathbf{\Hat{W}}_m$ can be computed as
\begin{eqnarray}\label{normW}
 \mathbf{\Hat{W}}_m = (\mathbf{\Lambda}_m^{\frac{1}{2}})^+ \mathbf{V}_m^T\mathbf{W}_m,
\end{eqnarray}
where the $+$ sign denotes pseudo-inverse. For notation simplicity, we set $\mathbf{{W}}_m=\mathbf{\Hat{W}}_m$.
\subsubsection{Non-linear MS-SVDD with Nonlinear Projection Trick}
The non-linear MS-SVDD using the kernel trick requires computing the eigendecomposition \eqref{eqforWm2} at every iteration. This is computationally expensive and, therefore, we propose an alternative non-linear approach using NPT \cite{kwak2013nonlinear}. Here, a non-linear mapping is applied only at the beginning of the process, while the optimization follows the linear MS-SVDD. In the NPT-based MS-SVDD, we first compute kernel matrix $\mathbf{K}_{m}$ using \eqref{RBFkernel}. In the next step, the computed kernel matrix is centralized as 
\begin{eqnarray}\label{centerK}
\mathbf{\Hat{K}}_{m} = (\mathbf{I}- \mathbf{E}_N) \mathbf{K}_{m} ( \mathbf{I}-\mathbf{E}_N)
\end{eqnarray}
where $\mathbf{\Hat{K}}_{m}$ is the centralized kernel matrix and $\mathbf{E}_N$ is $N\times N$ matrix defined as 
\begin{eqnarray}\label{ennpt}
\mathbf{E}_N = \frac{1}{N}\mathbf{1}_N \mathbf{1}_{N}^T. 
\end{eqnarray}
$\mathbf{1}_N \in \mathbb{R}^{N}$is a vector with each element having value of $1$. The centralized matrix $\mathbf{\Hat{K}}_{m}$ is decomposed by using eigendecomposition,
\begin{eqnarray}\label{eigen}
\mathbf{\Hat{K}}_{m} = \mathbf{U}_m\mathbf{A}_m\mathbf{U}_m^T, 
\end{eqnarray}
where $\mathbf{A}_m$ contains the non-negative eigenvalues of the centered kernel matrix and $\mathbf{U}_m$ contains the corresponding eigenvectors. The data in the reduced dimensional kernel space is obtained as
\begin{eqnarray}\label{nptdata}
\mathbf{\Phi}_{m} = (\mathbf{A}^{\frac{1}{2}}_m)^{+} \mathbf{U}_m^{+} {\mathbf{\Hat{K}}_{m} }
\end{eqnarray}
Since we consider NPT as a pure preprocessing step, we continue by considering $\mathbf{\Phi}_{m}$ as our input data, i.e., we set $\mathbf{X}_{m}=\mathbf{\Phi}_{m}$. Then we follow the linear MS-SVDD. Note that in cases where the number of training samples is high, this pre-processing step can be highly accelerated by following approximations, like the Nystr\"{o}m-based Approximate Kernel Subspace Learning method in \cite{iosifidis2016aksl}.

\subsection{Test Phase}\label{testing}
During the test phase, an instance $\mathbf{x}_{m*} \in \mathbb{R}^{D_m}$ (the $_*$ in subscript denotes test instance) coming from modality $m$ is projected to the common $d$-dimensional subspace using \eqref{eq:Y_i} for the linear case. For kernel case, first, the kernel vector is computed as
\begin{eqnarray}\label{kvector}
\mathbf{k}_{m*} = \mathbf{\Phi}_m^T \phi(\mathbf{x}_{m*})
\end{eqnarray}
and then projected to the common $d$-dimensional subspace using \eqref{kerneleq2}. For NPT, first the kernel vector $\mathbf{k}_{m*}$ is computed and then centralized as
\begin{eqnarray}\label{centerKtest}
\mathbf{{\Hat{k}}}_{m*}= (\mathbf{I}- \mathbf{E}_N) [  \mathbf{{{k}}}_{m*}-\frac{1}{N}\mathbf{K}_{m} \mathbf{1}_N].
\end{eqnarray}
The centralized kernel vector is mapped to
\begin{eqnarray}\label{npttest}
{\mathbf{\phi}}_{m*} = \mathbf{(\Phi}_{m}^T)^{+}\mathbf{\Hat{k}}_{m*}
\end{eqnarray}
and then to $d$-dimensional subspace using \eqref{eq:Y_i} (for notation simplicity ${\mathbf{\phi}}_{m*}$ is considered as ${\mathbf{x}}_{m*}$). 

The decision to classify the test instance $\mathbf{y}_{m*}$ as positive or negative is taken on the basis of its distance from the center of hypersphere, i.e.,
\begin{eqnarray}\label{testequation}
\|\mathbf{y}_{m*} - \mathbf{a}\|_2^2= \mathbf{y}_{m*}^T\mathbf{y}_{m*} - 2 \sum_{k=1}^{M}\sum_{i=1}^{N} \alpha_{k,i} \mathbf{y}_{m*}^T\mathbf{y}_{k,i} 
+  \sum_{k=1}^{M}\sum_{i=1}^{N}\sum_{n=1}^{M} \sum_{j=1}^{N}  \alpha_{k,i} \alpha_{n,j} \mathbf{y}_{k,i}^T\mathbf{y}_{n,j}. 
\end{eqnarray}
The representation $\mathbf{y}_{m*}$ is assigned to the positive class when $\|\mathbf{y}_{m*} - \mathbf{a}\|_2^2 \le R^2$ and to the negative class if $\|\mathbf{y}_{m*} - \mathbf{a}\|_2^2 > R^2$, where $R^2$ is the distance from center $\mathbf{a}$ to any support vector on the boundary,
\begin{eqnarray}\label{rsquare}
R^2&=& \mathbf{v}^T\mathbf{v}- 2 \sum_{m=1}^{M}\sum_{i=1}^{N} \alpha_{m,i} \mathbf{y}_{m,i}^T\mathbf{v} 
+  \sum_{m=1}^{M}\sum_{i=1}^{N}\sum_{n=1}^{M} \sum_{j=1}^{N}  \alpha_{m,i} \alpha_{n,j} \mathbf{y}_{m,i}^T\mathbf{y}_{n,j},
\end{eqnarray}
where $\mathbf{v}$ is any support vector in the training set with corresponding $\alpha$ having value $0 < \alpha < C$. Since the items are represented by $M$ different modalities, the final decision for assigning the item to a particular class (either positive or negative) can be taken using different strategies explained in Section \ref{decissions}.
\subsection{Complexity Analysis}\label{complexityanalysis}
The linear version of the proposed method has the following main steps: 1) Initializing the projection matrices via PCA, 2) mapping data from all modalities to a lower $d$-dimensional shared space, 3) SVDD for obtaining the $\alpha$ values and final data description for all data points coming from $M$ different modalities, 4) computing the gradient ($\Delta L$) for each modality, 5) updating the projection matrices and 6) QR decomposition for orthogonalizing and normalizing the projection matrices. We analyze each of these steps and then compute the overall complexity of the algorithm: 
\begin{enumerate}
\item PCA of a matrix is computed by the eigenvalue decomposition of its covariance matrix, so it involves two steps, i.e., computing the covariance matrix and then the eigenvalue decomposition of the obtained covariance matrix. The complexity of calculating covariance matrix and corresponding eigenvalue decomposition for a single modality is  $\mathcal{O}\big(ND_m\times min(N,D_m)\big)$ and  $\mathcal{O}\big(D_m^3\big)$, respectively \cite{elgamal2015analysis}. The complexity of computing PCA for all modalities is $\mathcal{O}\big(min(N^2D_1,D_1^2N)+D_1^3)+(min(N^2D_2,D_2^2N)+D_2^3)+\dots+(min(N^2D_M,$ $D_M^2N)+D_M^3\big)$. We denote the sum of dimensions of all modalities as $\Sigma_\mathcal{D} = D_1+D_2+\dots+D_M$ and similarly the sum of squared dimensions as $\Sigma_{\mathcal{D}^2} = D_1^2+D_2^2+\dots+D_M^2$ (note that $\Sigma_{\mathcal{D}^2} \neq (\Sigma_\mathcal{D})^2$) and sum of cubed dimensions as $\Sigma_{\mathcal{D}^3} = D_1^3+D_2^3+\dots+D_M^3$. Hence, the complexity of initializing the projection matrices via PCA becomes $\mathcal{O}\big(min(N^2\Sigma_{\mathcal{D}},\Sigma_{\mathcal{D}^2}N)+\Sigma_{\mathcal{D}^3}\big)$. 
\item The complexity of mapping data from the original $D_m$ dimensional space to a lower $d$-dimensional space is the complexity of multiplying $d\times D_m$ and $D_m \times N$, which has the complexity of $\mathcal{O}\big(dD_mN\big)$. Repeating this for all modalities we get  $\mathcal{O}\big(d\Sigma{_\mathcal{D}}N\big)$
\item The complexity of SVDD for $N$ data points is $\mathcal{O}\big({N}^3\big)$ \cite{zheng2016smoothly}. For all data points coming from $M$ different modalities it becomes $\mathcal{O}\big(M^3{N}^3\big)$. 
\item The gradient $\Delta$L to update $\mathbf{Q}_m$ is computed using \eqref{eqforQ1}, where the second term has the highest complexity (equally high as regularization terms 4-6). Its complexity is $\mathcal{O}(2dN^2D_m\Sigma_{\mathcal{D}})$. As this step is repeated for all modalities the total complexity becomes $\mathcal{O}(2dN^2{\Sigma_{\mathcal{D}}}^2)$.
\item Updating the projection matrices has $\mathcal{O}\big(d\Sigma_{\mathcal{D}}\big)$ complexity. 
\item The complexity of QR decomposition for a single modality is $\mathcal{O}(d{D_m}^2)$ \cite{sharma2013qr}. Thus, the overall complexity of QR decompositions for all the modalities is $\mathcal{O}(d\Sigma_{\mathcal{D}^2})$. \end{enumerate}
Dropping the relatively lower intensive computational steps and adding the rest, the full complexity of the proposed method reduces to
$\mathcal{O}\big(min(N^2\Sigma_{\mathcal{D}},\Sigma_{\mathcal{D}^2}N)+\Sigma_{\mathcal{D}^3} + M^3N^3\big)$. Assuming that the total number of samples $M*N$ is always greater than $\mathcal{D}$ and $M << N$, the time complexity of (a single iteration of) our proposed algorithm in terms of the big $\mathcal{O}$ notation is $\mathcal{O}(N^3)$. In the testing phase, each representation of a test sample in each modality is projected to the $d$-dimensional subspace and then its distance is compared to $R$. This has the total complexity of $\mathcal{O}(d\Sigma_{\mathcal{D}} + Md)$.

For the non-linear version with NPT, the kernel matrix $\mathbf{K}_m$ is first formed which has the complexity of $\mathcal{O}(D_m N^2)$. Then the kernel matrix is centralized and decomposed by using eigendecomposition. Both of these steps have the complexity of $\mathcal{O}(N^3)$. As the data dimensionality in the remaining steps of the proposed method changes from $D_m$ to $N$, the total complexity of the remaining steps becomes $\mathcal{O}\big(MN^3 + M^3N^3\big)$. Thus, the overall complexity in terms of the big $\mathcal{O}$ notation remains at $\mathcal{O}\big(N^3\big)$ for $M << N$, while in practice the computational complexity is higher (by a scalar multiplier $c$) than for the linear version. Also for the non-linear version with the standard kernel trick, the overall complexity remains the same, but the kernel mapping is repeated at every iteration and, thus, the scalar $c$ becomes larger for the overall training process. The testing complexity of the non-linear methods increases to $\mathcal{O}(N\Sigma_{\mathcal{D}} + dMN + Md)$.

\section{Experiments}\label{experiments}
\subsection{Datasets and prepossessing}\label{datasetandsetup}
To evaluate the proposed method, we performed different sets of experiments over 5 datasets. Robot Execution Failures dataset, Single Proton Emission Computed Tomography (SPECTF) heart dataset, and Ionosphere dataset were downloaded from UC Irvine (UCI) machine learning repository \cite{Dua:2017}. Caltech-7 dataset and Handwritten dataset were downloaded from a repository for multi-view learning \cite{li2015large}. The details of datasets and experiments are as follows.

The first set of experiments was performed on the Robot Execution Failures dataset \cite{lopes1998feature}. In Robot Execution Failures dataset, force and torque measurements are collected at regular intervals of time after a task failure is detected. The dataset is divided into five different learning problems (LP) corresponding to different triggering events:

\begin{itemize}
    \item\textbf{LP1:} Failures in approach to grasp position
    \item\textbf{LP2:} Failures in the transfer of a part
    \item\textbf{LP3:} Position of the part after a transfer failure
    \item\textbf{LP4:} Failures in approach to ungrasp position
    \item\textbf{LP5:} Failures in motion with part
\end{itemize}
The total number of instances and the distribution of the classes are given in Table \ref{robotdataset}. All instances are given as 15 samples collected at 315 ms regular time intervals for each sensor. For this dataset, we consider all the instances belonging to the normal class as the target class and the remaining classes as the non-target data. Hence, we have two modalities (torque and force measurements), and we consider the dataset as a one-class classification problem.

\begin{table}[t]\footnotesize\setlength{\tabcolsep}{2pt}
  \centering
   \caption{Robot Execution Failures dataset}
   
\begin{tabular}{lll}
\hline
Learning problem & Instances & Classes and Distribution                                                                                                                                    \\ \hline
LP1              & 88        & \begin{tabular}[c]{@{}l@{}}24\% normal\\ 19\% collision    \\ 18\% front collision\\ 39\% obstruction\end{tabular}                                          \\ \hline
LP2              & 47        & \begin{tabular}[c]{@{}l@{}}43\% normal\\ 13\% front collision\\ 15\% back collision\\ 11\% collision to the right\\ 19\% collision to the left\end{tabular} \\ \hline
LP3              & 47        & \begin{tabular}[c]{@{}l@{}}43\% ok\\ 19\% slightly moved\\ 32\% moved\\ 06\% lost\end{tabular}                                                               \\ \hline
LP4              & 117       & \begin{tabular}[c]{@{}l@{}}21\% normal\\ 62\% collision\\ 18\% obstruction\end{tabular}                                                                     \\ \hline
LP5              & 164       & \begin{tabular}[c]{@{}l@{}}27\% normal\\ 16\% bottom collision\\ 13\% bottom obstruction\\ 29\% collision in part\\ 16\% collision in tool\end{tabular}     \\ \hline
\end{tabular}\label{robotdataset}
\end{table}

The second set of experiments was performed SPECTF heart dataset \cite{kurgan2001knowledge}. The SPECTF heart dataset consists of two sets of features corresponding to rest and stress condition SPECTF images of different subjects. The training set consists of 40 examples diagnosed as healthy heart muscle perfusions and 40 diagnosed as pathological perfusions. The test set consists of 15 instances of healthy heart muscle perfusions and 172 from instances diagnosed as pathological perfusions. We convert this to a multimodal one-class classification problem by considering the rest and stress conditions as different modalities and by selecting the healthy heart muscle perfusions as our target class.

The third set of experiments was performed over the Caltech-7 dataset. We used Gabor feature and Wavelet moments as our two different modalities. The dataset contains 1474 total samples from 7 different classes. We selected faces (435 samples) as our target class and the rest of the classes all together (1039 samples) as the outlier class.

We used Ionosphere dataset for the fourth set of experiments. The categories in this dataset are described by two attributes per pulse number resulting from the complex electromagnetic signal, processed by an autocorrelation function. We used the two attributes (real and complex) for each pulse as two different modalities and the attribute ``good'' as our target class. The total number of samples in this dataset is 351, out of which 225 are from the target class (good), and the rest of 126 samples are from outlier class (bad).

For the fifth set of experiments, we used Handwritten dataset. We considered the samples of numeral 0 as the target. In the Handwritten dataset, the total number of samples is 2000, out of which 200 are from the target class. The rest of the 1800 samples are considered as an outlier class. We used the Zernike moment (ZER) and morphological (MOR) features as our two different modalities.

\subsection{Experimental setup}\label{expandeva}
For the Robot Execution Failures dataset, Ionosphere dataset, Caltech-7 dataset, and Handwritten dataset, we performed our experiments on 70-30$\%$ split for training and testing sets. We selected the 70-30$\%$ split randomly 5 times, keeping the distribution of classes similar to the original data. To tune the hyperparameters for final testing, we did 5-fold cross-validation on the training set, where the (70$\%$) training data are divided into 5 different sets, and each time one set is used for validation while all the others for training. The process was repeated 5 times until all the sets have been used as validation sets. For SPECTF heart dataset, the train and test sets are given with the dataset. We did 5-fold cross-validation on the training set to optimize the hyperparameters. 

For all datasets, the models were trained by using samples from the positive class only, while testing was carried out using all the classes. The hyperparameters were selected from the following ranges:
\begin{itemize}
  \item$\beta\in\{10^{-4},10^{-3},10^{-2},10^{-1},10^{0},10^{1},10^{2},10^{3},10^{4}\},$
  \item$C\in\{0.01,0.05,0.1,0.2,0.3,0.4,0.5,0.6\}$,
  \item$\sigma\in\{10^{-3},10^{-2},10^{-1},10^{0},10^{1},10^{2},10^{3}\}$,
  \item$d\in\{1,2,3,4,5,10,20,50, 100\}$,
  \item$\eta=0.1$.
\end{itemize}

Here, we restricted the dimension $d$ of the shared subspace as $d<min{\{D_1 ,...,D_M\}}$ for a given dataset, where $D_m$ is the dimensionality of modality $m$. For competing methods, the features from different modalities were concatenated before training the model. We also report the results of the competing methods by considering data from one modality at a time for training and testing. For competing methods, the hyperparameters were selected from the same ranges as mentioned above.

\subsection{Decision strategies}\label{decissions}
During testing, after the common compact representation of all modalities was formed, each representation (modality) of an instance was mapped to the lower-dimensional subspace via corresponding projection matrix and classified as described in Section \ref{testing}. The following four strategies were used to decide the final class for the instance:

\begin{itemize}
\item\textbf{Decision strategy 1} (also called the AND gate)\textbf{:} The test instance is assigned the target label if the representations from all modalities for that particular instance are classified to the target class and the non-target label otherwise.
\item\textbf{Decision strategy 2} (also called as the OR gate)\textbf{:} The final decision is taken on the basis of the OR gate principle, i.e., if a representation of an instance from any of the modalities is classified to the target class, the overall decision for that particular instance is taken in favor of the target class.
\item\textbf{Decision strategy 3:} The final classification decision is made on the basis of first modality, i.e., if the representation from the first modality is assigned to a particular class, the overall classification is made following that.
\item\textbf{Decision strategy 4:} The overall decision is taken on the basis of the label assigned to the representation from the second modality.  
  \end{itemize}
It should be noted that for more than two modalities, different decision strategies, such as majority vote, might be more suitable. 
\subsection{Evaluation criteria}\label{evaluation}
One-class classification models can be evaluated using different metrics. These metrics are decided on the basis of the goals of a given application. For example, in outlier detection, the focus is on detecting negative instances accurately. The most common metrics in one-class classification are true positive rate (\textit{tpr}), and true negative rate (\textit{tnr}). The former, also called as recall, sensitivity, or hit rate, is the proportion of positive instances that is classified by the trained model as positive correctly:
\begin{equation}\label{tpr}
 tpr=\frac{tp}{p},
\end{equation}
where \textit{tp} is the number of positive samples classified correctly and \textit{p} is the total number of positive samples in the test set. The latter, \textit{tnr}, also called as specificity, is defined as
\begin{equation}\label{tnr}
 tnr=\frac{tn}{n},
\end{equation}
where \textit{tn} is the number of negative samples classified correctly and \textit{n} is the total number of negative samples in the test set. Accuracy (\textit{accu}) is measured as the ratio of the number of correctly classified instances to the total number of instances:
\begin{equation}\label{accu}
 accu=\frac{tp+tn}{p+n}.
\end{equation}
Precision ($pre$) measures the proportion of instances classified positive which really are positive:
\begin{equation}\label{precission}
 pre=\frac{tp}{tp+fp},
\end{equation}
where $fp$ is the number of false positives. Another useful measure is \textit{F1} measure, which is the harmonic mean of \textit{pre} and \textit{tpr}:
\begin{equation}\label{f1}
 F1=2\times\frac{pre\times tpr}{pre+tpr}.
\end{equation}
Geometric mean (\textit{gm}) is defined as the square root of the product of sensitivity and specificity:
\begin{equation}\label{gm}
 gm=\sqrt {tpr \times tnr}.
\end{equation}
 \textit{gm} has been used by many researchers for imbalanced datasets. Since it takes into consideration both sensitivity and specificity, we opted to finetune hyperparameters based on the \textit{gm} score on the validation data.

\subsection{Experimental results and discussion}\label{resultsanddiscussion}
\begin{table*}[ht]  \footnotesize\setlength{\tabcolsep}{1.5pt}
  \centering 
       \caption{Test results for Robot Execution Failures dataset}
\begin{tabular}{lllllllllllllll}
\cline{3-8} \cline{10-15}
                                            & \multicolumn{1}{l|}{} & \multicolumn{6}{c|}{Linear}                                                                                                                                                    & \multicolumn{1}{l|}{} & \multicolumn{6}{c|}{Non-linear}                                                                                                                                                    \\ \cline{3-8} \cline{10-15} 
                                            & \multicolumn{1}{l|}{} & \multicolumn{1}{l|}{accu} & \multicolumn{1}{l|}{tpr}  & \multicolumn{1}{l|}{tnr}  & \multicolumn{1}{l|}{pre}  & \multicolumn{1}{l|}{F1}   & \multicolumn{1}{l|}{gm}            & \multicolumn{1}{l|}{} & \multicolumn{1}{l|}{accu} & \multicolumn{1}{l|}{tpr}  & \multicolumn{1}{l|}{tnr}  & \multicolumn{1}{l|}{pre}  & \multicolumn{1}{l|}{F1}   & \multicolumn{1}{l|}{gm}            \\ \cline{3-8} \cline{10-15} 
Proposed method                             &                       &                           &                           &                           &                           &                           &                                    &                       &                           &                           &                           &                           &                           &                                    \\ \cline{1-1} \cline{3-8} \cline{10-15} 
\multicolumn{1}{|l|}{MS-SVDD $\omega_2ds3$} & \multicolumn{1}{l|}{} & \multicolumn{1}{l|}{0.97} & \multicolumn{1}{l|}{0.97} & \multicolumn{1}{l|}{0.97} & \multicolumn{1}{l|}{0.93} & \multicolumn{1}{l|}{0.95} & \multicolumn{1}{l|}{\textbf{0.97}} & \multicolumn{1}{l|}{} & \multicolumn{1}{l|}{0.94} & \multicolumn{1}{l|}{0.98} & \multicolumn{1}{l|}{0.92} & \multicolumn{1}{l|}{0.83} & \multicolumn{1}{l|}{0.90} & \multicolumn{1}{l|}{0.95}          \\ \cline{1-1} \cline{3-8} \cline{10-15} 
\multicolumn{1}{|l|}{MS-SVDD $\omega_5ds3$}  & \multicolumn{1}{l|}{} & \multicolumn{1}{l|}{0.97} & \multicolumn{1}{l|}{0.95} & \multicolumn{1}{l|}{0.97} & \multicolumn{1}{l|}{0.93} & \multicolumn{1}{l|}{0.94} & \multicolumn{1}{l|}{0.96}          & \multicolumn{1}{l|}{} & \multicolumn{1}{l|}{0.94} & \multicolumn{1}{l|}{0.98} & \multicolumn{1}{l|}{0.92} & \multicolumn{1}{l|}{0.83} & \multicolumn{1}{l|}{0.90} & \multicolumn{1}{l|}{0.95}          \\ \cline{1-1} \cline{3-8} \cline{10-15} 
Concatenated features                       &                       &                           &                           &                           &                           &                           &                                    &                       &                           &                           &                           &                           &                           &                                    \\ \cline{1-1} \cline{3-8} \cline{10-15} 
\multicolumn{1}{|l|}{S-SVDD $\psi_1$}       & \multicolumn{1}{l|}{} & \multicolumn{1}{l|}{0.66} & \multicolumn{1}{l|}{0.89} & \multicolumn{1}{l|}{0.57} & \multicolumn{1}{l|}{0.46} & \multicolumn{1}{l|}{0.60} & \multicolumn{1}{l|}{0.71}          & \multicolumn{1}{l|}{} & \multicolumn{1}{l|}{0.94} & \multicolumn{1}{l|}{0.84} & \multicolumn{1}{l|}{0.98} & \multicolumn{1}{l|}{0.95} & \multicolumn{1}{l|}{0.89} & \multicolumn{1}{l|}{0.91}          \\ \cline{1-1} \cline{3-8} \cline{10-15} 
\multicolumn{1}{|l|}{S-SVDD $\psi_2$}       & \multicolumn{1}{l|}{} & \multicolumn{1}{l|}{0.70} & \multicolumn{1}{l|}{0.80} & \multicolumn{1}{l|}{0.66} & \multicolumn{1}{l|}{0.58} & \multicolumn{1}{l|}{0.60} & \multicolumn{1}{l|}{0.70}          & \multicolumn{1}{l|}{} & \multicolumn{1}{l|}{0.92} & \multicolumn{1}{l|}{0.90} & \multicolumn{1}{l|}{0.93} & \multicolumn{1}{l|}{0.84} & \multicolumn{1}{l|}{0.87} & \multicolumn{1}{l|}{0.91}          \\ \cline{1-1} \cline{3-8} \cline{10-15} 
\multicolumn{1}{|l|}{S-SVDD $\psi_3$}       & \multicolumn{1}{l|}{} & \multicolumn{1}{l|}{0.66} & \multicolumn{1}{l|}{0.78} & \multicolumn{1}{l|}{0.61} & \multicolumn{1}{l|}{0.46} & \multicolumn{1}{l|}{0.56} & \multicolumn{1}{l|}{0.67}          & \multicolumn{1}{l|}{} & \multicolumn{1}{l|}{0.93} & \multicolumn{1}{l|}{0.93} & \multicolumn{1}{l|}{0.93} & \multicolumn{1}{l|}{0.85} & \multicolumn{1}{l|}{0.89} & \multicolumn{1}{l|}{0.93}          \\ \cline{1-1} \cline{3-8} \cline{10-15} 
\multicolumn{1}{|l|}{S-SVDD $\psi_4$}       & \multicolumn{1}{l|}{} & \multicolumn{1}{l|}{0.64} & \multicolumn{1}{l|}{0.94} & \multicolumn{1}{l|}{0.52} & \multicolumn{1}{l|}{0.44} & \multicolumn{1}{l|}{0.60} & \multicolumn{1}{l|}{0.70}          & \multicolumn{1}{l|}{} & \multicolumn{1}{l|}{0.96} & \multicolumn{1}{l|}{0.90} & \multicolumn{1}{l|}{0.98} & \multicolumn{1}{l|}{0.96} & \multicolumn{1}{l|}{0.93} & \multicolumn{1}{l|}{0.94}          \\ \cline{1-1} \cline{3-8} \cline{10-15} 
\multicolumn{1}{|l|}{OC-SVM}                & \multicolumn{1}{l|}{} & \multicolumn{1}{l|}{0.51} & \multicolumn{1}{l|}{0.47} & \multicolumn{1}{l|}{0.52} & \multicolumn{1}{l|}{0.28} & \multicolumn{1}{l|}{0.35} & \multicolumn{1}{l|}{0.49}          & \multicolumn{1}{l|}{} & \multicolumn{1}{l|}{0.86} & \multicolumn{1}{l|}{0.49} & \multicolumn{1}{l|}{1.00} & \multicolumn{1}{l|}{1.00} & \multicolumn{1}{l|}{0.65} & \multicolumn{1}{l|}{0.70}          \\ \cline{1-1} \cline{3-8} \cline{10-15} 
\multicolumn{1}{|l|}{SVDD}                  & \multicolumn{1}{l|}{} & \multicolumn{1}{l|}{0.97} & \multicolumn{1}{l|}{0.91} & \multicolumn{1}{l|}{0.99} & \multicolumn{1}{l|}{0.98} & \multicolumn{1}{l|}{0.95} & \multicolumn{1}{l|}{0.95}          & \multicolumn{1}{l|}{} & \multicolumn{1}{l|}{0.95} & \multicolumn{1}{l|}{0.85} & \multicolumn{1}{l|}{0.99} & \multicolumn{1}{l|}{0.98} & \multicolumn{1}{l|}{0.91} & \multicolumn{1}{l|}{0.92}          \\ \cline{1-1} \cline{3-8} \cline{10-15} 
Force measurements                          &                       &                           &                           &                           &                           &                           &                                    &                       &                           &                           &                           &                           &                           &                                    \\ \cline{1-1} \cline{3-8} \cline{10-15} 
\multicolumn{1}{|l|}{S-SVDD $\psi_1$}       & \multicolumn{1}{l|}{} & \multicolumn{1}{l|}{0.76} & \multicolumn{1}{l|}{0.88} & \multicolumn{1}{l|}{0.71} & \multicolumn{1}{l|}{0.55} & \multicolumn{1}{l|}{0.67} & \multicolumn{1}{l|}{0.79}          & \multicolumn{1}{l|}{} & \multicolumn{1}{l|}{0.96} & \multicolumn{1}{l|}{0.90} & \multicolumn{1}{l|}{0.98} & \multicolumn{1}{l|}{0.95} & \multicolumn{1}{l|}{0.92} & \multicolumn{1}{l|}{0.94}          \\ \cline{1-1} \cline{3-8} \cline{10-15} 
\multicolumn{1}{|l|}{S-SVDD $\psi_2$}       & \multicolumn{1}{l|}{} & \multicolumn{1}{l|}{0.77} & \multicolumn{1}{l|}{0.94} & \multicolumn{1}{l|}{0.71} & \multicolumn{1}{l|}{0.56} & \multicolumn{1}{l|}{0.70} & \multicolumn{1}{l|}{0.82}          & \multicolumn{1}{l|}{} & \multicolumn{1}{l|}{0.96} & \multicolumn{1}{l|}{0.90} & \multicolumn{1}{l|}{0.98} & \multicolumn{1}{l|}{0.95} & \multicolumn{1}{l|}{0.92} & \multicolumn{1}{l|}{0.94}          \\ \cline{1-1} \cline{3-8} \cline{10-15} 
\multicolumn{1}{|l|}{S-SVDD $\psi_3$}       & \multicolumn{1}{l|}{} & \multicolumn{1}{l|}{0.73} & \multicolumn{1}{l|}{0.70} & \multicolumn{1}{l|}{0.74} & \multicolumn{1}{l|}{0.51} & \multicolumn{1}{l|}{0.58} & \multicolumn{1}{l|}{0.71}          & \multicolumn{1}{l|}{} & \multicolumn{1}{l|}{0.96} & \multicolumn{1}{l|}{0.91} & \multicolumn{1}{l|}{0.98} & \multicolumn{1}{l|}{0.95} & \multicolumn{1}{l|}{0.93} & \multicolumn{1}{l|}{0.94}          \\ \cline{1-1} \cline{3-8} \cline{10-15} 
\multicolumn{1}{|l|}{S-SVDD $\psi_4$}       & \multicolumn{1}{l|}{} & \multicolumn{1}{l|}{0.76} & \multicolumn{1}{l|}{0.85} & \multicolumn{1}{l|}{0.72} & \multicolumn{1}{l|}{0.54} & \multicolumn{1}{l|}{0.66} & \multicolumn{1}{l|}{0.78}          & \multicolumn{1}{l|}{} & \multicolumn{1}{l|}{0.93} & \multicolumn{1}{l|}{0.82} & \multicolumn{1}{l|}{0.98} & \multicolumn{1}{l|}{0.95} & \multicolumn{1}{l|}{0.84} & \multicolumn{1}{l|}{0.88}          \\ \cline{1-1} \cline{3-8} \cline{10-15} 
\multicolumn{1}{|l|}{OC-SVM}                & \multicolumn{1}{l|}{} & \multicolumn{1}{l|}{0.50} & \multicolumn{1}{l|}{0.53} & \multicolumn{1}{l|}{0.49} & \multicolumn{1}{l|}{0.29} & \multicolumn{1}{l|}{0.37} & \multicolumn{1}{l|}{0.51}          & \multicolumn{1}{l|}{} & \multicolumn{1}{l|}{0.86} & \multicolumn{1}{l|}{0.49} & \multicolumn{1}{l|}{1.00} & \multicolumn{1}{l|}{1.00} & \multicolumn{1}{l|}{0.65} & \multicolumn{1}{l|}{0.70}          \\ \cline{1-1} \cline{3-8} \cline{10-15} 
\multicolumn{1}{|l|}{SVDD}                  & \multicolumn{1}{l|}{} & \multicolumn{1}{l|}{0.97} & \multicolumn{1}{l|}{0.90} & \multicolumn{1}{l|}{0.99} & \multicolumn{1}{l|}{0.98} & \multicolumn{1}{l|}{0.94} & \multicolumn{1}{l|}{0.95}          & \multicolumn{1}{l|}{} & \multicolumn{1}{l|}{0.97} & \multicolumn{1}{l|}{0.92} & \multicolumn{1}{l|}{0.99} & \multicolumn{1}{l|}{0.98} & \multicolumn{1}{l|}{0.95} & \multicolumn{1}{l|}{\textbf{0.96}} \\ \cline{1-1} \cline{3-8} \cline{10-15} 
Torque measurements                         &                       &                           &                           &                           &                           &                           &                                    &                       &                           &                           &                           &                           &                           &                                    \\ \cline{1-1} \cline{3-8} \cline{10-15} 
\multicolumn{1}{|l|}{S-SVDD $\psi_1$}       & \multicolumn{1}{l|}{} & \multicolumn{1}{l|}{0.59} & \multicolumn{1}{l|}{0.96} & \multicolumn{1}{l|}{0.44} & \multicolumn{1}{l|}{0.41} & \multicolumn{1}{l|}{0.57} & \multicolumn{1}{l|}{0.65}          & \multicolumn{1}{l|}{} & \multicolumn{1}{l|}{0.97} & \multicolumn{1}{l|}{0.89} & \multicolumn{1}{l|}{1.00} & \multicolumn{1}{l|}{1.00} & \multicolumn{1}{l|}{0.94} & \multicolumn{1}{l|}{0.94}          \\ \cline{1-1} \cline{3-8} \cline{10-15} 
\multicolumn{1}{|l|}{S-SVDD $\psi_2$}       & \multicolumn{1}{l|}{} & \multicolumn{1}{l|}{0.61} & \multicolumn{1}{l|}{0.94} & \multicolumn{1}{l|}{0.48} & \multicolumn{1}{l|}{0.42} & \multicolumn{1}{l|}{0.57} & \multicolumn{1}{l|}{0.67}          & \multicolumn{1}{l|}{} & \multicolumn{1}{l|}{0.71} & \multicolumn{1}{l|}{0.66} & \multicolumn{1}{l|}{0.73} & \multicolumn{1}{l|}{0.51} & \multicolumn{1}{l|}{0.54} & \multicolumn{1}{l|}{0.51}          \\ \cline{1-1} \cline{3-8} \cline{10-15} 
\multicolumn{1}{|l|}{S-SVDD $\psi_3$}       & \multicolumn{1}{l|}{} & \multicolumn{1}{l|}{0.62} & \multicolumn{1}{l|}{0.92} & \multicolumn{1}{l|}{0.50} & \multicolumn{1}{l|}{0.43} & \multicolumn{1}{l|}{0.58} & \multicolumn{1}{l|}{0.67}          & \multicolumn{1}{l|}{} & \multicolumn{1}{l|}{0.92} & \multicolumn{1}{l|}{0.76} & \multicolumn{1}{l|}{0.99} & \multicolumn{1}{l|}{0.97} & \multicolumn{1}{l|}{0.82} & \multicolumn{1}{l|}{0.85}          \\ \cline{1-1} \cline{3-8} \cline{10-15} 
\multicolumn{1}{|l|}{S-SVDD $\psi_4$}       & \multicolumn{1}{l|}{} & \multicolumn{1}{l|}{0.61} & \multicolumn{1}{l|}{0.96} & \multicolumn{1}{l|}{0.48} & \multicolumn{1}{l|}{0.42} & \multicolumn{1}{l|}{0.58} & \multicolumn{1}{l|}{0.68}          & \multicolumn{1}{l|}{} & \multicolumn{1}{l|}{0.76} & \multicolumn{1}{l|}{0.76} & \multicolumn{1}{l|}{0.76} & \multicolumn{1}{l|}{0.76} & \multicolumn{1}{l|}{0.71} & \multicolumn{1}{l|}{0.66}          \\ \cline{1-1} \cline{3-8} \cline{10-15} 
\multicolumn{1}{|l|}{OC-SVM}                & \multicolumn{1}{l|}{} & \multicolumn{1}{l|}{0.52} & \multicolumn{1}{l|}{0.59} & \multicolumn{1}{l|}{0.49} & \multicolumn{1}{l|}{0.31} & \multicolumn{1}{l|}{0.40} & \multicolumn{1}{l|}{0.53}          & \multicolumn{1}{l|}{} & \multicolumn{1}{l|}{0.84} & \multicolumn{1}{l|}{0.58} & \multicolumn{1}{l|}{0.94} & \multicolumn{1}{l|}{0.81} & \multicolumn{1}{l|}{0.66} & \multicolumn{1}{l|}{0.73}          \\ \cline{1-1} \cline{3-8} \cline{10-15} 
\multicolumn{1}{|l|}{SVDD}                  & \multicolumn{1}{l|}{} & \multicolumn{1}{l|}{0.90} & \multicolumn{1}{l|}{0.95} & \multicolumn{1}{l|}{0.88} & \multicolumn{1}{l|}{0.76} & \multicolumn{1}{l|}{0.84} & \multicolumn{1}{l|}{0.91}          & \multicolumn{1}{l|}{} & \multicolumn{1}{l|}{0.91} & \multicolumn{1}{l|}{0.88} & \multicolumn{1}{l|}{0.92} & \multicolumn{1}{l|}{0.81} & \multicolumn{1}{l|}{0.84} & \multicolumn{1}{l|}{0.90}          \\ \cline{1-1} \cline{3-8} \cline{10-15} 
\end{tabular}
\label{resultsrobot}
\end{table*}

\begin{table*}[ht]  \footnotesize\setlength{\tabcolsep}{1.5pt}
  \centering 
       \caption{Test results for Caltech-7 dataset}
\begin{tabular}{lllllllllllllll}
\cline{3-8} \cline{10-15}
                                            & \multicolumn{1}{l|}{} & \multicolumn{6}{c|}{Linear}                                                                                                                                                    & \multicolumn{1}{l|}{} & \multicolumn{6}{c|}{Non-linear}                                                                                                                                                \\ \cline{3-8} \cline{10-15} 
                                            & \multicolumn{1}{l|}{} & \multicolumn{1}{l|}{accu} & \multicolumn{1}{l|}{tpr}  & \multicolumn{1}{l|}{tnr}  & \multicolumn{1}{l|}{pre}  & \multicolumn{1}{l|}{F1}   & \multicolumn{1}{l|}{gm}            & \multicolumn{1}{l|}{} & \multicolumn{1}{l|}{accu} & \multicolumn{1}{l|}{tpr}  & \multicolumn{1}{l|}{tnr}  & \multicolumn{1}{l|}{pre}  & \multicolumn{1}{l|}{F1}   & \multicolumn{1}{l|}{gm}            \\ \cline{3-8} \cline{10-15} 
Proposed method                             &                       &                           &                           &                           &                           &                           &                                    &                       &                           &                           &                           &                           &                           &                                    \\ \cline{1-1} \cline{3-8} \cline{10-15} 
\multicolumn{1}{|l|}{MS-SVDD $\omega_1ds1$} & \multicolumn{1}{l|}{} & \multicolumn{1}{l|}{0.91} & \multicolumn{1}{l|}{0.96} & \multicolumn{1}{l|}{0.89} & \multicolumn{1}{l|}{0.78} & \multicolumn{1}{l|}{0.86} & \multicolumn{1}{l|}{0.92}          & \multicolumn{1}{l|}{} & \multicolumn{1}{l|}{0.94} & \multicolumn{1}{l|}{0.98} & \multicolumn{1}{l|}{0.92} & \multicolumn{1}{l|}{0.85} & \multicolumn{1}{l|}{0.91} & \multicolumn{1}{l|}{\textbf{0.95}} \\ \cline{1-1} \cline{3-8} \cline{10-15} 
\multicolumn{1}{|l|}{MS-SVDD $\omega_4ds1$} & \multicolumn{1}{l|}{} & \multicolumn{1}{l|}{0.91} & \multicolumn{1}{l|}{0.95} & \multicolumn{1}{l|}{0.89} & \multicolumn{1}{l|}{0.78} & \multicolumn{1}{l|}{0.86} & \multicolumn{1}{l|}{0.92}          & \multicolumn{1}{l|}{} & \multicolumn{1}{l|}{0.94} & \multicolumn{1}{l|}{0.95} & \multicolumn{1}{l|}{0.94} & \multicolumn{1}{l|}{0.88} & \multicolumn{1}{l|}{0.91} & \multicolumn{1}{l|}{\textbf{0.95}} \\ \cline{1-1} \cline{3-8} \cline{10-15} 
Concatenated features                       &                       &                           &                           &                           &                           &                           &                                    &                       &                           &                           &                           &                           &                           &                                    \\ \cline{1-1} \cline{3-8} \cline{10-15} 
\multicolumn{1}{|l|}{S-SVDD $\psi_1$}       & \multicolumn{1}{l|}{} & \multicolumn{1}{l|}{0.65} & \multicolumn{1}{l|}{0.96} & \multicolumn{1}{l|}{0.52} & \multicolumn{1}{l|}{0.46} & \multicolumn{1}{l|}{0.62} & \multicolumn{1}{l|}{0.71}          & \multicolumn{1}{l|}{} & \multicolumn{1}{l|}{0.37} & \multicolumn{1}{l|}{0.35} & \multicolumn{1}{l|}{0.38} & \multicolumn{1}{l|}{0.15} & \multicolumn{1}{l|}{0.20} & \multicolumn{1}{l|}{0.23}          \\ \cline{1-1} \cline{3-8} \cline{10-15} 
\multicolumn{1}{|l|}{S-SVDD $\psi_2$}       & \multicolumn{1}{l|}{} & \multicolumn{1}{l|}{0.67} & \multicolumn{1}{l|}{0.92} & \multicolumn{1}{l|}{0.57} & \multicolumn{1}{l|}{0.48} & \multicolumn{1}{l|}{0.63} & \multicolumn{1}{l|}{0.72}          & \multicolumn{1}{l|}{} & \multicolumn{1}{l|}{0.66} & \multicolumn{1}{l|}{0.69} & \multicolumn{1}{l|}{0.64} & \multicolumn{1}{l|}{0.39} & \multicolumn{1}{l|}{0.48} & \multicolumn{1}{l|}{0.53}          \\ \cline{1-1} \cline{3-8} \cline{10-15} 
\multicolumn{1}{|l|}{S-SVDD $\psi_3$}       & \multicolumn{1}{l|}{} & \multicolumn{1}{l|}{0.71} & \multicolumn{1}{l|}{0.84} & \multicolumn{1}{l|}{0.66} & \multicolumn{1}{l|}{0.59} & \multicolumn{1}{l|}{0.65} & \multicolumn{1}{l|}{0.69}          & \multicolumn{1}{l|}{} & \multicolumn{1}{l|}{0.90} & \multicolumn{1}{l|}{0.79} & \multicolumn{1}{l|}{0.94} & \multicolumn{1}{l|}{0.86} & \multicolumn{1}{l|}{0.81} & \multicolumn{1}{l|}{0.86}          \\ \cline{1-1} \cline{3-8} \cline{10-15} 
\multicolumn{1}{|l|}{S-SVDD $\psi_4$}       & \multicolumn{1}{l|}{} & \multicolumn{1}{l|}{0.62} & \multicolumn{1}{l|}{0.96} & \multicolumn{1}{l|}{0.47} & \multicolumn{1}{l|}{0.46} & \multicolumn{1}{l|}{0.61} & \multicolumn{1}{l|}{0.66}          & \multicolumn{1}{l|}{} & \multicolumn{1}{l|}{0.87} & \multicolumn{1}{l|}{0.61} & \multicolumn{1}{l|}{0.97} & \multicolumn{1}{l|}{0.91} & \multicolumn{1}{l|}{0.72} & \multicolumn{1}{l|}{0.76}          \\ \cline{1-1} \cline{3-8} \cline{10-15} 
\multicolumn{1}{|l|}{OC-SVM}                & \multicolumn{1}{l|}{} & \multicolumn{1}{l|}{0.22} & \multicolumn{1}{l|}{0.47} & \multicolumn{1}{l|}{0.12} & \multicolumn{1}{l|}{0.18} & \multicolumn{1}{l|}{0.26} & \multicolumn{1}{l|}{0.22}          & \multicolumn{1}{l|}{} & \multicolumn{1}{l|}{0.86} & \multicolumn{1}{l|}{0.53} & \multicolumn{1}{l|}{1.00} & \multicolumn{1}{l|}{0.99} & \multicolumn{1}{l|}{0.69} & \multicolumn{1}{l|}{0.73}          \\ \cline{1-1} \cline{3-8} \cline{10-15} 
\multicolumn{1}{|l|}{SVDD}                  & \multicolumn{1}{l|}{} & \multicolumn{1}{l|}{0.92} & \multicolumn{1}{l|}{0.94} & \multicolumn{1}{l|}{0.91} & \multicolumn{1}{l|}{0.81} & \multicolumn{1}{l|}{0.87} & \multicolumn{1}{l|}{\textbf{0.93}} & \multicolumn{1}{l|}{} & \multicolumn{1}{l|}{0.96} & \multicolumn{1}{l|}{0.94} & \multicolumn{1}{l|}{0.97} & \multicolumn{1}{l|}{0.93} & \multicolumn{1}{l|}{0.94} & \multicolumn{1}{l|}{\textbf{0.95}} \\ \cline{1-1} \cline{3-8} \cline{10-15} 
Gabor feature                               &                       &                           &                           &                           &                           &                           &                                    &                       &                           &                           &                           &                           &                           &                                    \\ \cline{1-1} \cline{3-8} \cline{10-15} 
\multicolumn{1}{|l|}{S-SVDD $\psi_1$}       & \multicolumn{1}{l|}{} & \multicolumn{1}{l|}{0.68} & \multicolumn{1}{l|}{0.72} & \multicolumn{1}{l|}{0.67} & \multicolumn{1}{l|}{0.47} & \multicolumn{1}{l|}{0.57} & \multicolumn{1}{l|}{0.69}          & \multicolumn{1}{l|}{} & \multicolumn{1}{l|}{0.46} & \multicolumn{1}{l|}{0.84} & \multicolumn{1}{l|}{0.31} & \multicolumn{1}{l|}{0.33} & \multicolumn{1}{l|}{0.48} & \multicolumn{1}{l|}{0.50}          \\ \cline{1-1} \cline{3-8} \cline{10-15} 
\multicolumn{1}{|l|}{S-SVDD $\psi_2$}       & \multicolumn{1}{l|}{} & \multicolumn{1}{l|}{0.68} & \multicolumn{1}{l|}{0.72} & \multicolumn{1}{l|}{0.67} & \multicolumn{1}{l|}{0.47} & \multicolumn{1}{l|}{0.57} & \multicolumn{1}{l|}{0.69}          & \multicolumn{1}{l|}{} & \multicolumn{1}{l|}{0.54} & \multicolumn{1}{l|}{0.78} & \multicolumn{1}{l|}{0.44} & \multicolumn{1}{l|}{0.46} & \multicolumn{1}{l|}{0.52} & \multicolumn{1}{l|}{0.50}          \\ \cline{1-1} \cline{3-8} \cline{10-15} 
\multicolumn{1}{|l|}{S-SVDD $\psi_3$}       & \multicolumn{1}{l|}{} & \multicolumn{1}{l|}{0.61} & \multicolumn{1}{l|}{0.74} & \multicolumn{1}{l|}{0.55} & \multicolumn{1}{l|}{0.45} & \multicolumn{1}{l|}{0.52} & \multicolumn{1}{l|}{0.58}          & \multicolumn{1}{l|}{} & \multicolumn{1}{l|}{0.76} & \multicolumn{1}{l|}{0.68} & \multicolumn{1}{l|}{0.80} & \multicolumn{1}{l|}{0.65} & \multicolumn{1}{l|}{0.63} & \multicolumn{1}{l|}{0.71}          \\ \cline{1-1} \cline{3-8} \cline{10-15} 
\multicolumn{1}{|l|}{S-SVDD $\psi_4$}       & \multicolumn{1}{l|}{} & \multicolumn{1}{l|}{0.70} & \multicolumn{1}{l|}{0.74} & \multicolumn{1}{l|}{0.68} & \multicolumn{1}{l|}{0.49} & \multicolumn{1}{l|}{0.59} & \multicolumn{1}{l|}{0.71}          & \multicolumn{1}{l|}{} & \multicolumn{1}{l|}{0.78} & \multicolumn{1}{l|}{0.39} & \multicolumn{1}{l|}{0.94} & \multicolumn{1}{l|}{0.80} & \multicolumn{1}{l|}{0.46} & \multicolumn{1}{l|}{0.55}          \\ \cline{1-1} \cline{3-8} \cline{10-15} 
\multicolumn{1}{|l|}{OC-SVM}                & \multicolumn{1}{l|}{} & \multicolumn{1}{l|}{0.43} & \multicolumn{1}{l|}{0.53} & \multicolumn{1}{l|}{0.40} & \multicolumn{1}{l|}{0.27} & \multicolumn{1}{l|}{0.36} & \multicolumn{1}{l|}{0.45}          & \multicolumn{1}{l|}{} & \multicolumn{1}{l|}{0.79} & \multicolumn{1}{l|}{0.55} & \multicolumn{1}{l|}{0.89} & \multicolumn{1}{l|}{0.69} & \multicolumn{1}{l|}{0.61} & \multicolumn{1}{l|}{0.70}          \\ \cline{1-1} \cline{3-8} \cline{10-15} 
\multicolumn{1}{|l|}{SVDD}                  & \multicolumn{1}{l|}{} & \multicolumn{1}{l|}{0.76} & \multicolumn{1}{l|}{0.70} & \multicolumn{1}{l|}{0.78} & \multicolumn{1}{l|}{0.57} & \multicolumn{1}{l|}{0.63} & \multicolumn{1}{l|}{0.74}          & \multicolumn{1}{l|}{} & \multicolumn{1}{l|}{0.74} & \multicolumn{1}{l|}{0.92} & \multicolumn{1}{l|}{0.67} & \multicolumn{1}{l|}{0.55} & \multicolumn{1}{l|}{0.68} & \multicolumn{1}{l|}{0.78}          \\ \cline{1-1} \cline{3-8} \cline{10-15} 
Wavelet moments                             &                       &                           &                           &                           &                           &                           &                                    &                       &                           &                           &                           &                           &                           &                                    \\ \cline{1-1} \cline{3-8} \cline{10-15} 
\multicolumn{1}{|l|}{S-SVDD $\psi_1$}       & \multicolumn{1}{l|}{} & \multicolumn{1}{l|}{0.70} & \multicolumn{1}{l|}{0.73} & \multicolumn{1}{l|}{0.68} & \multicolumn{1}{l|}{0.50} & \multicolumn{1}{l|}{0.59} & \multicolumn{1}{l|}{0.69}          & \multicolumn{1}{l|}{} & \multicolumn{1}{l|}{0.54} & \multicolumn{1}{l|}{0.44} & \multicolumn{1}{l|}{0.58} & \multicolumn{1}{l|}{0.22} & \multicolumn{1}{l|}{0.26} & \multicolumn{1}{l|}{0.24}          \\ \cline{1-1} \cline{3-8} \cline{10-15} 
\multicolumn{1}{|l|}{S-SVDD $\psi_2$}       & \multicolumn{1}{l|}{} & \multicolumn{1}{l|}{0.71} & \multicolumn{1}{l|}{0.73} & \multicolumn{1}{l|}{0.70} & \multicolumn{1}{l|}{0.52} & \multicolumn{1}{l|}{0.60} & \multicolumn{1}{l|}{0.70}          & \multicolumn{1}{l|}{} & \multicolumn{1}{l|}{0.51} & \multicolumn{1}{l|}{0.93} & \multicolumn{1}{l|}{0.33} & \multicolumn{1}{l|}{0.41} & \multicolumn{1}{l|}{0.55} & \multicolumn{1}{l|}{0.42}          \\ \cline{1-1} \cline{3-8} \cline{10-15} 
\multicolumn{1}{|l|}{S-SVDD $\psi_3$}       & \multicolumn{1}{l|}{} & \multicolumn{1}{l|}{0.50} & \multicolumn{1}{l|}{0.93} & \multicolumn{1}{l|}{0.33} & \multicolumn{1}{l|}{0.38} & \multicolumn{1}{l|}{0.54} & \multicolumn{1}{l|}{0.50}          & \multicolumn{1}{l|}{} & \multicolumn{1}{l|}{0.79} & \multicolumn{1}{l|}{0.38} & \multicolumn{1}{l|}{0.96} & \multicolumn{1}{l|}{0.65} & \multicolumn{1}{l|}{0.44} & \multicolumn{1}{l|}{0.51}          \\ \cline{1-1} \cline{3-8} \cline{10-15} 
\multicolumn{1}{|l|}{S-SVDD $\psi_4$}       & \multicolumn{1}{l|}{} & \multicolumn{1}{l|}{0.56} & \multicolumn{1}{l|}{0.88} & \multicolumn{1}{l|}{0.42} & \multicolumn{1}{l|}{0.40} & \multicolumn{1}{l|}{0.54} & \multicolumn{1}{l|}{0.59}          & \multicolumn{1}{l|}{} & \multicolumn{1}{l|}{0.61} & \multicolumn{1}{l|}{0.51} & \multicolumn{1}{l|}{0.65} & \multicolumn{1}{l|}{0.50} & \multicolumn{1}{l|}{0.36} & \multicolumn{1}{l|}{0.30}          \\ \cline{1-1} \cline{3-8} \cline{10-15} 
\multicolumn{1}{|l|}{OC-SVM}                & \multicolumn{1}{l|}{} & \multicolumn{1}{l|}{0.21} & \multicolumn{1}{l|}{0.48} & \multicolumn{1}{l|}{0.10} & \multicolumn{1}{l|}{0.18} & \multicolumn{1}{l|}{0.26} & \multicolumn{1}{l|}{0.21}          & \multicolumn{1}{l|}{} & \multicolumn{1}{l|}{0.84} & \multicolumn{1}{l|}{0.48} & \multicolumn{1}{l|}{0.99} & \multicolumn{1}{l|}{0.97} & \multicolumn{1}{l|}{0.64} & \multicolumn{1}{l|}{0.69}          \\ \cline{1-1} \cline{3-8} \cline{10-15} 
\multicolumn{1}{|l|}{SVDD}                  & \multicolumn{1}{l|}{} & \multicolumn{1}{l|}{0.91} & \multicolumn{1}{l|}{0.94} & \multicolumn{1}{l|}{0.89} & \multicolumn{1}{l|}{0.79} & \multicolumn{1}{l|}{0.85} & \multicolumn{1}{l|}{0.91}          & \multicolumn{1}{l|}{} & \multicolumn{1}{l|}{0.94} & \multicolumn{1}{l|}{0.97} & \multicolumn{1}{l|}{0.93} & \multicolumn{1}{l|}{0.85} & \multicolumn{1}{l|}{0.91} & \multicolumn{1}{l|}{\textbf{0.95}} \\ \cline{1-1} \cline{3-8} \cline{10-15} 
\end{tabular}
\label{resultscaltech}
\end{table*}

\begin{table*}[ht]  \footnotesize\setlength{\tabcolsep}{1.5pt}
  \centering 
       \caption{Test results for Ionosphere dataset}
\begin{tabular}{lllllllllllllll}
\cline{3-8} \cline{10-15}
                                            & \multicolumn{1}{l|}{} & \multicolumn{6}{c|}{Linear}                                                                                                                                                    & \multicolumn{1}{l|}{} & \multicolumn{6}{c|}{Non-linear}                                                                                                                                                \\ \cline{3-8} \cline{10-15} 
                                            & \multicolumn{1}{l|}{} & \multicolumn{1}{l|}{accu} & \multicolumn{1}{l|}{tpr}  & \multicolumn{1}{l|}{tnr}  & \multicolumn{1}{l|}{pre}  & \multicolumn{1}{l|}{F1}   & \multicolumn{1}{l|}{gm}            & \multicolumn{1}{l|}{} & \multicolumn{1}{l|}{accu} & \multicolumn{1}{l|}{tpr}  & \multicolumn{1}{l|}{tnr}  & \multicolumn{1}{l|}{pre}  & \multicolumn{1}{l|}{F1}   & \multicolumn{1}{l|}{gm}            \\ \cline{3-8} \cline{10-15} 
Proposed method                             &                       &                           &                           &                           &                           &                           &                                    &                       &                           &                           &                           &                           &                           &                                    \\ \cline{1-1} \cline{3-8} \cline{10-15} 
\multicolumn{1}{|l|}{MS-SVDD $\omega_2ds4$} & \multicolumn{1}{l|}{} & \multicolumn{1}{l|}{0.87} & \multicolumn{1}{l|}{0.95} & \multicolumn{1}{l|}{0.73} & \multicolumn{1}{l|}{0.87} & \multicolumn{1}{l|}{0.91} & \multicolumn{1}{l|}{0.83}          & \multicolumn{1}{l|}{} & \multicolumn{1}{l|}{0.76} & \multicolumn{1}{l|}{0.86} & \multicolumn{1}{l|}{0.59} & \multicolumn{1}{l|}{0.79} & \multicolumn{1}{l|}{0.82} & \multicolumn{1}{l|}{0.71}          \\ \cline{1-1} \cline{3-8} \cline{10-15} 
\multicolumn{1}{|l|}{MS-SVDD $\omega_1ds4$} & \multicolumn{1}{l|}{} & \multicolumn{1}{l|}{0.83} & \multicolumn{1}{l|}{0.91} & \multicolumn{1}{l|}{0.69} & \multicolumn{1}{l|}{0.84} & \multicolumn{1}{l|}{0.87} & \multicolumn{1}{l|}{0.79}          & \multicolumn{1}{l|}{} & \multicolumn{1}{l|}{0.88} & \multicolumn{1}{l|}{0.95} & \multicolumn{1}{l|}{0.74} & \multicolumn{1}{l|}{0.87} & \multicolumn{1}{l|}{0.91} & \multicolumn{1}{l|}{0.84}          \\ \cline{1-1} \cline{3-8} \cline{10-15} 
Concatenated features                       &                       &                           &                           &                           &                           &                           &                                    &                       &                           &                           &                           &                           &                           &                                    \\ \cline{1-1} \cline{3-8} \cline{10-15} 
\multicolumn{1}{|l|}{S-SVDD $\psi_1$}       & \multicolumn{1}{l|}{} & \multicolumn{1}{l|}{0.69} & \multicolumn{1}{l|}{0.88} & \multicolumn{1}{l|}{0.32} & \multicolumn{1}{l|}{0.69} & \multicolumn{1}{l|}{0.77} & \multicolumn{1}{l|}{0.53}          & \multicolumn{1}{l|}{} & \multicolumn{1}{l|}{0.49} & \multicolumn{1}{l|}{0.37} & \multicolumn{1}{l|}{0.69} & \multicolumn{1}{l|}{0.55} & \multicolumn{1}{l|}{0.39} & \multicolumn{1}{l|}{0.28}          \\ \cline{1-1} \cline{3-8} \cline{10-15} 
\multicolumn{1}{|l|}{S-SVDD $\psi_2$}       & \multicolumn{1}{l|}{} & \multicolumn{1}{l|}{0.69} & \multicolumn{1}{l|}{0.89} & \multicolumn{1}{l|}{0.31} & \multicolumn{1}{l|}{0.69} & \multicolumn{1}{l|}{0.77} & \multicolumn{1}{l|}{0.51}          & \multicolumn{1}{l|}{} & \multicolumn{1}{l|}{0.74} & \multicolumn{1}{l|}{0.60} & \multicolumn{1}{l|}{0.98} & \multicolumn{1}{l|}{0.98} & \multicolumn{1}{l|}{0.75} & \multicolumn{1}{l|}{0.77}          \\ \cline{1-1} \cline{3-8} \cline{10-15} 
\multicolumn{1}{|l|}{S-SVDD $\psi_3$}       & \multicolumn{1}{l|}{} & \multicolumn{1}{l|}{0.58} & \multicolumn{1}{l|}{0.63} & \multicolumn{1}{l|}{0.48} & \multicolumn{1}{l|}{0.66} & \multicolumn{1}{l|}{0.62} & \multicolumn{1}{l|}{0.51}          & \multicolumn{1}{l|}{} & \multicolumn{1}{l|}{0.72} & \multicolumn{1}{l|}{0.77} & \multicolumn{1}{l|}{0.62} & \multicolumn{1}{l|}{0.83} & \multicolumn{1}{l|}{0.77} & \multicolumn{1}{l|}{0.63}          \\ \cline{1-1} \cline{3-8} \cline{10-15} 
\multicolumn{1}{|l|}{S-SVDD $\psi_4$}       & \multicolumn{1}{l|}{} & \multicolumn{1}{l|}{0.72} & \multicolumn{1}{l|}{0.98} & \multicolumn{1}{l|}{0.23} & \multicolumn{1}{l|}{0.70} & \multicolumn{1}{l|}{0.82} & \multicolumn{1}{l|}{0.43}          & \multicolumn{1}{l|}{} & \multicolumn{1}{l|}{0.66} & \multicolumn{1}{l|}{0.61} & \multicolumn{1}{l|}{0.77} & \multicolumn{1}{l|}{0.88} & \multicolumn{1}{l|}{0.67} & \multicolumn{1}{l|}{0.62}          \\ \cline{1-1} \cline{3-8} \cline{10-15} 
\multicolumn{1}{|l|}{OC-SVM}                & \multicolumn{1}{l|}{} & \multicolumn{1}{l|}{0.38} & \multicolumn{1}{l|}{0.39} & \multicolumn{1}{l|}{0.34} & \multicolumn{1}{l|}{0.52} & \multicolumn{1}{l|}{0.45} & \multicolumn{1}{l|}{0.37}          & \multicolumn{1}{l|}{} & \multicolumn{1}{l|}{0.66} & \multicolumn{1}{l|}{0.48} & \multicolumn{1}{l|}{0.97} & \multicolumn{1}{l|}{0.97} & \multicolumn{1}{l|}{0.63} & \multicolumn{1}{l|}{0.67}          \\ \cline{1-1} \cline{3-8} \cline{10-15} 
\multicolumn{1}{|l|}{SVDD}                  & \multicolumn{1}{l|}{} & \multicolumn{1}{l|}{0.87} & \multicolumn{1}{l|}{0.93} & \multicolumn{1}{l|}{0.76} & \multicolumn{1}{l|}{0.88} & \multicolumn{1}{l|}{0.90} & \multicolumn{1}{l|}{\textbf{0.84}} & \multicolumn{1}{l|}{} & \multicolumn{1}{l|}{0.89} & \multicolumn{1}{l|}{0.94} & \multicolumn{1}{l|}{0.78} & \multicolumn{1}{l|}{0.89} & \multicolumn{1}{l|}{0.92} & \multicolumn{1}{l|}{0.86}          \\ \cline{1-1} \cline{3-8} \cline{10-15} 
Real                                        &                       &                           &                           &                           &                           &                           &                                    &                       &                           &                           &                           &                           &                           &                                    \\ \cline{1-1} \cline{3-8} \cline{10-15} 
\multicolumn{1}{|l|}{S-SVDD $\psi_1$}       & \multicolumn{1}{l|}{} & \multicolumn{1}{l|}{0.81} & \multicolumn{1}{l|}{0.99} & \multicolumn{1}{l|}{0.50} & \multicolumn{1}{l|}{0.78} & \multicolumn{1}{l|}{0.87} & \multicolumn{1}{l|}{0.69}          & \multicolumn{1}{l|}{} & \multicolumn{1}{l|}{0.54} & \multicolumn{1}{l|}{0.36} & \multicolumn{1}{l|}{0.86} & \multicolumn{1}{l|}{0.67} & \multicolumn{1}{l|}{0.43} & \multicolumn{1}{l|}{0.46}          \\ \cline{1-1} \cline{3-8} \cline{10-15} 
\multicolumn{1}{|l|}{S-SVDD $\psi_2$}       & \multicolumn{1}{l|}{} & \multicolumn{1}{l|}{0.80} & \multicolumn{1}{l|}{0.99} & \multicolumn{1}{l|}{0.47} & \multicolumn{1}{l|}{0.78} & \multicolumn{1}{l|}{0.87} & \multicolumn{1}{l|}{0.67}          & \multicolumn{1}{l|}{} & \multicolumn{1}{l|}{0.62} & \multicolumn{1}{l|}{0.49} & \multicolumn{1}{l|}{0.86} & \multicolumn{1}{l|}{0.87} & \multicolumn{1}{l|}{0.61} & \multicolumn{1}{l|}{0.64}          \\ \cline{1-1} \cline{3-8} \cline{10-15} 
\multicolumn{1}{|l|}{S-SVDD $\psi_3$}       & \multicolumn{1}{l|}{} & \multicolumn{1}{l|}{0.81} & \multicolumn{1}{l|}{0.99} & \multicolumn{1}{l|}{0.49} & \multicolumn{1}{l|}{0.78} & \multicolumn{1}{l|}{0.87} & \multicolumn{1}{l|}{0.68}          & \multicolumn{1}{l|}{} & \multicolumn{1}{l|}{0.68} & \multicolumn{1}{l|}{0.63} & \multicolumn{1}{l|}{0.78} & \multicolumn{1}{l|}{0.86} & \multicolumn{1}{l|}{0.70} & \multicolumn{1}{l|}{0.68}          \\ \cline{1-1} \cline{3-8} \cline{10-15} 
\multicolumn{1}{|l|}{S-SVDD $\psi_4$}       & \multicolumn{1}{l|}{} & \multicolumn{1}{l|}{0.81} & \multicolumn{1}{l|}{0.99} & \multicolumn{1}{l|}{0.50} & \multicolumn{1}{l|}{0.78} & \multicolumn{1}{l|}{0.87} & \multicolumn{1}{l|}{0.70}          & \multicolumn{1}{l|}{} & \multicolumn{1}{l|}{0.58} & \multicolumn{1}{l|}{0.45} & \multicolumn{1}{l|}{0.83} & \multicolumn{1}{l|}{0.85} & \multicolumn{1}{l|}{0.53} & \multicolumn{1}{l|}{0.56}          \\ \cline{1-1} \cline{3-8} \cline{10-15} 
\multicolumn{1}{|l|}{OC-SVM}                & \multicolumn{1}{l|}{} & \multicolumn{1}{l|}{0.49} & \multicolumn{1}{l|}{0.52} & \multicolumn{1}{l|}{0.42} & \multicolumn{1}{l|}{0.61} & \multicolumn{1}{l|}{0.56} & \multicolumn{1}{l|}{0.46}          & \multicolumn{1}{l|}{} & \multicolumn{1}{l|}{0.68} & \multicolumn{1}{l|}{0.56} & \multicolumn{1}{l|}{0.89} & \multicolumn{1}{l|}{0.93} & \multicolumn{1}{l|}{0.67} & \multicolumn{1}{l|}{0.69}          \\ \cline{1-1} \cline{3-8} \cline{10-15} 
\multicolumn{1}{|l|}{SVDD}                  & \multicolumn{1}{l|}{} & \multicolumn{1}{l|}{0.88} & \multicolumn{1}{l|}{0.95} & \multicolumn{1}{l|}{0.74} & \multicolumn{1}{l|}{0.87} & \multicolumn{1}{l|}{0.91} & \multicolumn{1}{l|}{\textbf{0.84}} & \multicolumn{1}{l|}{} & \multicolumn{1}{l|}{0.89} & \multicolumn{1}{l|}{0.94} & \multicolumn{1}{l|}{0.81} & \multicolumn{1}{l|}{0.90} & \multicolumn{1}{l|}{0.92} & \multicolumn{1}{l|}{\textbf{0.87}} \\ \cline{1-1} \cline{3-8} \cline{10-15} 
Complex                                     &                       &                           &                           &                           &                           &                           &                                    &                       &                           &                           &                           &                           &                           &                                    \\ \cline{1-1} \cline{3-8} \cline{10-15} 
\multicolumn{1}{|l|}{S-SVDD $\psi_1$}       & \multicolumn{1}{l|}{} & \multicolumn{1}{l|}{0.50} & \multicolumn{1}{l|}{0.37} & \multicolumn{1}{l|}{0.72} & \multicolumn{1}{l|}{0.70} & \multicolumn{1}{l|}{0.49} & \multicolumn{1}{l|}{0.51}          & \multicolumn{1}{l|}{} & \multicolumn{1}{l|}{0.43} & \multicolumn{1}{l|}{0.27} & \multicolumn{1}{l|}{0.71} & \multicolumn{1}{l|}{0.52} & \multicolumn{1}{l|}{0.30} & \multicolumn{1}{l|}{0.34}          \\ \cline{1-1} \cline{3-8} \cline{10-15} 
\multicolumn{1}{|l|}{S-SVDD $\psi_2$}       & \multicolumn{1}{l|}{} & \multicolumn{1}{l|}{0.47} & \multicolumn{1}{l|}{0.35} & \multicolumn{1}{l|}{0.69} & \multicolumn{1}{l|}{0.67} & \multicolumn{1}{l|}{0.46} & \multicolumn{1}{l|}{0.49}          & \multicolumn{1}{l|}{} & \multicolumn{1}{l|}{0.66} & \multicolumn{1}{l|}{0.56} & \multicolumn{1}{l|}{0.83} & \multicolumn{1}{l|}{0.85} & \multicolumn{1}{l|}{0.68} & \multicolumn{1}{l|}{0.68}          \\ \cline{1-1} \cline{3-8} \cline{10-15} 
\multicolumn{1}{|l|}{S-SVDD $\psi_3$}       & \multicolumn{1}{l|}{} & \multicolumn{1}{l|}{0.53} & \multicolumn{1}{l|}{0.57} & \multicolumn{1}{l|}{0.46} & \multicolumn{1}{l|}{0.67} & \multicolumn{1}{l|}{0.58} & \multicolumn{1}{l|}{0.39}          & \multicolumn{1}{l|}{} & \multicolumn{1}{l|}{0.65} & \multicolumn{1}{l|}{0.65} & \multicolumn{1}{l|}{0.65} & \multicolumn{1}{l|}{0.78} & \multicolumn{1}{l|}{0.70} & \multicolumn{1}{l|}{0.63}          \\ \cline{1-1} \cline{3-8} \cline{10-15} 
\multicolumn{1}{|l|}{S-SVDD $\psi_4$}       & \multicolumn{1}{l|}{} & \multicolumn{1}{l|}{0.50} & \multicolumn{1}{l|}{0.38} & \multicolumn{1}{l|}{0.72} & \multicolumn{1}{l|}{0.70} & \multicolumn{1}{l|}{0.49} & \multicolumn{1}{l|}{0.52}          & \multicolumn{1}{l|}{} & \multicolumn{1}{l|}{0.63} & \multicolumn{1}{l|}{0.64} & \multicolumn{1}{l|}{0.62} & \multicolumn{1}{l|}{0.76} & \multicolumn{1}{l|}{0.69} & \multicolumn{1}{l|}{0.62}          \\ \cline{1-1} \cline{3-8} \cline{10-15} 
\multicolumn{1}{|l|}{OC-SVM}                & \multicolumn{1}{l|}{} & \multicolumn{1}{l|}{0.40} & \multicolumn{1}{l|}{0.31} & \multicolumn{1}{l|}{0.57} & \multicolumn{1}{l|}{0.56} & \multicolumn{1}{l|}{0.40} & \multicolumn{1}{l|}{0.42}          & \multicolumn{1}{l|}{} & \multicolumn{1}{l|}{0.66} & \multicolumn{1}{l|}{0.59} & \multicolumn{1}{l|}{0.78} & \multicolumn{1}{l|}{0.84} & \multicolumn{1}{l|}{0.69} & \multicolumn{1}{l|}{0.67}          \\ \cline{1-1} \cline{3-8} \cline{10-15} 
\multicolumn{1}{|l|}{SVDD}                  & \multicolumn{1}{l|}{} & \multicolumn{1}{l|}{0.77} & \multicolumn{1}{l|}{0.89} & \multicolumn{1}{l|}{0.55} & \multicolumn{1}{l|}{0.79} & \multicolumn{1}{l|}{0.83} & \multicolumn{1}{l|}{0.70}          & \multicolumn{1}{l|}{} & \multicolumn{1}{l|}{0.79} & \multicolumn{1}{l|}{0.91} & \multicolumn{1}{l|}{0.58} & \multicolumn{1}{l|}{0.80} & \multicolumn{1}{l|}{0.85} & \multicolumn{1}{l|}{0.72}          \\ \cline{1-1} \cline{3-8} \cline{10-15} 
\end{tabular}
\label{resultsionosphere}
\end{table*}

\begin{table*}[ht]  \footnotesize\setlength{\tabcolsep}{1.5pt}
  \centering 
       \caption{Test results for Handwritten dataset}
\begin{tabular}{lllllllllllllll}
\cline{3-8} \cline{10-15}
                                            & \multicolumn{1}{l|}{} & \multicolumn{6}{c|}{Linear}                                                                                                                                                    & \multicolumn{1}{l|}{} & \multicolumn{6}{c|}{Non-linear}                                                                                                                                         \\ \cline{3-8} \cline{10-15} 
                                            & \multicolumn{1}{l|}{} & \multicolumn{1}{l|}{accu} & \multicolumn{1}{l|}{tpr}  & \multicolumn{1}{l|}{tnr}  & \multicolumn{1}{l|}{pre}  & \multicolumn{1}{l|}{F1}   & \multicolumn{1}{l|}{gm}            & \multicolumn{1}{l|}{} & \multicolumn{1}{l|}{accu} & \multicolumn{1}{l|}{tpr}  & \multicolumn{1}{l|}{tnr}  & \multicolumn{1}{l|}{pre}  & \multicolumn{1}{l|}{F1}   & \multicolumn{1}{l|}{gm}            \\ \cline{3-8} \cline{10-15} 
Proposed method                             &                       &                           &                           &                           &                           &                           &                                    &                       &                           &                           &                           &                           &                           &                                    \\ \cline{1-1} \cline{3-8} \cline{10-15} 
\multicolumn{1}{|l|}{MS-SVDD $\omega_4ds4$} & \multicolumn{1}{l|}{} & \multicolumn{1}{l|}{0.98} & \multicolumn{1}{l|}{0.99} & \multicolumn{1}{l|}{0.98} & \multicolumn{1}{l|}{0.90} & \multicolumn{1}{l|}{0.93} & \multicolumn{1}{l|}{\textbf{0.98}} & \multicolumn{1}{l|}{} & \multicolumn{1}{l|}{0.99} & \multicolumn{1}{l|}{0.99} & \multicolumn{1}{l|}{1.00} & \multicolumn{1}{l|}{0.98} & \multicolumn{1}{l|}{0.98} & \multicolumn{1}{l|}{\textbf{0.99}} \\ \cline{1-1} \cline{3-8} \cline{10-15} 
\multicolumn{1}{|l|}{MS-SVDD $\omega_4ds1$} & \multicolumn{1}{l|}{} & \multicolumn{1}{l|}{0.98} & \multicolumn{1}{l|}{0.90} & \multicolumn{1}{l|}{0.99} & \multicolumn{1}{l|}{0.89} & \multicolumn{1}{l|}{0.89} & \multicolumn{1}{l|}{0.94}          & \multicolumn{1}{l|}{} & \multicolumn{1}{l|}{0.98} & \multicolumn{1}{l|}{0.95} & \multicolumn{1}{l|}{0.99} & \multicolumn{1}{l|}{0.91} & \multicolumn{1}{l|}{0.93} & \multicolumn{1}{l|}{0.97}          \\ \cline{1-1} \cline{3-8} \cline{10-15} 
Concatenated features                       &                       &                           &                           &                           &                           &                           &                                    &                       &                           &                           &                           &                           &                           &                                    \\ \cline{1-1} \cline{3-8} \cline{10-15} 
\multicolumn{1}{|l|}{S-SVDD $\psi_1$}       & \multicolumn{1}{l|}{} & \multicolumn{1}{l|}{0.78} & \multicolumn{1}{l|}{0.92} & \multicolumn{1}{l|}{0.76} & \multicolumn{1}{l|}{0.34} & \multicolumn{1}{l|}{0.49} & \multicolumn{1}{l|}{0.83}          & \multicolumn{1}{l|}{} & \multicolumn{1}{l|}{0.53} & \multicolumn{1}{l|}{0.40} & \multicolumn{1}{l|}{0.54} & \multicolumn{1}{l|}{0.05} & \multicolumn{1}{l|}{0.09} & \multicolumn{1}{l|}{0.14}          \\ \cline{1-1} \cline{3-8} \cline{10-15} 
\multicolumn{1}{|l|}{S-SVDD $\psi_2$}       & \multicolumn{1}{l|}{} & \multicolumn{1}{l|}{0.82} & \multicolumn{1}{l|}{0.88} & \multicolumn{1}{l|}{0.81} & \multicolumn{1}{l|}{0.40} & \multicolumn{1}{l|}{0.54} & \multicolumn{1}{l|}{0.84}          & \multicolumn{1}{l|}{} & \multicolumn{1}{l|}{0.62} & \multicolumn{1}{l|}{0.66} & \multicolumn{1}{l|}{0.61} & \multicolumn{1}{l|}{0.18} & \multicolumn{1}{l|}{0.25} & \multicolumn{1}{l|}{0.44}          \\ \cline{1-1} \cline{3-8} \cline{10-15} 
\multicolumn{1}{|l|}{S-SVDD $\psi_3$}       & \multicolumn{1}{l|}{} & \multicolumn{1}{l|}{0.82} & \multicolumn{1}{l|}{0.97} & \multicolumn{1}{l|}{0.81} & \multicolumn{1}{l|}{0.39} & \multicolumn{1}{l|}{0.55} & \multicolumn{1}{l|}{0.88}          & \multicolumn{1}{l|}{} & \multicolumn{1}{l|}{0.63} & \multicolumn{1}{l|}{0.58} & \multicolumn{1}{l|}{0.64} & \multicolumn{1}{l|}{0.20} & \multicolumn{1}{l|}{0.25} & \multicolumn{1}{l|}{0.30}          \\ \cline{1-1} \cline{3-8} \cline{10-15} 
\multicolumn{1}{|l|}{S-SVDD $\psi_4$}       & \multicolumn{1}{l|}{} & \multicolumn{1}{l|}{0.84} & \multicolumn{1}{l|}{0.92} & \multicolumn{1}{l|}{0.83} & \multicolumn{1}{l|}{0.42} & \multicolumn{1}{l|}{0.56} & \multicolumn{1}{l|}{0.87}          & \multicolumn{1}{l|}{} & \multicolumn{1}{l|}{0.71} & \multicolumn{1}{l|}{0.39} & \multicolumn{1}{l|}{0.75} & \multicolumn{1}{l|}{0.08} & \multicolumn{1}{l|}{0.13} & \multicolumn{1}{l|}{0.17}          \\ \cline{1-1} \cline{3-8} \cline{10-15} 
\multicolumn{1}{|l|}{OC-SVM}                & \multicolumn{1}{l|}{} & \multicolumn{1}{l|}{0.50} & \multicolumn{1}{l|}{0.51} & \multicolumn{1}{l|}{0.50} & \multicolumn{1}{l|}{0.12} & \multicolumn{1}{l|}{0.19} & \multicolumn{1}{l|}{0.49}          & \multicolumn{1}{l|}{} & \multicolumn{1}{l|}{0.95} & \multicolumn{1}{l|}{0.51} & \multicolumn{1}{l|}{1.00} & \multicolumn{1}{l|}{1.00} & \multicolumn{1}{l|}{0.68} & \multicolumn{1}{l|}{0.71}          \\ \cline{1-1} \cline{3-8} \cline{10-15} 
\multicolumn{1}{|l|}{SVDD}                  & \multicolumn{1}{l|}{} & \multicolumn{1}{l|}{0.95} & \multicolumn{1}{l|}{0.93} & \multicolumn{1}{l|}{0.95} & \multicolumn{1}{l|}{0.69} & \multicolumn{1}{l|}{0.79} & \multicolumn{1}{l|}{0.94}          & \multicolumn{1}{l|}{} & \multicolumn{1}{l|}{0.95} & \multicolumn{1}{l|}{0.92} & \multicolumn{1}{l|}{0.96} & \multicolumn{1}{l|}{0.74} & \multicolumn{1}{l|}{0.81} & \multicolumn{1}{l|}{0.94}          \\ \cline{1-1} \cline{3-8} \cline{10-15} 
ZER                                         &                       &                           &                           &                           &                           &                           &                                    &                       &                           &                           &                           &                           &                           &                                    \\ \cline{1-1} \cline{3-8} \cline{10-15} 
\multicolumn{1}{|l|}{S-SVDD $\psi_1$}       & \multicolumn{1}{l|}{} & \multicolumn{1}{l|}{0.55} & \multicolumn{1}{l|}{0.92} & \multicolumn{1}{l|}{0.51} & \multicolumn{1}{l|}{0.18} & \multicolumn{1}{l|}{0.30} & \multicolumn{1}{l|}{0.68}          & \multicolumn{1}{l|}{} & \multicolumn{1}{l|}{0.59} & \multicolumn{1}{l|}{0.41} & \multicolumn{1}{l|}{0.61} & \multicolumn{1}{l|}{0.06} & \multicolumn{1}{l|}{0.10} & \multicolumn{1}{l|}{0.24}          \\ \cline{1-1} \cline{3-8} \cline{10-15} 
\multicolumn{1}{|l|}{S-SVDD $\psi_2$}       & \multicolumn{1}{l|}{} & \multicolumn{1}{l|}{0.52} & \multicolumn{1}{l|}{0.88} & \multicolumn{1}{l|}{0.48} & \multicolumn{1}{l|}{0.17} & \multicolumn{1}{l|}{0.28} & \multicolumn{1}{l|}{0.64}          & \multicolumn{1}{l|}{} & \multicolumn{1}{l|}{0.62} & \multicolumn{1}{l|}{0.78} & \multicolumn{1}{l|}{0.60} & \multicolumn{1}{l|}{0.17} & \multicolumn{1}{l|}{0.27} & \multicolumn{1}{l|}{0.48}          \\ \cline{1-1} \cline{3-8} \cline{10-15} 
\multicolumn{1}{|l|}{S-SVDD $\psi_3$}       & \multicolumn{1}{l|}{} & \multicolumn{1}{l|}{0.50} & \multicolumn{1}{l|}{0.96} & \multicolumn{1}{l|}{0.45} & \multicolumn{1}{l|}{0.19} & \multicolumn{1}{l|}{0.31} & \multicolumn{1}{l|}{0.63}          & \multicolumn{1}{l|}{} & \multicolumn{1}{l|}{0.57} & \multicolumn{1}{l|}{0.61} & \multicolumn{1}{l|}{0.57} & \multicolumn{1}{l|}{0.31} & \multicolumn{1}{l|}{0.20} & \multicolumn{1}{l|}{0.37}          \\ \cline{1-1} \cline{3-8} \cline{10-15} 
\multicolumn{1}{|l|}{S-SVDD $\psi_4$}       & \multicolumn{1}{l|}{} & \multicolumn{1}{l|}{0.64} & \multicolumn{1}{l|}{0.90} & \multicolumn{1}{l|}{0.61} & \multicolumn{1}{l|}{0.21} & \multicolumn{1}{l|}{0.34} & \multicolumn{1}{l|}{0.74}          & \multicolumn{1}{l|}{} & \multicolumn{1}{l|}{0.55} & \multicolumn{1}{l|}{0.60} & \multicolumn{1}{l|}{0.54} & \multicolumn{1}{l|}{0.09} & \multicolumn{1}{l|}{0.15} & \multicolumn{1}{l|}{0.24}          \\ \cline{1-1} \cline{3-8} \cline{10-15} 
\multicolumn{1}{|l|}{OC-SVM}                & \multicolumn{1}{l|}{} & \multicolumn{1}{l|}{0.43} & \multicolumn{1}{l|}{0.42} & \multicolumn{1}{l|}{0.43} & \multicolumn{1}{l|}{0.09} & \multicolumn{1}{l|}{0.14} & \multicolumn{1}{l|}{0.41}          & \multicolumn{1}{l|}{} & \multicolumn{1}{l|}{0.95} & \multicolumn{1}{l|}{0.52} & \multicolumn{1}{l|}{1.00} & \multicolumn{1}{l|}{0.93} & \multicolumn{1}{l|}{0.67} & \multicolumn{1}{l|}{0.72}          \\ \cline{1-1} \cline{3-8} \cline{10-15} 
\multicolumn{1}{|l|}{SVDD}                  & \multicolumn{1}{l|}{} & \multicolumn{1}{l|}{0.88} & \multicolumn{1}{l|}{0.90} & \multicolumn{1}{l|}{0.88} & \multicolumn{1}{l|}{0.47} & \multicolumn{1}{l|}{0.61} & \multicolumn{1}{l|}{0.89}          & \multicolumn{1}{l|}{} & \multicolumn{1}{l|}{0.92} & \multicolumn{1}{l|}{0.88} & \multicolumn{1}{l|}{0.92} & \multicolumn{1}{l|}{0.56} & \multicolumn{1}{l|}{0.68} & \multicolumn{1}{l|}{0.90}          \\ \cline{1-1} \cline{3-8} \cline{10-15} 
MOR                                         &                       &                           &                           &                           &                           &                           &                                    &                       &                           &                           &                           &                           &                           &                                    \\ \cline{1-1} \cline{3-8} \cline{10-15} 
\multicolumn{1}{|l|}{S-SVDD $\psi_1$}       & \multicolumn{1}{l|}{} & \multicolumn{1}{l|}{0.84} & \multicolumn{1}{l|}{0.99} & \multicolumn{1}{l|}{0.82} & \multicolumn{1}{l|}{0.48} & \multicolumn{1}{l|}{0.61} & \multicolumn{1}{l|}{0.90}          & \multicolumn{1}{l|}{} & \multicolumn{1}{l|}{0.84} & \multicolumn{1}{l|}{0.01} & \multicolumn{1}{l|}{0.93} & \multicolumn{1}{l|}{0.00} & \multicolumn{1}{l|}{0.00} & \multicolumn{1}{l|}{0.03}          \\ \cline{1-1} \cline{3-8} \cline{10-15} 
\multicolumn{1}{|l|}{S-SVDD $\psi_2$}       & \multicolumn{1}{l|}{} & \multicolumn{1}{l|}{0.92} & \multicolumn{1}{l|}{0.99} & \multicolumn{1}{l|}{0.91} & \multicolumn{1}{l|}{0.66} & \multicolumn{1}{l|}{0.76} & \multicolumn{1}{l|}{0.95}          & \multicolumn{1}{l|}{} & \multicolumn{1}{l|}{0.58} & \multicolumn{1}{l|}{0.44} & \multicolumn{1}{l|}{0.60} & \multicolumn{1}{l|}{0.43} & \multicolumn{1}{l|}{0.22} & \multicolumn{1}{l|}{0.20}          \\ \cline{1-1} \cline{3-8} \cline{10-15} 
\multicolumn{1}{|l|}{S-SVDD $\psi_3$}       & \multicolumn{1}{l|}{} & \multicolumn{1}{l|}{0.86} & \multicolumn{1}{l|}{0.99} & \multicolumn{1}{l|}{0.84} & \multicolumn{1}{l|}{0.52} & \multicolumn{1}{l|}{0.64} & \multicolumn{1}{l|}{0.91}          & \multicolumn{1}{l|}{} & \multicolumn{1}{l|}{0.61} & \multicolumn{1}{l|}{0.70} & \multicolumn{1}{l|}{0.60} & \multicolumn{1}{l|}{0.44} & \multicolumn{1}{l|}{0.42} & \multicolumn{1}{l|}{0.36}          \\ \cline{1-1} \cline{3-8} \cline{10-15} 
\multicolumn{1}{|l|}{S-SVDD $\psi_4$}       & \multicolumn{1}{l|}{} & \multicolumn{1}{l|}{0.84} & \multicolumn{1}{l|}{0.99} & \multicolumn{1}{l|}{0.82} & \multicolumn{1}{l|}{0.48} & \multicolumn{1}{l|}{0.61} & \multicolumn{1}{l|}{0.90}          & \multicolumn{1}{l|}{} & \multicolumn{1}{l|}{0.25} & \multicolumn{1}{l|}{0.67} & \multicolumn{1}{l|}{0.20} & \multicolumn{1}{l|}{0.27} & \multicolumn{1}{l|}{0.14} & \multicolumn{1}{l|}{0.04}          \\ \cline{1-1} \cline{3-8} \cline{10-15} 
\multicolumn{1}{|l|}{OC-SVM}                & \multicolumn{1}{l|}{} & \multicolumn{1}{l|}{0.54} & \multicolumn{1}{l|}{0.45} & \multicolumn{1}{l|}{0.55} & \multicolumn{1}{l|}{0.13} & \multicolumn{1}{l|}{0.18} & \multicolumn{1}{l|}{0.39}          & \multicolumn{1}{l|}{} & \multicolumn{1}{l|}{0.99} & \multicolumn{1}{l|}{0.87} & \multicolumn{1}{l|}{1.00} & \multicolumn{1}{l|}{1.00} & \multicolumn{1}{l|}{0.93} & \multicolumn{1}{l|}{0.93}          \\ \cline{1-1} \cline{3-8} \cline{10-15} 
\multicolumn{1}{|l|}{SVDD}                  & \multicolumn{1}{l|}{} & \multicolumn{1}{l|}{0.93} & \multicolumn{1}{l|}{0.91} & \multicolumn{1}{l|}{0.93} & \multicolumn{1}{l|}{0.75} & \multicolumn{1}{l|}{0.78} & \multicolumn{1}{l|}{0.92}          & \multicolumn{1}{l|}{} & \multicolumn{1}{l|}{0.99} & \multicolumn{1}{l|}{0.96} & \multicolumn{1}{l|}{1.00} & \multicolumn{1}{l|}{1.00} & \multicolumn{1}{l|}{0.98} & \multicolumn{1}{l|}{0.98}          \\ \cline{1-1} \cline{3-8} \cline{10-15} 
\end{tabular}
\label{resultshandwritten}
\end{table*}

\begin{table*}[ht]  \footnotesize\setlength{\tabcolsep}{1.5pt}
  \centering 
       \caption{Test results for SPECTF heart dataset}
\begin{tabular}{lllllllllllllll}
\cline{3-8} \cline{10-15}
                                             & \multicolumn{1}{l|}{} & \multicolumn{6}{c|}{Linear}                                                                                                                                                    & \multicolumn{1}{l|}{} & \multicolumn{6}{c|}{Non-linear}                                                                                                                                                \\ \cline{3-8} \cline{10-15} 
                                             & \multicolumn{1}{l|}{} & \multicolumn{1}{l|}{accu} & \multicolumn{1}{l|}{tpr}  & \multicolumn{1}{l|}{tnr}  & \multicolumn{1}{l|}{pre}  & \multicolumn{1}{l|}{F1}   & \multicolumn{1}{l|}{gm}            & \multicolumn{1}{l|}{} & \multicolumn{1}{l|}{accu} & \multicolumn{1}{l|}{tpr}  & \multicolumn{1}{l|}{tnr}  & \multicolumn{1}{l|}{pre}  & \multicolumn{1}{l|}{F1}   & \multicolumn{1}{l|}{gm}            \\ \cline{3-8} \cline{10-15} 
Proposed method                              &                       &                           &                           &                           &                           &                           &                                    &                       &                           &                           &                           &                           &                           &                                    \\ \cline{1-1} \cline{3-8} \cline{10-15} 
\multicolumn{1}{|l|}{MS-SVDD $\omega_0ds1$} & \multicolumn{1}{l|}{} & \multicolumn{1}{l|}{0.78} & \multicolumn{1}{l|}{0.80} & \multicolumn{1}{l|}{0.78} & \multicolumn{1}{l|}{0.24} & \multicolumn{1}{l|}{0.37} & \multicolumn{1}{l|}{\textbf{0.79}} & \multicolumn{1}{l|}{} & \multicolumn{1}{l|}{0.55} & \multicolumn{1}{l|}{0.60} & \multicolumn{1}{l|}{0.55} & \multicolumn{1}{l|}{0.10} & \multicolumn{1}{l|}{0.18} & \multicolumn{1}{l|}{0.57}          \\ \cline{1-1} \cline{3-8} \cline{10-15} 
\multicolumn{1}{|l|}{MS-SVDD $\omega_2ds1$} & \multicolumn{1}{l|}{} & \multicolumn{1}{l|}{0.78} & \multicolumn{1}{l|}{0.80} & \multicolumn{1}{l|}{0.77} & \multicolumn{1}{l|}{0.24} & \multicolumn{1}{l|}{0.36} & \multicolumn{1}{l|}{\textbf{0.79}} & \multicolumn{1}{l|}{} & \multicolumn{1}{l|}{0.80} & \multicolumn{1}{l|}{0.73} & \multicolumn{1}{l|}{0.80} & \multicolumn{1}{l|}{0.24} & \multicolumn{1}{l|}{0.37} & \multicolumn{1}{l|}{\textbf{0.77}} \\ \cline{1-1} \cline{3-8} \cline{10-15} 
Concatenated features                        &                       &                           &                           &                           &                           &                           &                                    &                       &                           &                           &                           &                           &                           &                                    \\ \cline{1-1} \cline{3-8} \cline{10-15} 
\multicolumn{1}{|l|}{S-SVDD $\psi_1$}        & \multicolumn{1}{l|}{} & \multicolumn{1}{l|}{0.71} & \multicolumn{1}{l|}{0.53} & \multicolumn{1}{l|}{0.73} & \multicolumn{1}{l|}{0.15} & \multicolumn{1}{l|}{0.23} & \multicolumn{1}{l|}{0.62}          & \multicolumn{1}{l|}{} & \multicolumn{1}{l|}{0.77} & \multicolumn{1}{l|}{0.60} & \multicolumn{1}{l|}{0.78} & \multicolumn{1}{l|}{0.20} & \multicolumn{1}{l|}{0.30} & \multicolumn{1}{l|}{0.69}          \\ \cline{1-1} \cline{3-8} \cline{10-15} 
\multicolumn{1}{|l|}{S-SVDD $\psi_2$}        & \multicolumn{1}{l|}{} & \multicolumn{1}{l|}{0.69} & \multicolumn{1}{l|}{0.87} & \multicolumn{1}{l|}{0.67} & \multicolumn{1}{l|}{0.19} & \multicolumn{1}{l|}{0.31} & \multicolumn{1}{l|}{0.76}          & \multicolumn{1}{l|}{} & \multicolumn{1}{l|}{0.77} & \multicolumn{1}{l|}{0.60} & \multicolumn{1}{l|}{0.78} & \multicolumn{1}{l|}{0.20} & \multicolumn{1}{l|}{0.30} & \multicolumn{1}{l|}{0.69}          \\ \cline{1-1} \cline{3-8} \cline{10-15} 
\multicolumn{1}{|l|}{S-SVDD $\psi_3$}        & \multicolumn{1}{l|}{} & \multicolumn{1}{l|}{0.66} & \multicolumn{1}{l|}{0.93} & \multicolumn{1}{l|}{0.64} & \multicolumn{1}{l|}{0.18} & \multicolumn{1}{l|}{0.31} & \multicolumn{1}{l|}{0.77}          & \multicolumn{1}{l|}{} & \multicolumn{1}{l|}{0.77} & \multicolumn{1}{l|}{0.60} & \multicolumn{1}{l|}{0.78} & \multicolumn{1}{l|}{0.20} & \multicolumn{1}{l|}{0.30} & \multicolumn{1}{l|}{0.69}          \\ \cline{1-1} \cline{3-8} \cline{10-15} 
\multicolumn{1}{|l|}{S-SVDD $\psi_4$}        & \multicolumn{1}{l|}{} & \multicolumn{1}{l|}{0.56} & \multicolumn{1}{l|}{0.67} & \multicolumn{1}{l|}{0.55} & \multicolumn{1}{l|}{0.11} & \multicolumn{1}{l|}{0.19} & \multicolumn{1}{l|}{0.60}          & \multicolumn{1}{l|}{} & \multicolumn{1}{l|}{0.77} & \multicolumn{1}{l|}{0.60} & \multicolumn{1}{l|}{0.78} & \multicolumn{1}{l|}{0.20} & \multicolumn{1}{l|}{0.30} & \multicolumn{1}{l|}{0.69}          \\ \cline{1-1} \cline{3-8} \cline{10-15} 
\multicolumn{1}{|l|}{OC-SVM}                 & \multicolumn{1}{l|}{} & \multicolumn{1}{l|}{0.86} & \multicolumn{1}{l|}{0.27} & \multicolumn{1}{l|}{0.91} & \multicolumn{1}{l|}{0.20} & \multicolumn{1}{l|}{0.23} & \multicolumn{1}{l|}{0.49}          & \multicolumn{1}{l|}{} & \multicolumn{1}{l|}{0.76} & \multicolumn{1}{l|}{0.73} & \multicolumn{1}{l|}{0.77} & \multicolumn{1}{l|}{0.22} & \multicolumn{1}{l|}{0.33} & \multicolumn{1}{l|}{0.75}          \\ \cline{1-1} \cline{3-8} \cline{10-15} 
\multicolumn{1}{|l|}{SVDD}                   & \multicolumn{1}{l|}{} & \multicolumn{1}{l|}{0.69} & \multicolumn{1}{l|}{0.73} & \multicolumn{1}{l|}{0.69} & \multicolumn{1}{l|}{0.17} & \multicolumn{1}{l|}{0.28} & \multicolumn{1}{l|}{0.71}          & \multicolumn{1}{l|}{} & \multicolumn{1}{l|}{0.75} & \multicolumn{1}{l|}{0.67} & \multicolumn{1}{l|}{0.76} & \multicolumn{1}{l|}{0.19} & \multicolumn{1}{l|}{0.30} & \multicolumn{1}{l|}{0.71}          \\ \cline{1-1} \cline{3-8} \cline{10-15} 
Rest Mode                                    &                       &                           &                           &                           &                           &                           &                                    &                       &                           &                           &                           &                           &                           &                                    \\ \cline{1-1} \cline{3-8} \cline{10-15} 
\multicolumn{1}{|l|}{S-SVDD $\psi_1$}        & \multicolumn{1}{l|}{} & \multicolumn{1}{l|}{0.50} & \multicolumn{1}{l|}{0.73} & \multicolumn{1}{l|}{0.48} & \multicolumn{1}{l|}{0.11} & \multicolumn{1}{l|}{0.19} & \multicolumn{1}{l|}{0.59}          & \multicolumn{1}{l|}{} & \multicolumn{1}{l|}{0.46} & \multicolumn{1}{l|}{0.87} & \multicolumn{1}{l|}{0.42} & \multicolumn{1}{l|}{0.12} & \multicolumn{1}{l|}{0.20} & \multicolumn{1}{l|}{0.61}          \\ \cline{1-1} \cline{3-8} \cline{10-15} 
\multicolumn{1}{|l|}{S-SVDD $\psi_2$}        & \multicolumn{1}{l|}{} & \multicolumn{1}{l|}{0.58} & \multicolumn{1}{l|}{0.87} & \multicolumn{1}{l|}{0.55} & \multicolumn{1}{l|}{0.14} & \multicolumn{1}{l|}{0.25} & \multicolumn{1}{l|}{0.69}          & \multicolumn{1}{l|}{} & \multicolumn{1}{l|}{0.77} & \multicolumn{1}{l|}{0.53} & \multicolumn{1}{l|}{0.79} & \multicolumn{1}{l|}{0.18} & \multicolumn{1}{l|}{0.27} & \multicolumn{1}{l|}{0.65}          \\ \cline{1-1} \cline{3-8} \cline{10-15} 
\multicolumn{1}{|l|}{S-SVDD $\psi_3$}        & \multicolumn{1}{l|}{} & \multicolumn{1}{l|}{0.40} & \multicolumn{1}{l|}{0.80} & \multicolumn{1}{l|}{0.37} & \multicolumn{1}{l|}{0.10} & \multicolumn{1}{l|}{0.18} & \multicolumn{1}{l|}{0.54}          & \multicolumn{1}{l|}{} & \multicolumn{1}{l|}{0.79} & \multicolumn{1}{l|}{0.47} & \multicolumn{1}{l|}{0.81} & \multicolumn{1}{l|}{0.18} & \multicolumn{1}{l|}{0.26} & \multicolumn{1}{l|}{0.62}          \\ \cline{1-1} \cline{3-8} \cline{10-15} 
\multicolumn{1}{|l|}{S-SVDD $\psi_4$}        & \multicolumn{1}{l|}{} & \multicolumn{1}{l|}{0.38} & \multicolumn{1}{l|}{0.87} & \multicolumn{1}{l|}{0.34} & \multicolumn{1}{l|}{0.10} & \multicolumn{1}{l|}{0.18} & \multicolumn{1}{l|}{0.54}          & \multicolumn{1}{l|}{} & \multicolumn{1}{l|}{0.60} & \multicolumn{1}{l|}{0.87} & \multicolumn{1}{l|}{0.58} & \multicolumn{1}{l|}{0.15} & \multicolumn{1}{l|}{0.26} & \multicolumn{1}{l|}{0.71}          \\ \cline{1-1} \cline{3-8} \cline{10-15} 
\multicolumn{1}{|l|}{OC-SVM}                 & \multicolumn{1}{l|}{} & \multicolumn{1}{l|}{0.76} & \multicolumn{1}{l|}{0.60} & \multicolumn{1}{l|}{0.77} & \multicolumn{1}{l|}{0.19} & \multicolumn{1}{l|}{0.29} & \multicolumn{1}{l|}{0.68}          & \multicolumn{1}{l|}{} & \multicolumn{1}{l|}{0.61} & \multicolumn{1}{l|}{0.80} & \multicolumn{1}{l|}{0.60} & \multicolumn{1}{l|}{0.15} & \multicolumn{1}{l|}{0.25} & \multicolumn{1}{l|}{0.69}          \\ \cline{1-1} \cline{3-8} \cline{10-15} 
\multicolumn{1}{|l|}{SVDD}                   & \multicolumn{1}{l|}{} & \multicolumn{1}{l|}{0.59} & \multicolumn{1}{l|}{0.73} & \multicolumn{1}{l|}{0.58} & \multicolumn{1}{l|}{0.13} & \multicolumn{1}{l|}{0.22} & \multicolumn{1}{l|}{0.65}          & \multicolumn{1}{l|}{} & \multicolumn{1}{l|}{0.59} & \multicolumn{1}{l|}{0.73} & \multicolumn{1}{l|}{0.58} & \multicolumn{1}{l|}{0.13} & \multicolumn{1}{l|}{0.22} & \multicolumn{1}{l|}{0.65}          \\ \cline{1-1} \cline{3-8} \cline{10-15} 
Stress Mode                                  &                       &                           &                           &                           &                           &                           &                                    &                       &                           &                           &                           &                           &                           &                                    \\ \cline{1-1} \cline{3-8} \cline{10-15} 
\multicolumn{1}{|l|}{S-SVDD $\psi_1$}        & \multicolumn{1}{l|}{} & \multicolumn{1}{l|}{0.53} & \multicolumn{1}{l|}{0.47} & \multicolumn{1}{l|}{0.53} & \multicolumn{1}{l|}{0.08} & \multicolumn{1}{l|}{0.14} & \multicolumn{1}{l|}{0.50}          & \multicolumn{1}{l|}{} & \multicolumn{1}{l|}{0.68} & \multicolumn{1}{l|}{0.73} & \multicolumn{1}{l|}{0.67} & \multicolumn{1}{l|}{0.16} & \multicolumn{1}{l|}{0.27} & \multicolumn{1}{l|}{0.70}          \\ \cline{1-1} \cline{3-8} \cline{10-15} 
\multicolumn{1}{|l|}{S-SVDD $\psi_2$}        & \multicolumn{1}{l|}{} & \multicolumn{1}{l|}{0.65} & \multicolumn{1}{l|}{0.80} & \multicolumn{1}{l|}{0.63} & \multicolumn{1}{l|}{0.16} & \multicolumn{1}{l|}{0.27} & \multicolumn{1}{l|}{0.71}          & \multicolumn{1}{l|}{} & \multicolumn{1}{l|}{0.75} & \multicolumn{1}{l|}{0.53} & \multicolumn{1}{l|}{0.77} & \multicolumn{1}{l|}{0.17} & \multicolumn{1}{l|}{0.26} & \multicolumn{1}{l|}{0.64}          \\ \cline{1-1} \cline{3-8} \cline{10-15} 
\multicolumn{1}{|l|}{S-SVDD $\psi_3$}        & \multicolumn{1}{l|}{} & \multicolumn{1}{l|}{0.73} & \multicolumn{1}{l|}{0.67} & \multicolumn{1}{l|}{0.73} & \multicolumn{1}{l|}{0.18} & \multicolumn{1}{l|}{0.28} & \multicolumn{1}{l|}{0.70}          & \multicolumn{1}{l|}{} & \multicolumn{1}{l|}{0.70} & \multicolumn{1}{l|}{0.73} & \multicolumn{1}{l|}{0.70} & \multicolumn{1}{l|}{0.17} & \multicolumn{1}{l|}{0.28} & \multicolumn{1}{l|}{0.72}          \\ \cline{1-1} \cline{3-8} \cline{10-15} 
\multicolumn{1}{|l|}{S-SVDD $\psi_4$}        & \multicolumn{1}{l|}{} & \multicolumn{1}{l|}{0.55} & \multicolumn{1}{l|}{0.93} & \multicolumn{1}{l|}{0.52} & \multicolumn{1}{l|}{0.14} & \multicolumn{1}{l|}{0.25} & \multicolumn{1}{l|}{0.69}          & \multicolumn{1}{l|}{} & \multicolumn{1}{l|}{0.75} & \multicolumn{1}{l|}{0.53} & \multicolumn{1}{l|}{0.77} & \multicolumn{1}{l|}{0.17} & \multicolumn{1}{l|}{0.26} & \multicolumn{1}{l|}{0.64}          \\ \cline{1-1} \cline{3-8} \cline{10-15} 
\multicolumn{1}{|l|}{OC-SVM}                 & \multicolumn{1}{l|}{} & \multicolumn{1}{l|}{0.86} & \multicolumn{1}{l|}{0.20} & \multicolumn{1}{l|}{0.91} & \multicolumn{1}{l|}{0.17} & \multicolumn{1}{l|}{0.18} & \multicolumn{1}{l|}{0.43}          & \multicolumn{1}{l|}{} & \multicolumn{1}{l|}{0.73} & \multicolumn{1}{l|}{0.60} & \multicolumn{1}{l|}{0.74} & \multicolumn{1}{l|}{0.17} & \multicolumn{1}{l|}{0.26} & \multicolumn{1}{l|}{0.67}          \\ \cline{1-1} \cline{3-8} \cline{10-15} 
\multicolumn{1}{|l|}{SVDD}                   & \multicolumn{1}{l|}{} & \multicolumn{1}{l|}{0.76} & \multicolumn{1}{l|}{0.60} & \multicolumn{1}{l|}{0.77} & \multicolumn{1}{l|}{0.19} & \multicolumn{1}{l|}{0.29} & \multicolumn{1}{l|}{0.68}          & \multicolumn{1}{l|}{} & \multicolumn{1}{l|}{0.78} & \multicolumn{1}{l|}{0.53} & \multicolumn{1}{l|}{0.80} & \multicolumn{1}{l|}{0.19} & \multicolumn{1}{l|}{0.28} & \multicolumn{1}{l|}{0.65}          \\ \cline{1-1} \cline{3-8} \cline{10-15} 
\end{tabular}
\label{resultsheart}
\end{table*}

In Tables \ref{resultsrobot}, \ref{resultscaltech}, \ref{resultsionosphere}, and \ref{resultshandwritten}, we report the average of different evaluation metrics over the five data splits for Robot Execution Failures dataset, Caltech-7 dataset, Ionosphere dataset, and Handwritten dataset, respectively, for both linear and non-linear versions of the applied methods. In Table \ref{resultsheart}, we report the results on the test set for the SPECTF heart dataset. In these tables, we only show the best performing versions of the proposed method, along with all competing methods. We compare our results with OC-SVM \cite{scholkopfu1999sv}, SVDD \cite{tax2004support}, and S-SVDD \cite{sohrab2018subspace}. In S-SVDD, different regularization terms ($\psi$'s) were proposed and, hence, we compare MS-SVDD with all proposed regularization terms of S-SVDD. We use kernel version of the competing methods for non-linear comparisons. In these tables, we report the best performing non-linear version of MS-SVDD for corresponding datasets. To analyze the different regularization terms and decision strategies for the proposed method, we also report the exhaustive results obtained by different settings in the supplementary material in Tables 1-5. The best results in terms of \textit{gm} are reported as in bold formatting.

For the Robot Execution Failures dataset (Table \ref{resultsrobot}), our proposed method outperforms all the competing methods in the linear case. The results achieved by the linear version of the proposed MS-SVDD method are overall best also compared to the non-linear methods. Table \ref{resultsrobot} shows that using decision strategy 3 with constraint $\omega_2$ (all support vectors and outliers from the corresponding modality considered for the update of the corresponding $\mathbf{Q}_m$) yields the best overall results for the robot dataset. In the non-linear case, the best performance for the proposed method is achieved by using the kernel trick with either constraint type $\omega_2$ or $\omega_5$, both with decision strategy 3. 

We also notice that the first modality (force measurements) is vital in taking the final decision as in both linear and non-linear cases, the best results are obtained when the decision is taken based on the first modality (decision strategy 3). The importance of the first modality is also evident from the results of the competing methods as the best results are obtained when using force measurements only. The results on the concatenated features are slightly worse, and the results using the torque measurements are clearly worse. Nevertheless, the proposed multimodal approach has managed to boost the results by combining information from both modalities.

For the Caltech-7 dataset, in the linear case, MS-SVDD performs better than all other methods with a single modality. Overall, only SVDD using concatenated features outperforms MS-SVDD and the margin is small. In the non-linear case, MS-SVDD obtains the best results along with SVDD. In terms of \textit{tpr}, MS-SVDD outperforms all the other methods in the non-linear case while maintaining reasonably good \textit{tnr}. We also notice that both modalities are vital in taking the final decision as the best performance of MS-SVDD is obtained by decision strategy 1 (AND gate).

For Ionosphere dataset, only SVDD applied on concatenated features or the first modality outperforms MS-SVDD in terms of \textit{gm}. Nevertheless, the performance of MS-SVDD is competitive as shown also by the top results obtained by the other performance metrics such as  \textit{F1} measure. In case of MS-SVDD, the second modality (Complex) is found to be more vital for taking the final decision.  

For the Handwritten dataset, MS-SVDD outperforms all competing methods in both linear and non-linear cases. It is noticed that decision strategy 4 yields the best results in both linear and non-linear cases for MS-SVDD, i.e., MOR features are more vital than ZER features.

For SPECTF heart dataset, in both linear and non-linear cases, the best results are achieved by MS-SVDD. We note that $\omega_0$ (no constraint used) and $\omega_2$, where all support vectors and outliers are used to describe the class variance for the update of the corresponding $\mathbf{Q}_m$ in decision strategy 1 yield the best overall results.

 \begin{figure*}[ht]
	\centering
	\includegraphics[width=\textwidth]{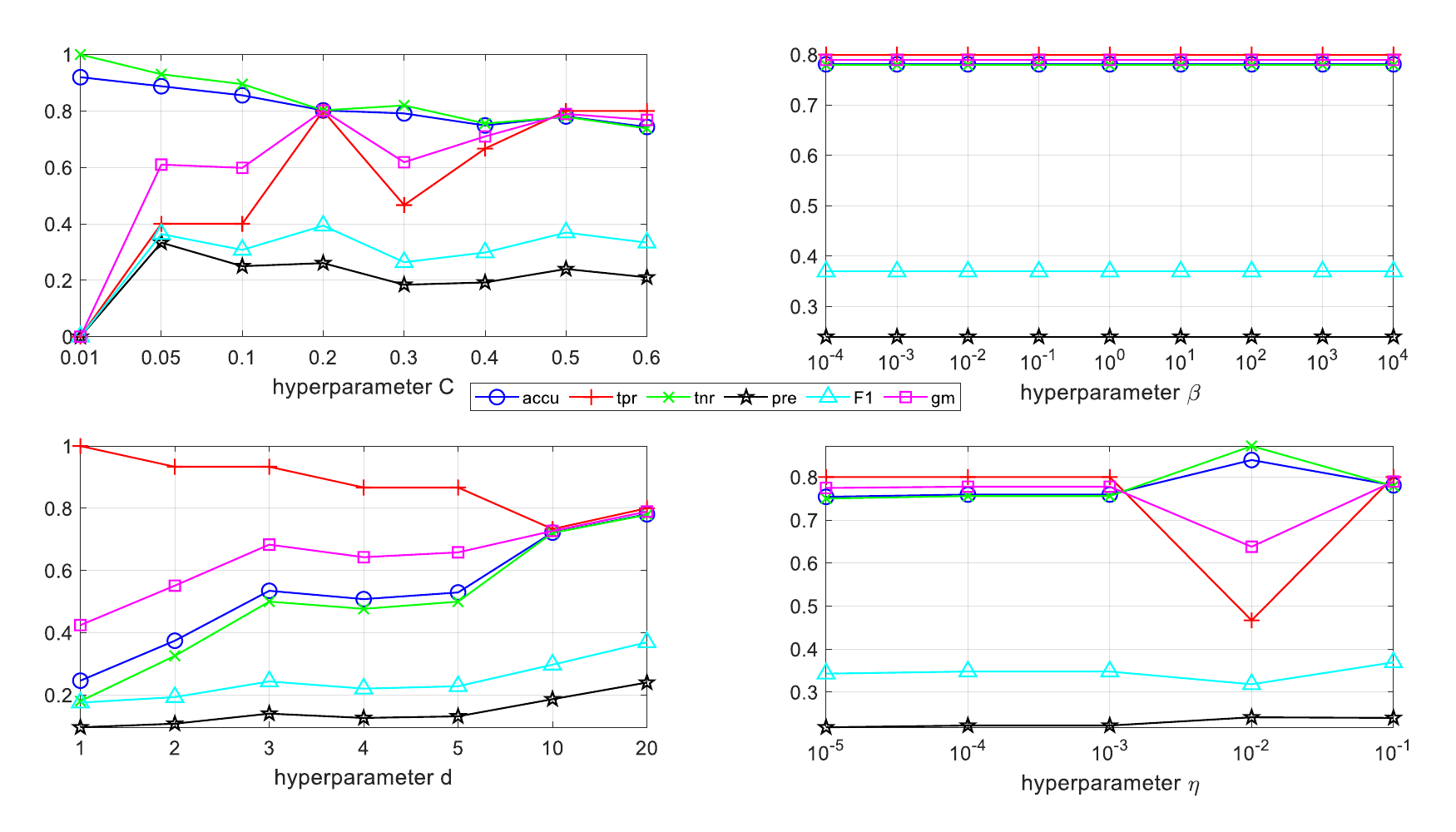}
	\caption{Hyperparameters sensitivity analysis for $\omega_0ds1$}
	\label{w0d1}
\end{figure*}

We compare the results for different variant of MS-SVDD in Tables 1-5 of the supplementary material. Overall in all datasets, NPT is found to be more robust than the kernel version. Linear MS-SVDD is found to perform best over 2 datasets, similar to the NPT version, which performs best on two datasets as well. The kernel MS-SVDD performs best on one out of five datasets as compared to linear and NPT version of MS-SVDD. All the relevant codes (implementation) for the proposed method are available online at \cite{mssvddcode}.

We also carried out a sensitivity analysis of different hyperparameters for linear MS-SVDD over SPECTF heart dataset. To analyze the sensitivity of MS-SVDD for each hyperparameter, we fix the other hyperparameters to their optimal values and record the performance with all the considered hyperparameter values. Figure \ref{w0d1} shows as an example the results for decision strategy 1 without any constraint. For the other decision strategies and constraints, we show the results in Figures 1-27 in the supplementary material. We note a trend of increase in \textit{tpr} and decrease in \textit{tnr} with the increase of value for hyperparameter $C$. We also noticed that the performance of trained models are relatively less sensitive to the hyperparameter $\beta$ as compared to other hyperparameters. For hyperparameter $d$, initially, there is a noticeable rise in the performance of the trained model; however, after certain value, the change seems to be very small. For hyperparameter $\eta$, we notice that precision and \textit{F1} measure stay stable with changing its value.

We also report the numerical training and testing time (in milliseconds) in the supplementary material (Tables 6-10) for all methods over all datasets used in the experiments. In the majority of cases, the proposed method has a higher computational cost than the competing methods, but generally, the difference is in the fractions of a second, which is negligible for datasets used in this work. It is also evident from the numerical results that the time complexity of the proposed method is higher mainly in the training phase, while in the testing phase the difference is negligible. This is as expected based on the complexity analysis in Section \ref{complexityanalysis}.

\section{Conclusion}\label{conclusion}
In this paper, a new multimodal one-class classification method is proposed. The proposed method iteratively transforms data from all the modalities to a new shared subspace optimized for data description in multimodal one-class classification tasks. We derived linear and two different non-linear versions along with a selection of different regularization terms. According to the best of our knowledge, this is the first work in the field of subspace learning for multimodal one-class classification. We conducted experiments comparing the different versions of MS-SVDD and performed comparisons against other one-class classification methods using either concatenated representations or a single modality at a time. 

In most cases, linear and NPT version of MS-SVDD outperformed all the competing methods in our experiments. NPT turned out to be more stable than the kernel version. We noticed that the optimal decision strategy depends on the usefulness of different modalities. If a particular modality is more informative than other(s), then it is useful to use that particular modality for making the final decision. Nevertheless, MS-SVDD can improve the results as compared to using a single modality only. If the modalities are more balanced, the AND gate strategy seems to perform better.

MS-SVDD can be interpreted and used in many ways for different one-class multimodal problems. It can be used for anomaly detection and detection of a specific class such as speaker verification and face recognition. In the future, we intend to try different kernels and model-based decision strategies for the proposed method. We also intend to propose changes in the boundary shape (other than spherical) for enclosing the target data in subspace. There is also room for research in other one-class classification techniques for multimodal subspace learning.

\section{Acknowledgement}\label{Acknowledgement}
This work was supported by the NSF-Business Finland Center for Visual and Decision Informatics project Co-Botics, jointly sponsored by Tieto Oy Finland and CA Software.

\bibliographystyle{elsarticle-num}
\bibliography{bibliography}

\journal{pattern recognition}

\newpage
\setcounter{figure}{0}  
\setcounter{table}{0} 
\setcounter{section}{0}  
\setcounter{page}{1}  
\begin{center}\textbf{Multimodal Subspace Support Vector Data Description\\Supplementary Material}
\end{center}

\begin{center}
Fahad Sohrab, Jenni Raitoharju, Alexandros Iosifidis, Moncef Gabbouj
\end{center}

This document contains the extensive results comparing different variants of the proposed MS-SVDD method over 5 different datasets in Section \ref{extensiveresults}, train and test time in Section \ref{trainandtesttime}, and the figures for sensitivity analysis of different hyperparameters over SPECTF heart data set for different regularization terms ($\omega$) and decision strategies ($ds$) for linear MS-SVDD in Section \ref{sensitivityAnslysis}.
\section{MS-SVDD results}
\label{extensiveresults}
\begin{table*}[htbp]
\footnotesize\setlength{\tabcolsep}{1.4pt}
    \renewcommand{\arraystretch}{0.9}
  \centering 
    \caption{Robot Execution Failures dataset test results for MS-SVDD using different decision strategies and constraints (hyperparameters selected using cross-validation on basis of maximum \textit{gm} score on training set)}
\begin{tabular}{llllllllllllllllllllll}
\cline{3-8} \cline{10-15} \cline{17-22}
                                         & \multicolumn{1}{l|}{} & \multicolumn{6}{c|}{Linear}                                                                                                                                            & \multicolumn{1}{l|}{} & \multicolumn{6}{c|}{Kernel}                                                                                                                                            & \multicolumn{1}{l|}{} & \multicolumn{6}{c|}{NPT}                                                                                                                                        \\ \cline{3-8} \cline{10-15} \cline{17-22} 
                                         & \multicolumn{1}{l|}{} & \multicolumn{1}{l|}{accu} & \multicolumn{1}{l|}{tpr}  & \multicolumn{1}{l|}{tnr}  & \multicolumn{1}{l|}{pre}  & \multicolumn{1}{l|}{F1}   & \multicolumn{1}{l|}{gm} & \multicolumn{1}{l|}{} & \multicolumn{1}{l|}{accu} & \multicolumn{1}{l|}{tpr}  & \multicolumn{1}{l|}{tnr}  & \multicolumn{1}{l|}{pre}  & \multicolumn{1}{l|}{F1}   & \multicolumn{1}{l|}{gm} & \multicolumn{1}{l|}{} & \multicolumn{1}{l|}{accu} & \multicolumn{1}{l|}{tpr}  & \multicolumn{1}{l|}{tnr}  & \multicolumn{1}{l|}{pre}  & \multicolumn{1}{l|}{F1}   & \multicolumn{1}{l|}{gm} \\
\end{tabular}
\label{resultsrobotmssvdd}
\end{table*}

\begin{table*}[htbp] 
\footnotesize\setlength{\tabcolsep}{1.4pt}
    \renewcommand{\arraystretch}{0.9}
  \centering   
    \caption{Caltech-7 dataset test results for MS-SVDD using different decision strategies and constraints (hyperparameters selected using cross-validation on basis of maximum \textit{gm} score on training set)}
\begin{tabular}{llllllllllllllllllllll}
\cline{3-8} \cline{10-15} \cline{17-22}
                                         & \multicolumn{1}{l|}{} & \multicolumn{6}{c|}{Linear}                                                                                                                                                    & \multicolumn{1}{l|}{} & \multicolumn{6}{c|}{Kernel}                                                                                                                                                    & \multicolumn{1}{l|}{} & \multicolumn{6}{c|}{NPT}                                                                                                                                                       \\ \cline{3-8} \cline{10-15} \cline{17-22} 
                                         & \multicolumn{1}{l|}{} & \multicolumn{1}{l|}{accu} & \multicolumn{1}{l|}{tpr}  & \multicolumn{1}{l|}{tnr}  & \multicolumn{1}{l|}{pre}  & \multicolumn{1}{l|}{F1}   & \multicolumn{1}{l|}{gm}            & \multicolumn{1}{l|}{} & \multicolumn{1}{l|}{accu} & \multicolumn{1}{l|}{tpr}  & \multicolumn{1}{l|}{tnr}  & \multicolumn{1}{l|}{pre}  & \multicolumn{1}{l|}{F1}   & \multicolumn{1}{l|}{gm}            & \multicolumn{1}{l|}{} & \multicolumn{1}{l|}{accu} & \multicolumn{1}{l|}{tpr}  & \multicolumn{1}{l|}{tnr}  & \multicolumn{1}{l|}{pre}  & \multicolumn{1}{l|}{F1}   & \multicolumn{1}{l|}{gm}            \\
\end{tabular}
\label{resultscaltechmssvdd}
\end{table*}

\begin{table*}[htbp]  
\footnotesize\setlength{\tabcolsep}{1.4pt}
    \renewcommand{\arraystretch}{0.9}
  \centering   
    \caption{Ionosphere dataset test results for MS-SVDD using different decision strategies and constraints (hyperparameters selected using cross-validation on basis of maximum \textit{gm} score on training set)}
\begin{tabular}{llllllllllllllllllllll}
\cline{3-8} \cline{10-15} \cline{17-22}
                                         & \multicolumn{1}{l|}{} & \multicolumn{6}{c|}{Linear}                                                                                                                                                    & \multicolumn{1}{l|}{} & \multicolumn{6}{c|}{Kernel}                                                                                                                                                    & \multicolumn{1}{l|}{} & \multicolumn{6}{c|}{NPT}                                                                                                                                                       \\ \cline{3-8} \cline{10-15} \cline{17-22} 
                                         & \multicolumn{1}{l|}{} & \multicolumn{1}{l|}{accu} & \multicolumn{1}{l|}{tpr}  & \multicolumn{1}{l|}{tnr}  & \multicolumn{1}{l|}{pre}  & \multicolumn{1}{l|}{F1}   & \multicolumn{1}{l|}{gm}            & \multicolumn{1}{l|}{} & \multicolumn{1}{l|}{accu} & \multicolumn{1}{l|}{tpr}  & \multicolumn{1}{l|}{tnr}  & \multicolumn{1}{l|}{pre}  & \multicolumn{1}{l|}{F1}   & \multicolumn{1}{l|}{gm}            & \multicolumn{1}{l|}{} & \multicolumn{1}{l|}{accu} & \multicolumn{1}{l|}{tpr}  & \multicolumn{1}{l|}{tnr}  & \multicolumn{1}{l|}{pre}  & \multicolumn{1}{l|}{F1}   & \multicolumn{1}{l|}{gm}            \\
\end{tabular}
\label{resultsionospheremssvdd}
\end{table*}

\begin{table*}[htbp]  
\footnotesize\setlength{\tabcolsep}{1.4pt}
    \renewcommand{\arraystretch}{0.9}
  \centering   
    \caption{Handwritten dataset test results for MS-SVDD using different decision strategies and constraints (hyperparameters selected using cross-validation on basis of maximum \textit{gm} score on training set)}
\begin{tabular}{llllllllllllllllllllll}
\cline{3-8} \cline{10-15} \cline{17-22}
                                         & \multicolumn{1}{l|}{} & \multicolumn{6}{c|}{Linear}                                                                                                                                                    & \multicolumn{1}{l|}{} & \multicolumn{6}{c|}{Kernel}                                                                                                                                                    & \multicolumn{1}{l|}{} & \multicolumn{6}{c|}{NPT}                                                                                                                                                       \\ \cline{3-8} \cline{10-15} \cline{17-22} 
                                         & \multicolumn{1}{l|}{} & \multicolumn{1}{l|}{accu} & \multicolumn{1}{l|}{tpr}  & \multicolumn{1}{l|}{tnr}  & \multicolumn{1}{l|}{pre}  & \multicolumn{1}{l|}{F1}   & \multicolumn{1}{l|}{gm}            & \multicolumn{1}{l|}{} & \multicolumn{1}{l|}{accu} & \multicolumn{1}{l|}{tpr}  & \multicolumn{1}{l|}{tnr}  & \multicolumn{1}{l|}{pre}  & \multicolumn{1}{l|}{F1}   & \multicolumn{1}{l|}{gm}            & \multicolumn{1}{l|}{} & \multicolumn{1}{l|}{accu} & \multicolumn{1}{l|}{tpr}  & \multicolumn{1}{l|}{tnr}  & \multicolumn{1}{l|}{pre}  & \multicolumn{1}{l|}{F1}   & \multicolumn{1}{l|}{gm}            \\
\end{tabular}
\label{resultshandwrittenmssvdd}
\end{table*}

\begin{table*}[htbp]  
\footnotesize\setlength{\tabcolsep}{1.4pt}
    \renewcommand{\arraystretch}{0.9}
  \centering   
    \caption{SPECTF heart dataset test results for MS-SVDD using different decision strategies and constraints (hyperparameters selected using cross-validation on basis of maximum \textit{gm} score on training set)}
\begin{tabular}{llllllllllllllllllllll}
\cline{3-8} \cline{10-15} \cline{17-22}
                                         & \multicolumn{1}{l|}{} & \multicolumn{6}{c|}{Linear}                                                                                                                                                    & \multicolumn{1}{l|}{} & \multicolumn{6}{c|}{Kernel}                                                                                                                                                    & \multicolumn{1}{l|}{} & \multicolumn{6}{c|}{NPT}                                                                                                                                                       \\ \cline{3-8} \cline{10-15} \cline{17-22} 
                                         & \multicolumn{1}{l|}{} & \multicolumn{1}{l|}{accu} & \multicolumn{1}{l|}{tpr}  & \multicolumn{1}{l|}{tnr}  & \multicolumn{1}{l|}{pre}  & \multicolumn{1}{l|}{F1}   & \multicolumn{1}{l|}{gm}            & \multicolumn{1}{l|}{} & \multicolumn{1}{l|}{accu} & \multicolumn{1}{l|}{tpr}  & \multicolumn{1}{l|}{tnr}  & \multicolumn{1}{l|}{pre}  & \multicolumn{1}{l|}{F1}   & \multicolumn{1}{l|}{gm}            & \multicolumn{1}{l|}{} & \multicolumn{1}{l|}{accu} & \multicolumn{1}{l|}{tpr}  & \multicolumn{1}{l|}{tnr}  & \multicolumn{1}{l|}{pre}  & \multicolumn{1}{l|}{F1}   & \multicolumn{1}{l|}{gm}            \\
\end{tabular}
\label{resultsheartmssvdd}
\end{table*}

\newpage
\section{Train and test time}
The numerical results of the train and test time (Tables 6-10) were computed using Matlab version 9.6.0.1072779 (R2019a) over, Intel64 Family 6 Model 158 Stepping 9 GenuineIntel ~2808 Mhz processor.
\label{trainandtesttime}
\begin{table*}[htbp]  
\footnotesize\setlength{\tabcolsep}{1.8pt}
    \renewcommand{\arraystretch}{1}
  \centering   
    \caption{Train and test time (in milliseconds) for Robot Execution Failures dataset}
\begin{tabular}{lllllll}
\cline{3-4} \cline{6-7}
                                              & \multicolumn{1}{l|}{} & \multicolumn{2}{c|}{Linear}                                      & \multicolumn{1}{l|}{} & \multicolumn{2}{c|}{Non-linear}                                  \\ \cline{3-4} \cline{6-7} 
                                              & \multicolumn{1}{l|}{} & \multicolumn{1}{l|}{Train Time} & \multicolumn{1}{l|}{Test Time} & \multicolumn{1}{l|}{} & \multicolumn{1}{l|}{Train Time} & \multicolumn{1}{l|}{Test Time} \\ \cline{3-4} \cline{6-7} 
Proposed method                               &                       &                                 &                                &                       &                                 &                                \\ \cline{1-1} \cline{3-4} \cline{6-7} 
\multicolumn{1}{|l|}{MS-SVDD   $\omega_2ds3$} & \multicolumn{1}{l|}{} & \multicolumn{1}{l|}{1.01E+02}   & \multicolumn{1}{l|}{3.43E-01}  & \multicolumn{1}{l|}{} & \multicolumn{1}{l|}{1.66E+02}   & \multicolumn{1}{l|}{3.26E-01}  \\ \cline{1-1} \cline{3-4} \cline{6-7} 
\multicolumn{1}{|l|}{MS-SVDD $\omega_5ds3$}   & \multicolumn{1}{l|}{} & \multicolumn{1}{l|}{9.03E+01}   & \multicolumn{1}{l|}{3.58E-01}  & \multicolumn{1}{l|}{} & \multicolumn{1}{l|}{1.65E+02}   & \multicolumn{1}{l|}{3.63E-01}  \\ \cline{1-1} \cline{3-4} \cline{6-7} 
Concatenated features                         &                       &                                 &                                &                       &                                 &                                \\ \cline{1-1} \cline{3-4} \cline{6-7} 
\multicolumn{1}{|l|}{S-SVDD $\psi_1$}         & \multicolumn{1}{l|}{} & \multicolumn{1}{l|}{4.66E+01}   & \multicolumn{1}{l|}{1.04E-01}  & \multicolumn{1}{l|}{} & \multicolumn{1}{l|}{8.52E+01}   & \multicolumn{1}{l|}{1.97E-01}  \\
\end{tabular}
\label{robothtime}
\end{table*}

\begin{table*}[htbp]  
\footnotesize\setlength{\tabcolsep}{1.8pt}
    \renewcommand{\arraystretch}{1}
  \centering   
    \caption{Train and test time (in milliseconds) for Caltec-7 dataset}
\begin{tabular}{lllllll}
\cline{3-4} \cline{6-7}
                                              & \multicolumn{1}{l|}{} & \multicolumn{2}{c|}{Linear}                                      & \multicolumn{1}{l|}{} & \multicolumn{2}{c|}{Non-linear}                                  \\ \cline{3-4} \cline{6-7} 
                                              & \multicolumn{1}{l|}{} & \multicolumn{1}{l|}{Train Time} & \multicolumn{1}{l|}{Test Time} & \multicolumn{1}{l|}{} & \multicolumn{1}{l|}{Train Time} & \multicolumn{1}{l|}{Test Time} \\ \cline{3-4} \cline{6-7} 
Proposed method                               &                       &                                 &                                &                       &                                 &                                \\ \cline{1-1} \cline{3-4} \cline{6-7} 
\multicolumn{1}{|l|}{MS-SVDD   $\omega_1ds1$} & \multicolumn{1}{l|}{} & \multicolumn{1}{l|}{1.81E+02}   & \multicolumn{1}{l|}{7.61E-01}  & \multicolumn{1}{l|}{} & \multicolumn{1}{l|}{5.36E+02}   & \multicolumn{1}{l|}{1.29E+00}  \\ \cline{1-1} \cline{3-4} \cline{6-7} 
\multicolumn{1}{|l|}{MS-SVDD $\omega_4ds1$}   & \multicolumn{1}{l|}{} & \multicolumn{1}{l|}{1.94E+02}   & \multicolumn{1}{l|}{7.95E-01}  & \multicolumn{1}{l|}{} & \multicolumn{1}{l|}{4.11E+02}   & \multicolumn{1}{l|}{1.28E+00}  \\ \cline{1-1} \cline{3-4} \cline{6-7} 
Concatenated features                                &                       &                                 &                                &                       &                                 &                                \\ \cline{1-1} \cline{3-4} \cline{6-7} 
\multicolumn{1}{|l|}{S-SVDD $\psi_1$}         & \multicolumn{1}{l|}{} & \multicolumn{1}{l|}{1.37E+02}   & \multicolumn{1}{l|}{2.46E-01}  & \multicolumn{1}{l|}{} & \multicolumn{1}{l|}{7.15E+02}   & \multicolumn{1}{l|}{4.33E-01}  \\
\end{tabular}
\label{caltechtime}
\end{table*}

\begin{table*}[htbp]  
\footnotesize\setlength{\tabcolsep}{1.8pt}
    \renewcommand{\arraystretch}{1}
          \centering   
    \caption{Train and test time (in milliseconds) for Ionosphere dataset}
\begin{tabular}{lllllll}
\cline{3-4} \cline{6-7}
                                              & \multicolumn{1}{l|}{} & \multicolumn{2}{c|}{Linear}                                      & \multicolumn{1}{l|}{} & \multicolumn{2}{c|}{Non-linear}                                  \\ \cline{3-4} \cline{6-7} 
                                              & \multicolumn{1}{l|}{} & \multicolumn{1}{l|}{Train Time} & \multicolumn{1}{l|}{Test Time} & \multicolumn{1}{l|}{} & \multicolumn{1}{l|}{Train Time} & \multicolumn{1}{l|}{Test Time} \\ \cline{3-4} \cline{6-7} 
Proposed method                               &                       &                                 &                                &                       &                                 &                                \\ \cline{1-1} \cline{3-4} \cline{6-7} 
\multicolumn{1}{|l|}{MS-SVDD   $\omega_2ds4$} & \multicolumn{1}{l|}{} & \multicolumn{1}{l|}{1.68E+02}   & \multicolumn{1}{l|}{3.82E-01}  & \multicolumn{1}{l|}{} & \multicolumn{1}{l|}{2.11E+02}   & \multicolumn{1}{l|}{3.99E-01}  \\ \cline{1-1} \cline{3-4} \cline{6-7} 
\multicolumn{1}{|l|}{MS-SVDD $\omega_1ds4$}   & \multicolumn{1}{l|}{} & \multicolumn{1}{l|}{1.79E+02}   & \multicolumn{1}{l|}{3.78E-01}  & \multicolumn{1}{l|}{} & \multicolumn{1}{l|}{2.29E+02}   & \multicolumn{1}{l|}{3.99E-01}  \\ \cline{1-1} \cline{3-4} \cline{6-7} 
Concatenated features                         &                       &                                 &                                &                       &                                 &                                \\ \cline{1-1} \cline{3-4} \cline{6-7} 
\multicolumn{1}{|l|}{S-SVDD $\psi_1$}         & \multicolumn{1}{l|}{} & \multicolumn{1}{l|}{5.65E+01}   & \multicolumn{1}{l|}{6.92E-02}  & \multicolumn{1}{l|}{} & \multicolumn{1}{l|}{1.50E+02}   & \multicolumn{1}{l|}{2.29E-01}  \\
\end{tabular}
\label{ionospheretime}
\end{table*}

\begin{table*}[htbp]  
\footnotesize\setlength{\tabcolsep}{1.8pt}
    \renewcommand{\arraystretch}{1}
      \centering   
    \caption{Train and test time (in milliseconds) for Handwritten dataset}
\begin{tabular}{lllllll}
\cline{3-4} \cline{6-7}
                                              & \multicolumn{1}{l|}{} & \multicolumn{2}{c|}{Linear}                                      & \multicolumn{1}{l|}{} & \multicolumn{2}{c|}{Non-linear}                                  \\ \cline{3-4} \cline{6-7} 
                                              & \multicolumn{1}{l|}{} & \multicolumn{1}{l|}{Train Time} & \multicolumn{1}{l|}{Test Time} & \multicolumn{1}{l|}{} & \multicolumn{1}{l|}{Train Time} & \multicolumn{1}{l|}{Test Time} \\ \cline{3-4} \cline{6-7} 
Proposed method                               &                       &                                 &                                &                       &                                 &                                \\ \cline{1-1} \cline{3-4} \cline{6-7} 
\multicolumn{1}{|l|}{MS-SVDD   $\omega_4ds4$} & \multicolumn{1}{l|}{} & \multicolumn{1}{l|}{1.18E+02}   & \multicolumn{1}{l|}{1.28E+00}  & \multicolumn{1}{l|}{} & \multicolumn{1}{l|}{1.07E+02}   & \multicolumn{1}{l|}{1.14E+00}  \\ \cline{1-1} \cline{3-4} \cline{6-7} 
\multicolumn{1}{|l|}{MS-SVDD $\omega_4ds1$}   & \multicolumn{1}{l|}{} & \multicolumn{1}{l|}{1.30E+02}   & \multicolumn{1}{l|}{1.39E+00}  & \multicolumn{1}{l|}{} & \multicolumn{1}{l|}{1.42E+02}   & \multicolumn{1}{l|}{1.94E+00}  \\ \cline{1-1} \cline{3-4} \cline{6-7} 
Concatenated features                         &                       &                                 &                                &                       &                                 &                                \\ \cline{1-1} \cline{3-4} \cline{6-7} 
\multicolumn{1}{|l|}{S-SVDD $\psi_1$}         & \multicolumn{1}{l|}{} & \multicolumn{1}{l|}{4.19E+01}   & \multicolumn{1}{l|}{2.27E-01}  & \multicolumn{1}{l|}{} & \multicolumn{1}{l|}{2.47E+02}   & \multicolumn{1}{l|}{6.25E-01}  \\
\end{tabular}
\label{handwrittentime}
\end{table*}

\begin{table*}[htbp]  
\footnotesize\setlength{\tabcolsep}{1.8pt}
    \renewcommand{\arraystretch}{1}
  \centering   
    \caption{Train and test time (in milliseconds) for SPECTF heart dataset}
\begin{tabular}{lllllll}
\cline{3-4} \cline{6-7}
                                              & \multicolumn{1}{l|}{} & \multicolumn{2}{c|}{Linear}                                      & \multicolumn{1}{l|}{} & \multicolumn{2}{c|}{Non-linear}                                  \\ \cline{3-4} \cline{6-7} 
                                              & \multicolumn{1}{l|}{} & \multicolumn{1}{l|}{Train Time} & \multicolumn{1}{l|}{Test Time} & \multicolumn{1}{l|}{} & \multicolumn{1}{l|}{Train Time} & \multicolumn{1}{l|}{Test Time} \\ \cline{3-4} \cline{6-7} 
Proposed method                               &                       &                                 &                                &                       &                                 &                                \\ \cline{1-1} \cline{3-4} \cline{6-7} 
\multicolumn{1}{|l|}{MS-SVDD   $\omega_0ds1$} & \multicolumn{1}{l|}{} & \multicolumn{1}{l|}{4.16E+01}   & \multicolumn{1}{l|}{5.44E-01}  & \multicolumn{1}{l|}{} & \multicolumn{1}{l|}{5.20E+01}   & \multicolumn{1}{l|}{5.41E-01}  \\ \cline{1-1} \cline{3-4} \cline{6-7} 
\multicolumn{1}{|l|}{MS-SVDD $\omega_2ds1$}   & \multicolumn{1}{l|}{} & \multicolumn{1}{l|}{4.60E+01}   & \multicolumn{1}{l|}{5.82E-01}  & \multicolumn{1}{l|}{} & \multicolumn{1}{l|}{5.07E+01}   & \multicolumn{1}{l|}{2.29E-01}  \\ \cline{1-1} \cline{3-4} \cline{6-7} 
Concatenated features                         &                       &                                 &                                &                       &                                 &                                \\ \cline{1-1} \cline{3-4} \cline{6-7} 
\multicolumn{1}{|l|}{S-SVDD $\psi_1$}         & \multicolumn{1}{l|}{} & \multicolumn{1}{l|}{6.27E+01}   & \multicolumn{1}{l|}{7.59E-02}  & \multicolumn{1}{l|}{} & \multicolumn{1}{l|}{6.19E+01}   & \multicolumn{1}{l|}{3.35E-01}  \\
\end{tabular}
\label{hearttime}
\end{table*}

\newpage
\section{Sensitivity analysis}
\label{sensitivityAnslysis}
 \begin{figure*}[ht]
	\centering
	\includegraphics[width=\textwidth]{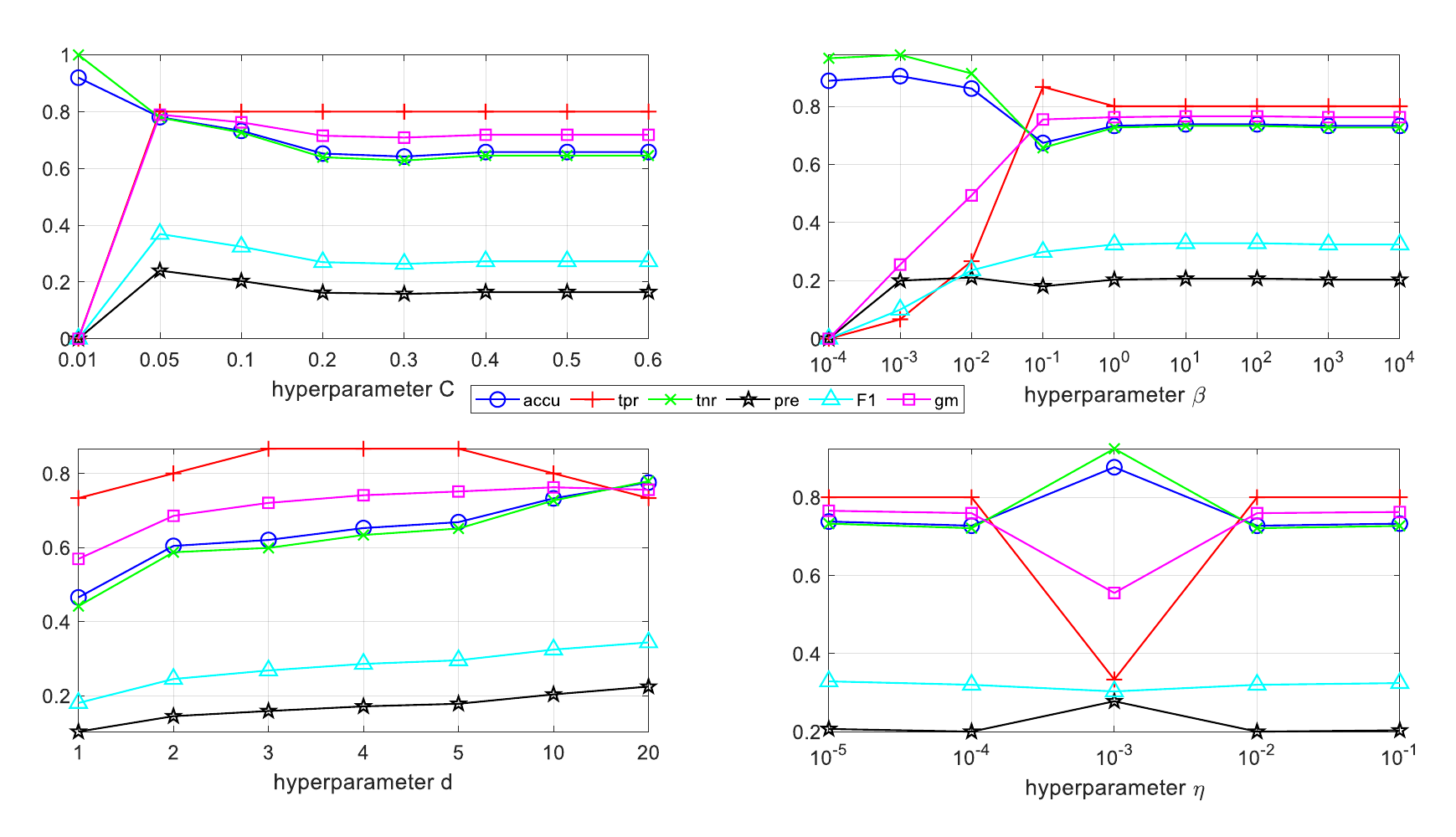}
	\caption{Hyperparameter sensitivity analysis for linear MS-SVDD $\omega_1ds1$ on SPECTF heart dataset}
	\label{w1d1}
\end{figure*}
 \begin{figure*}[ht]
	\centering
	\includegraphics[width=\textwidth]{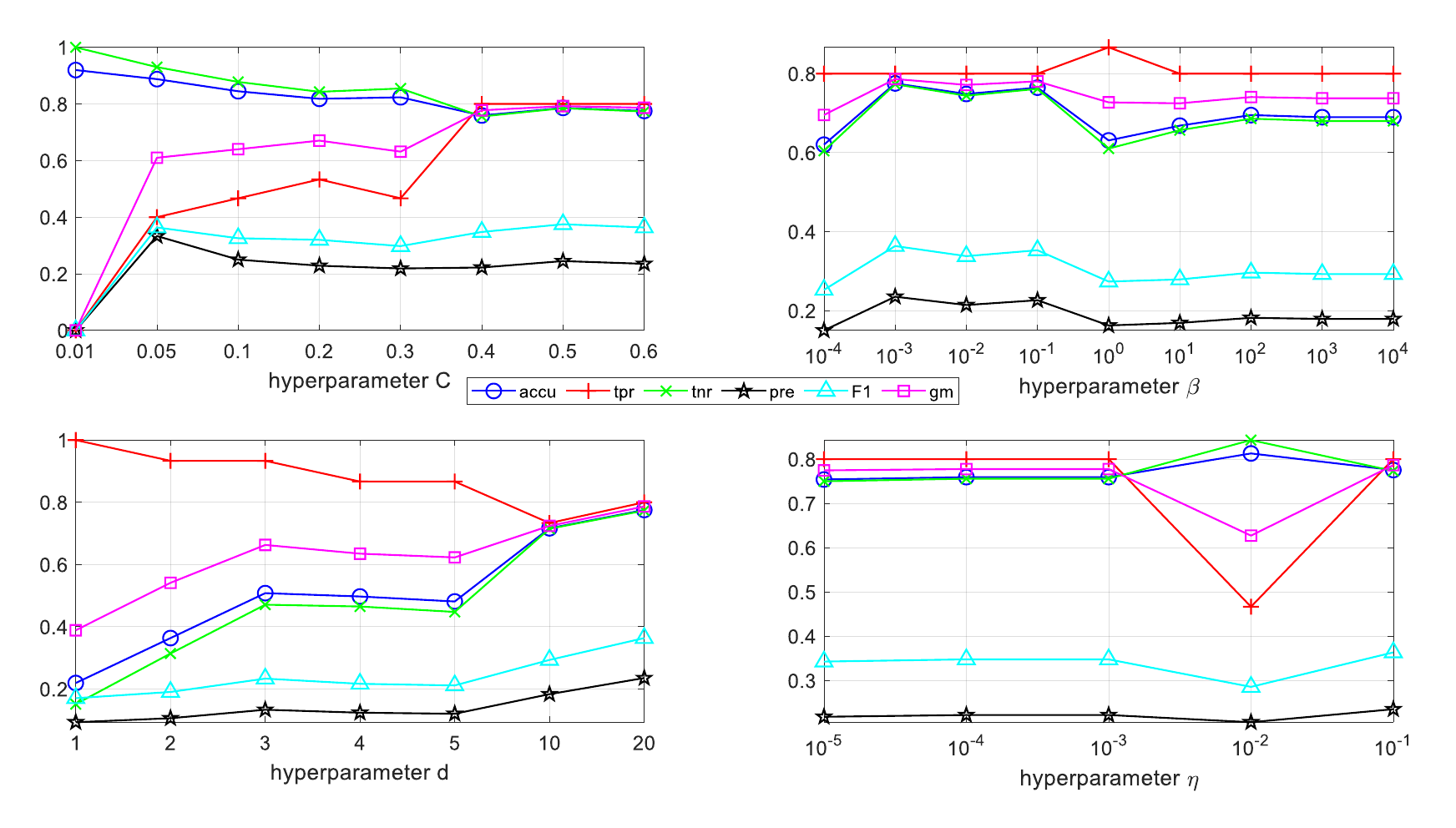}
	\caption{Hyperparameter sensitivity analysis for linear MS-SVDD $\omega_2ds1$ on SPECTF heart dataset}
	\label{w2d1}
\end{figure*}
 \begin{figure*}[ht]
	\centering
	\includegraphics[width=\textwidth]{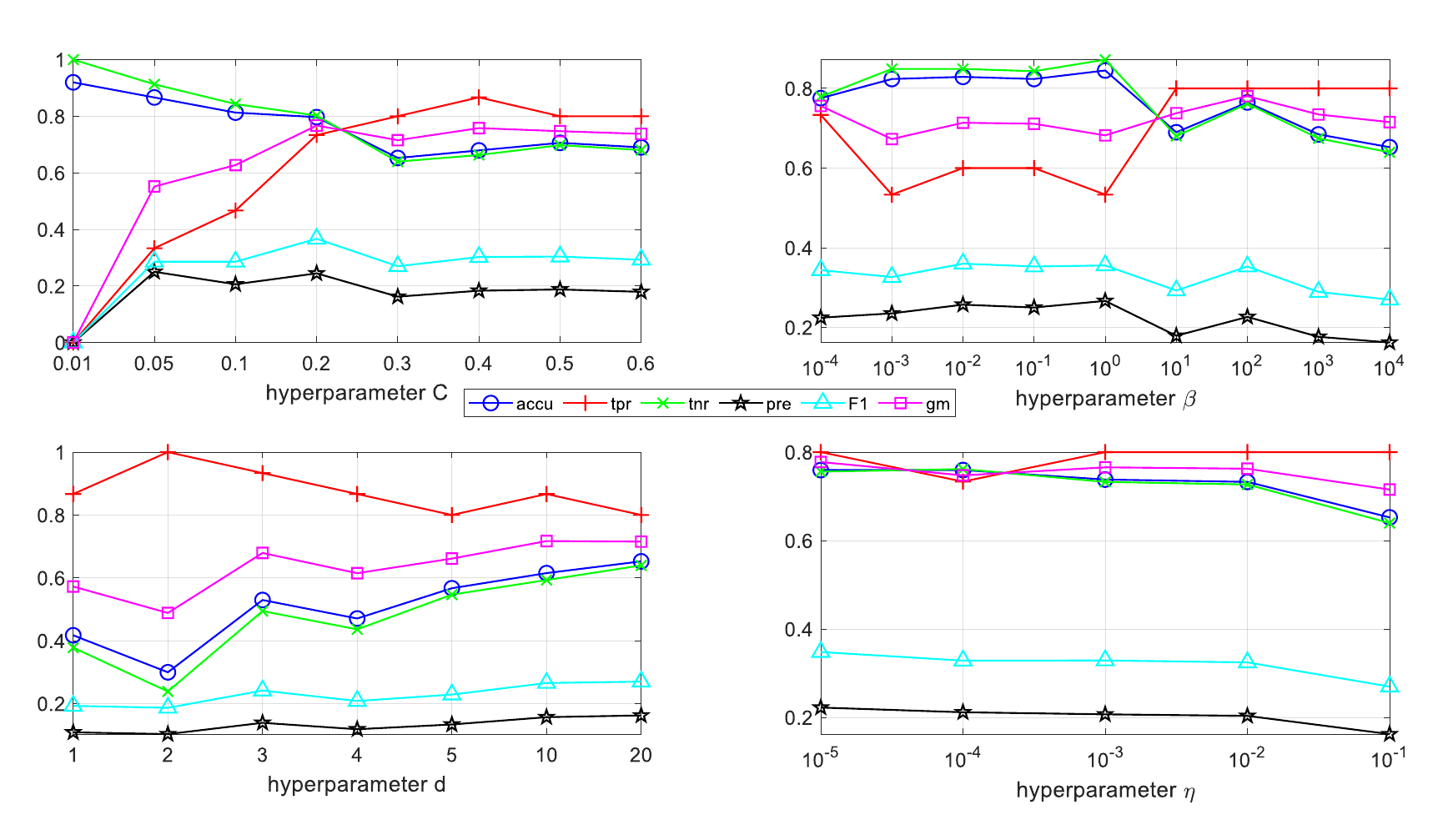}
	\caption{Hyperparameter sensitivity analysis for linear MS-SVDD $\omega_3ds1$ on SPECTF heart dataset}
	\label{w3d1}
\end{figure*}
 \begin{figure*}[ht]
	\centering
	\includegraphics[width=\textwidth]{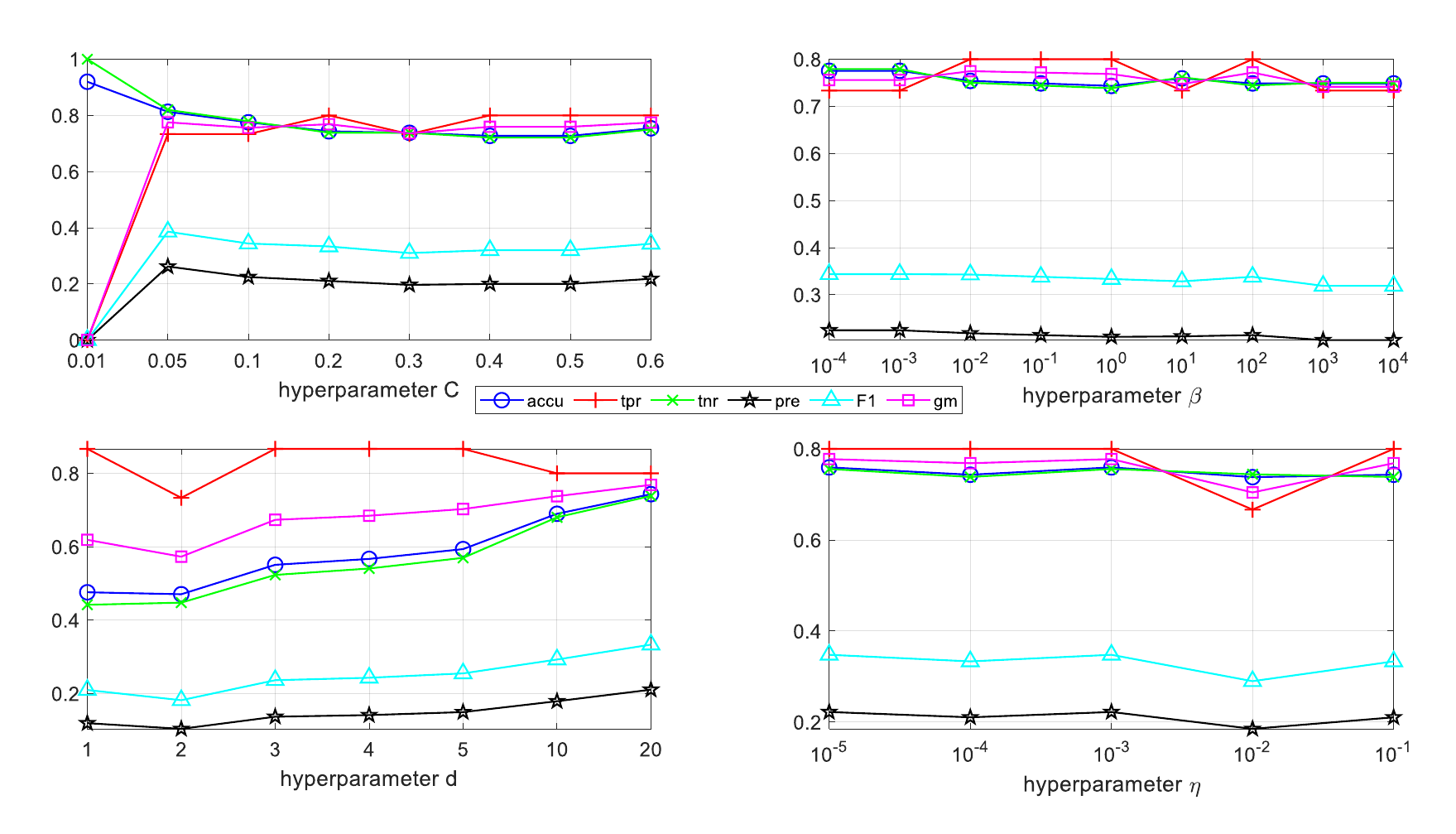}
	\caption{Hyperparameter sensitivity analysis for linear MS-SVDD $\omega_4ds1$ on SPECTF heart dataset}
	\label{w4d1}
\end{figure*}
 \begin{figure*}[ht]
	\centering
	\includegraphics[width=\textwidth]{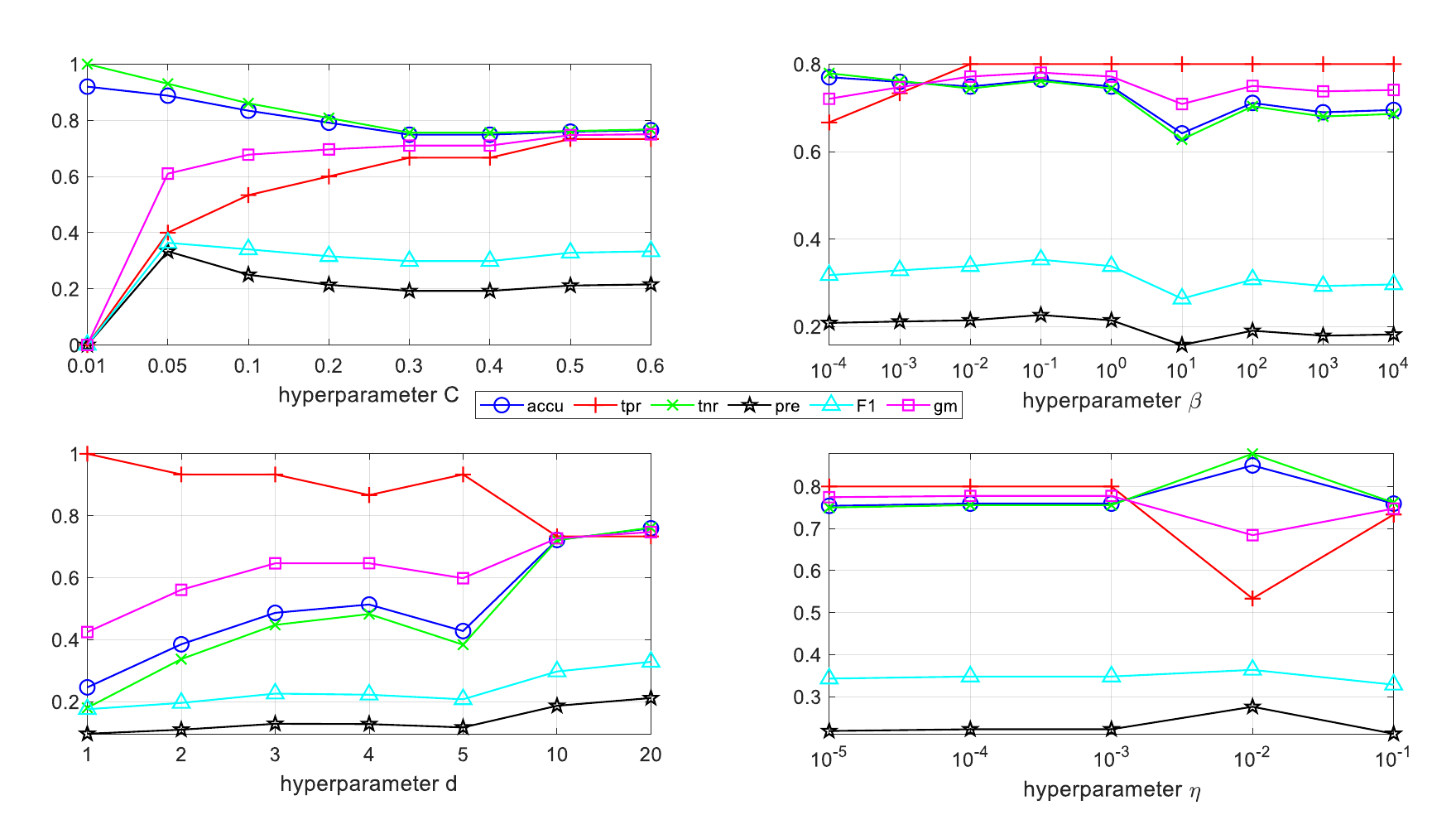}
	\caption{Hyperparameter sensitivity analysis for linear MS-SVDD $\omega_5ds1$ on SPECTF heart dataset}
\end{figure*}
 \begin{figure*}[ht]
	\centering
	\includegraphics[width=\textwidth]{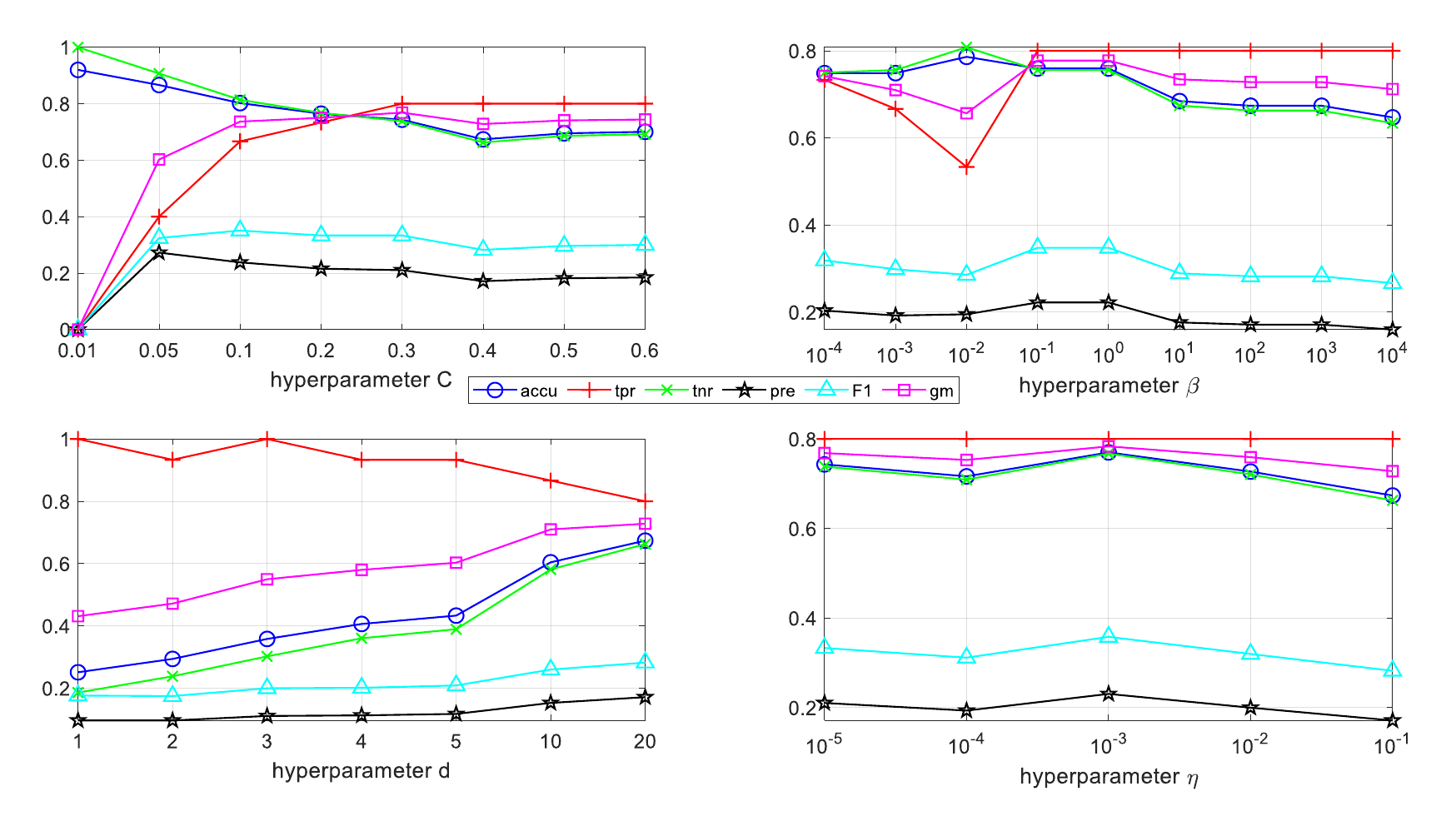}
	\caption{Hyperparameter sensitivity analysis for linear MS-SVDD $\omega_6ds1$ on SPECTF heart dataset}
	\label{w6d1}
\end{figure*}
 \begin{figure*}[ht]
	\centering
	\includegraphics[width=\textwidth]{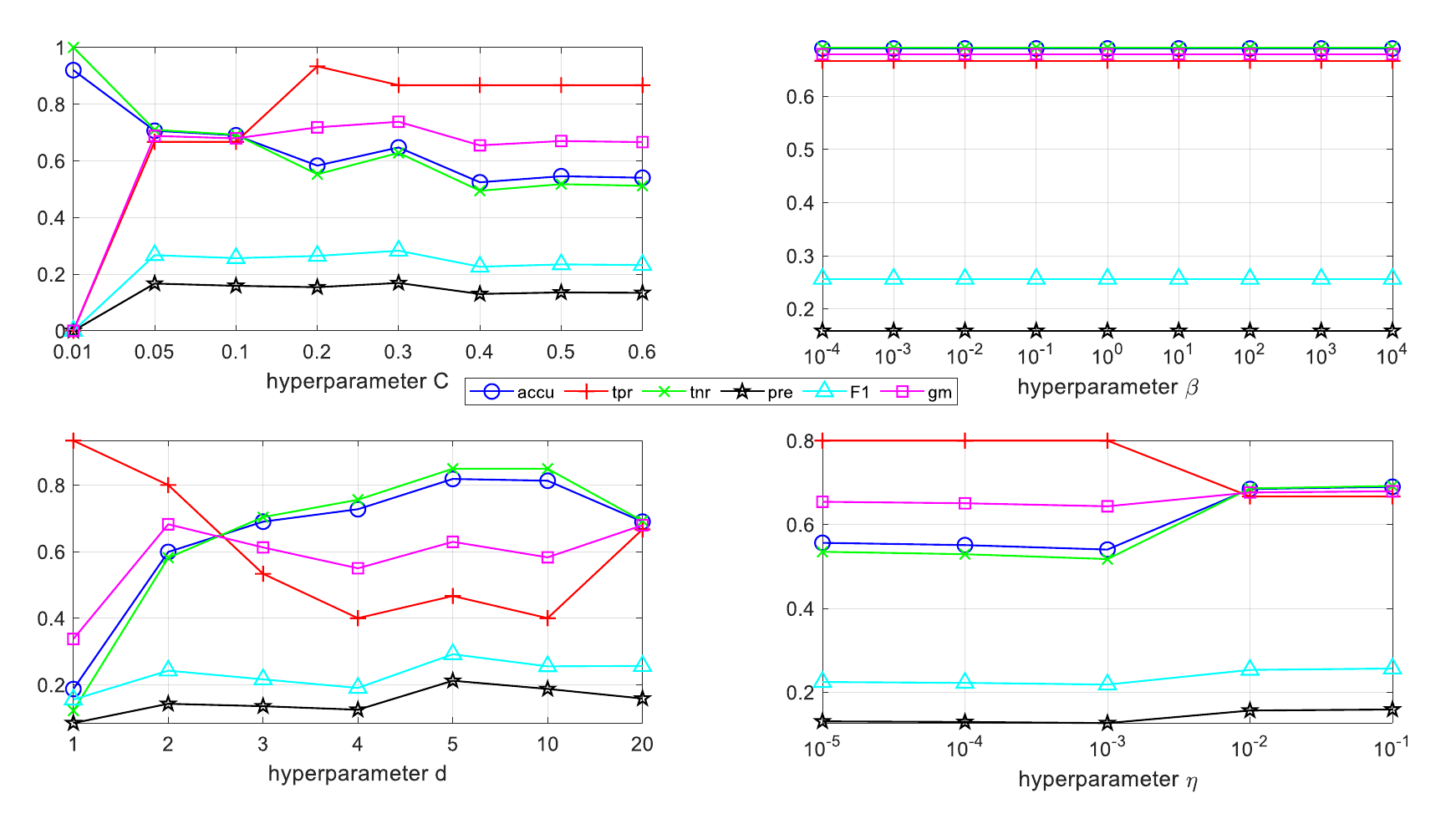}
	\caption{Hyperparameter sensitivity analysis for linear MS-SVDD $\omega_0ds2$ on SPECTF heart dataset}
	\label{w0d2}
\end{figure*}
 \begin{figure*}[ht]
	\centering
	\includegraphics[width=\textwidth]{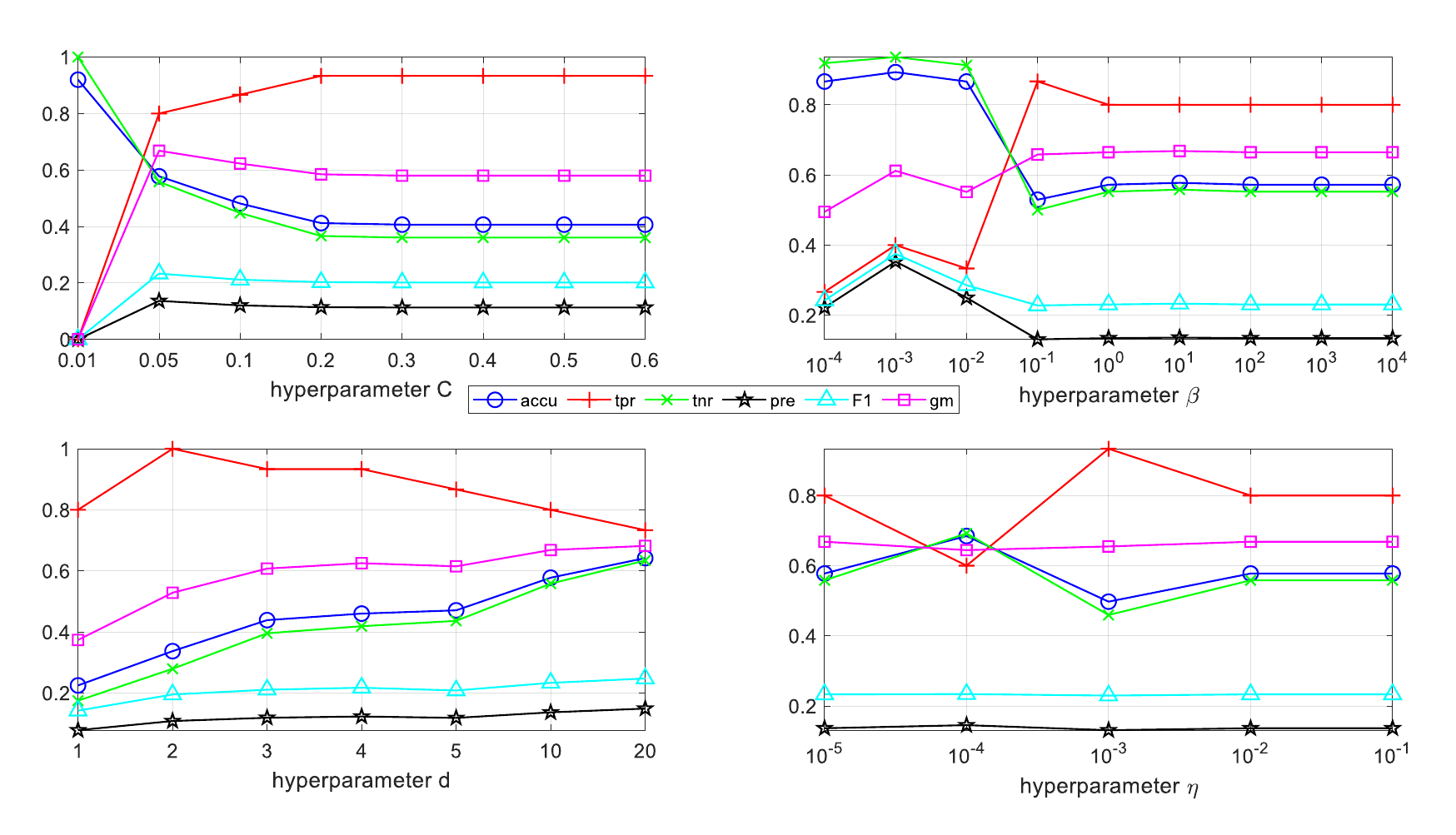}
	\caption{Hyperparameter sensitivity analysis for linear MS-SVDD $\omega_1ds2$ on SPECTF heart dataset}
	\label{w1d2}
\end{figure*}
 \begin{figure*}[ht]
	\centering
	\includegraphics[width=\textwidth]{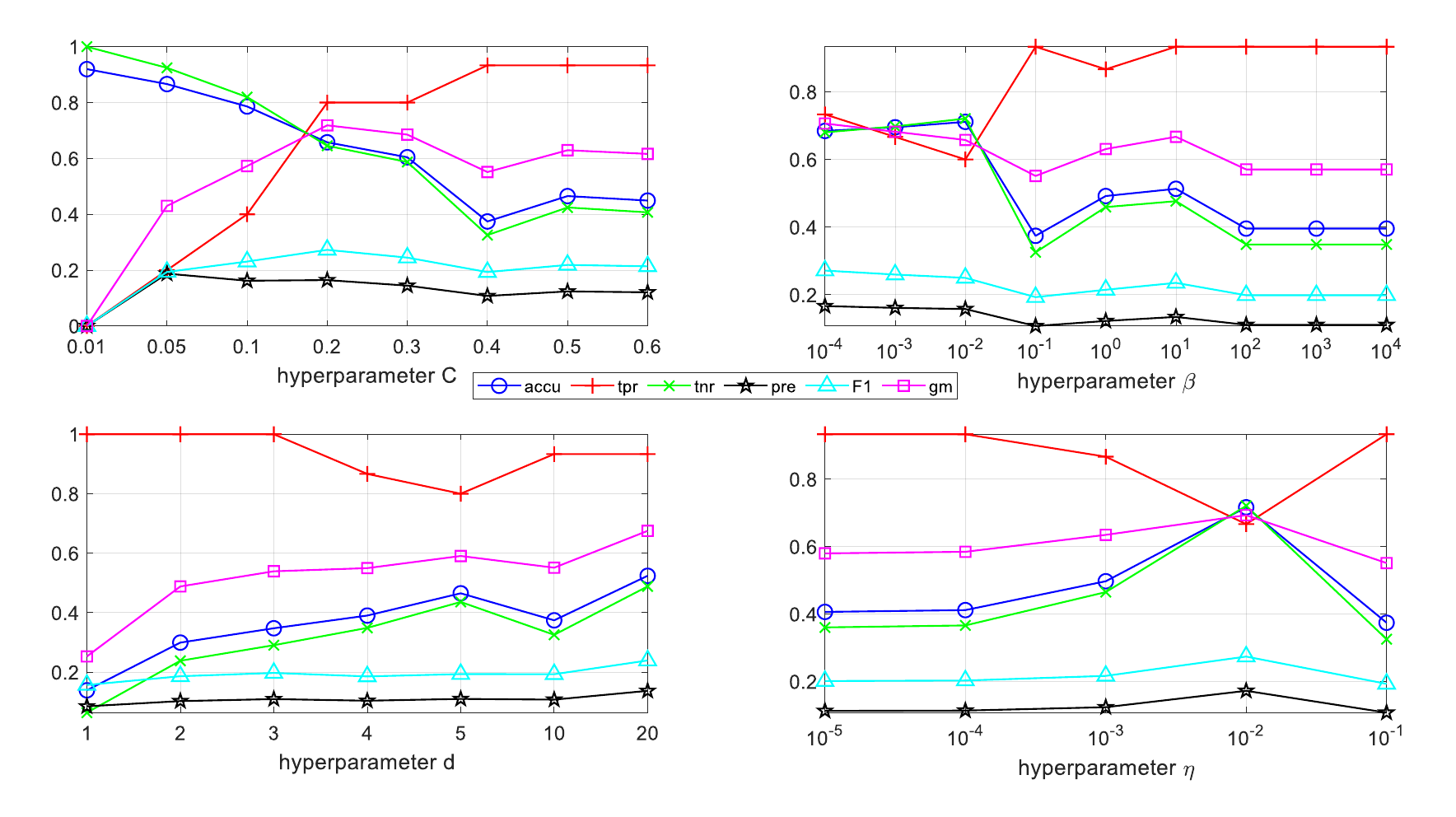}
	\caption{Hyperparameter sensitivity analysis for linear MS-SVDD $\omega_2ds2$ on SPECTF heart dataset}
	\label{w2d2}
\end{figure*}
 \begin{figure*}[ht]
	\centering
	\includegraphics[width=\textwidth]{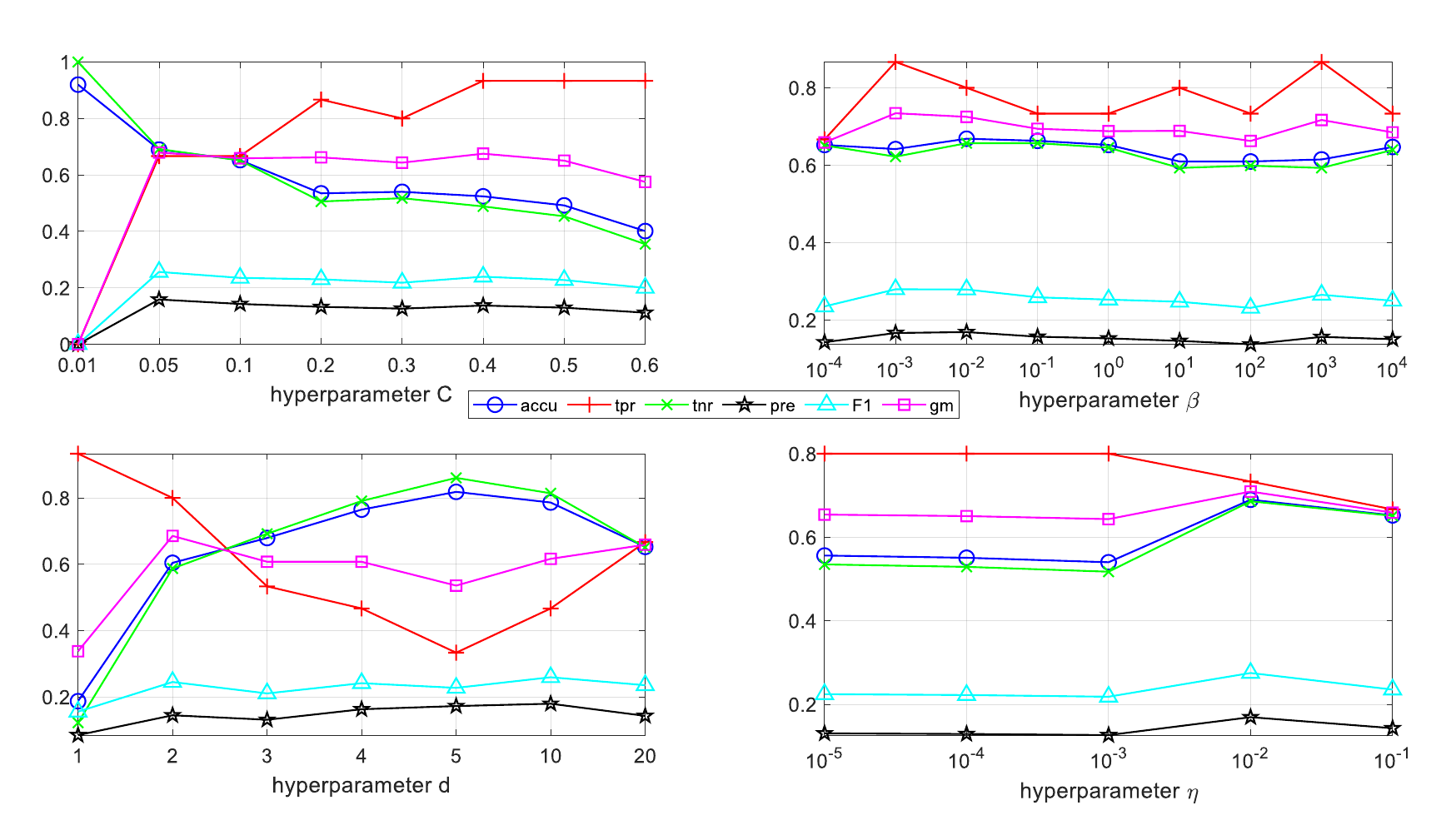}
	\caption{Hyperparameter sensitivity analysis for linear MS-SVDD $\omega_3ds2$ on SPECTF heart dataset}
	\label{w3d2}
\end{figure*}
 \begin{figure*}[ht]
	\centering
	\includegraphics[width=\textwidth]{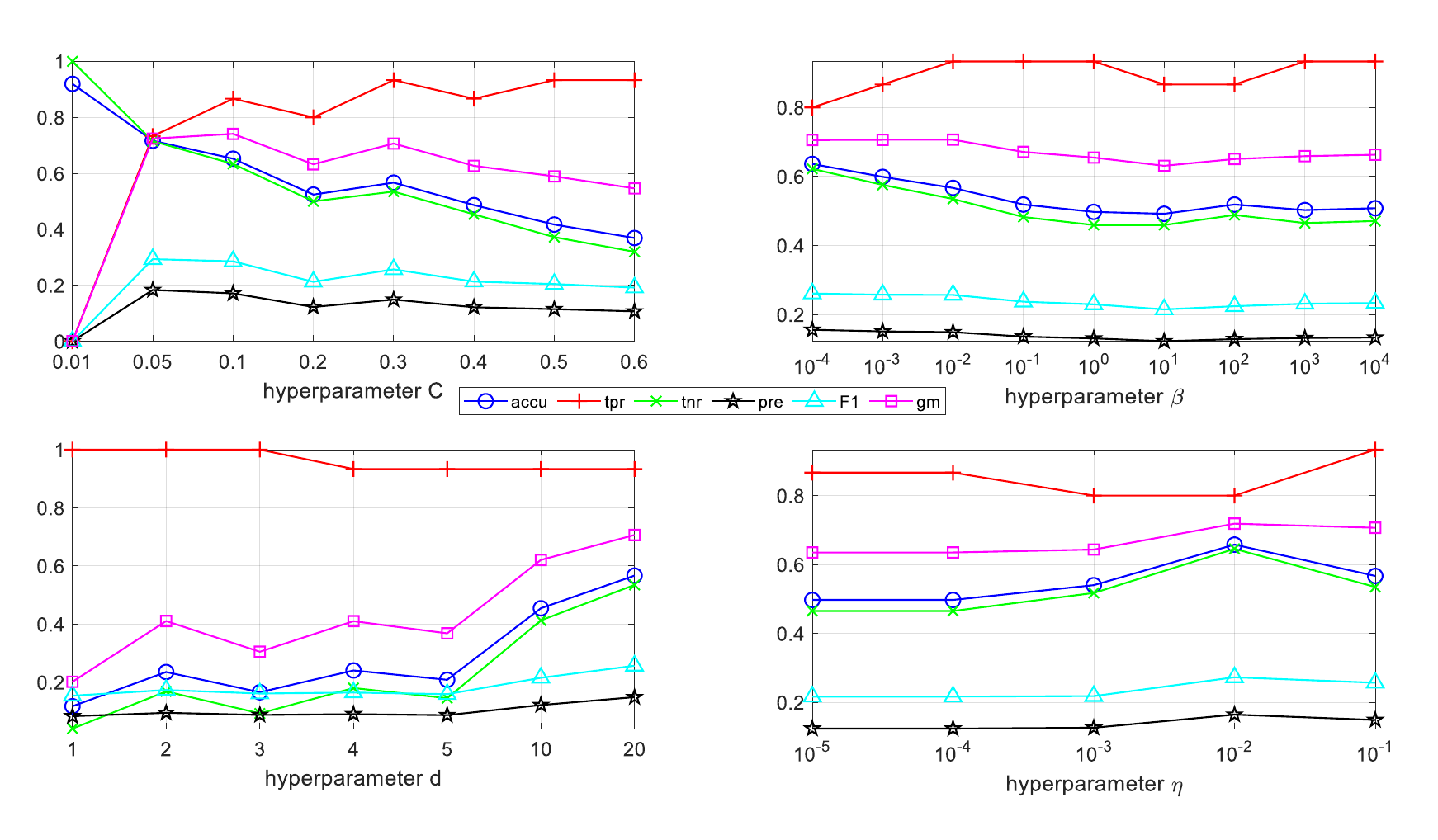}
	\caption{Hyperparameter sensitivity analysis for linear MS-SVDD $\omega_4ds2$ on SPECTF heart dataset}
	\label{w4d2}
\end{figure*}
 \begin{figure*}[ht]
	\centering
	\includegraphics[width=\textwidth]{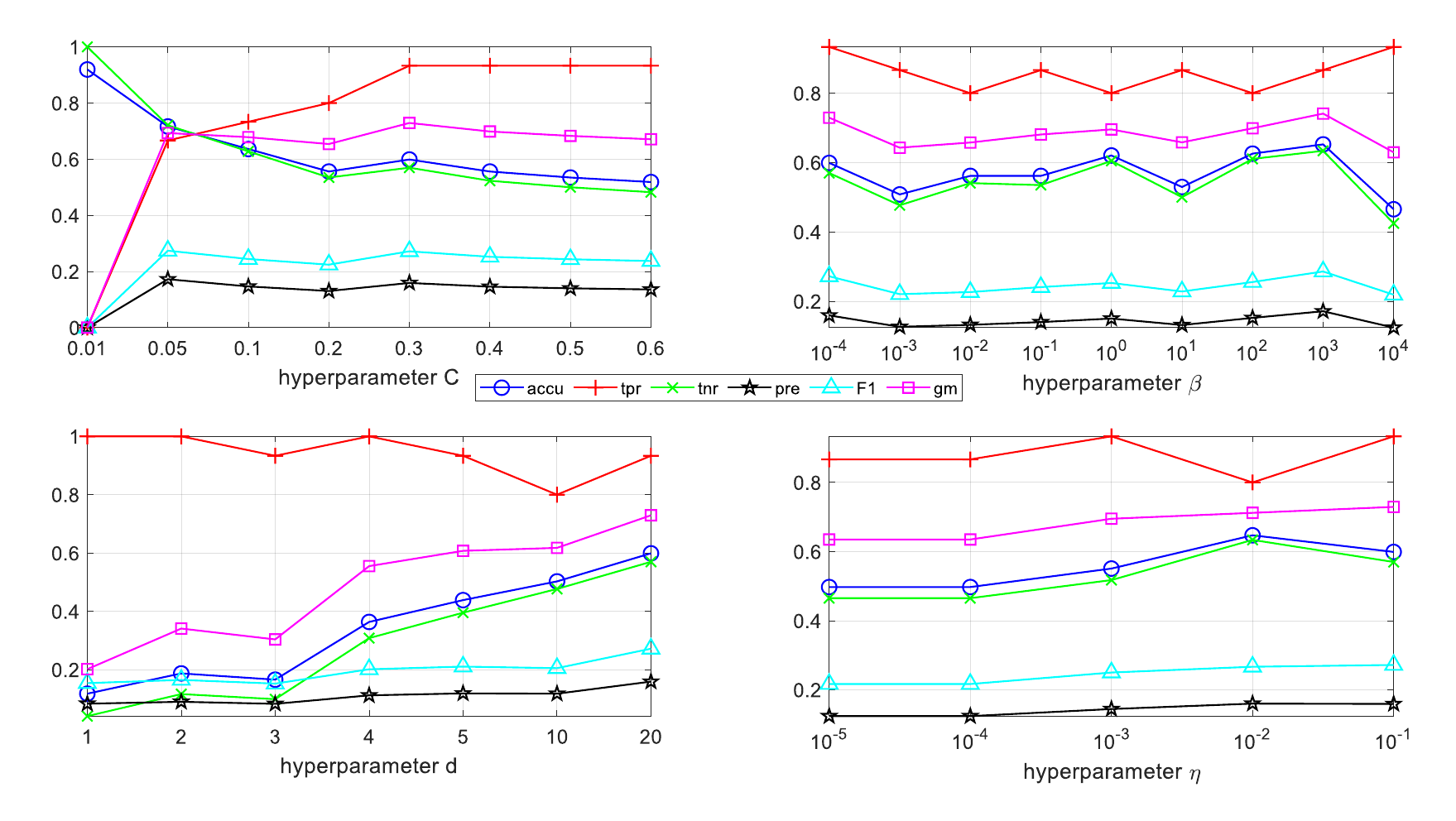}
	\caption{Hyperparameter sensitivity analysis for linear MS-SVDD $\omega_5ds2$ on SPECTF heart dataset}
	\label{w5d2}
\end{figure*}
 \begin{figure*}[ht]
	\centering
	\includegraphics[width=\textwidth]{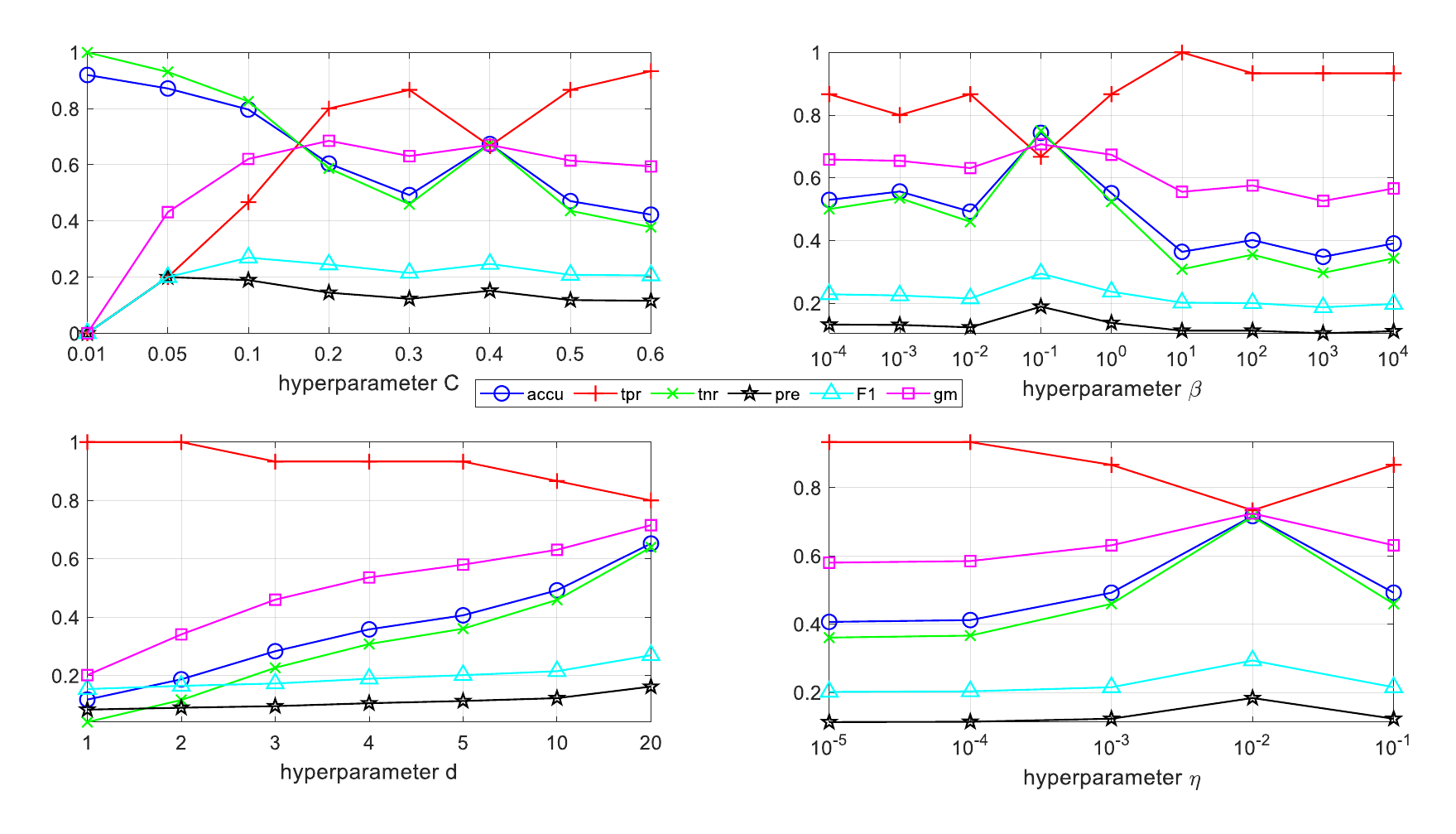}
	\caption{Hyperparameter sensitivity analysis for linear MS-SVDD $\omega_6ds2$ on SPECTF heart dataset}
	\label{w6d2}
\end{figure*}
 \begin{figure*}[ht]
	\centering
	\includegraphics[width=\textwidth]{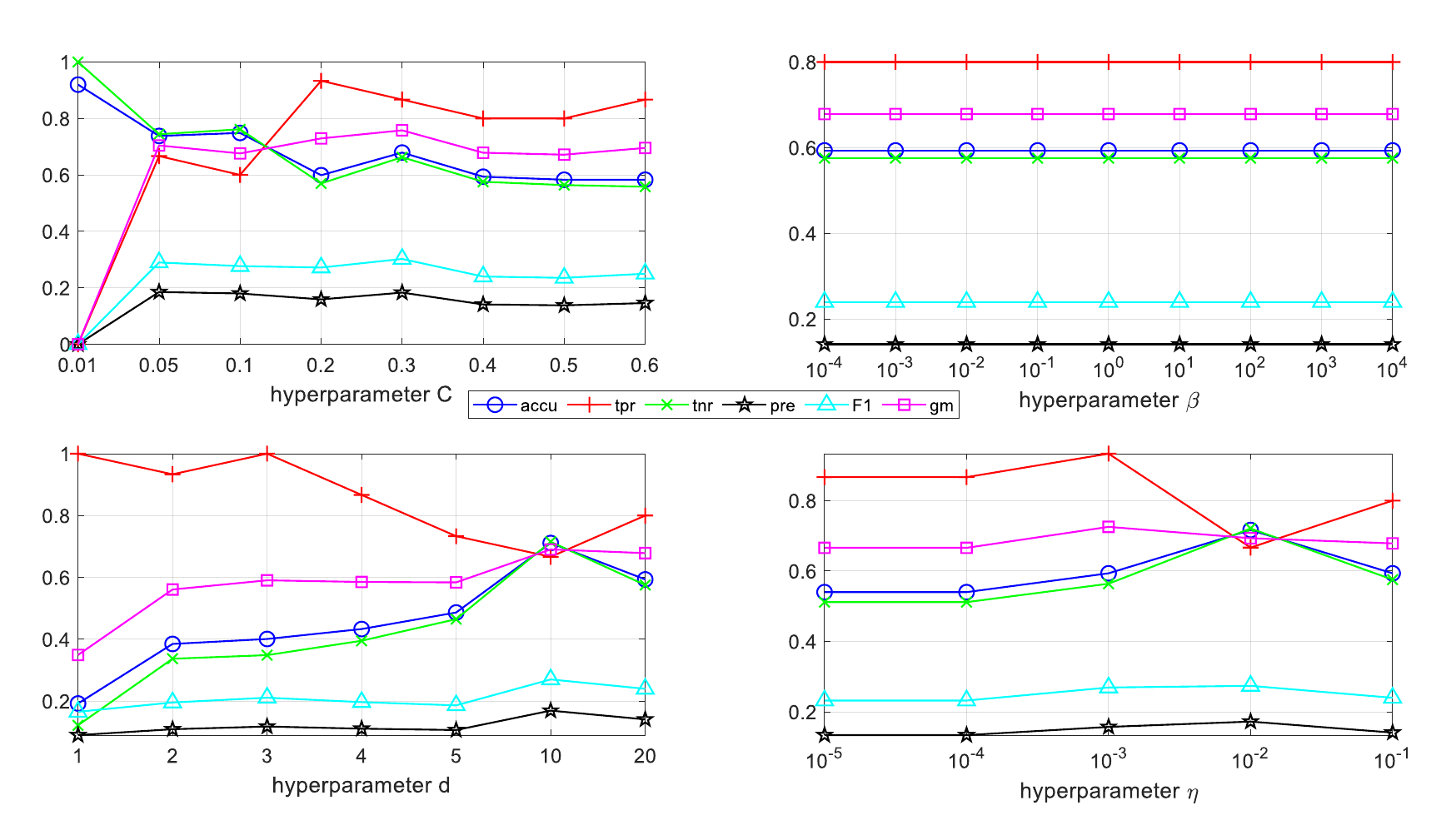}
	\caption{Hyperparameter sensitivity analysis for linear MS-SVDD $\omega_0ds3$ on SPECTF heart dataset}
	\label{w0d3}
\end{figure*}
 \begin{figure*}[ht]
	\centering
	\includegraphics[width=\textwidth]{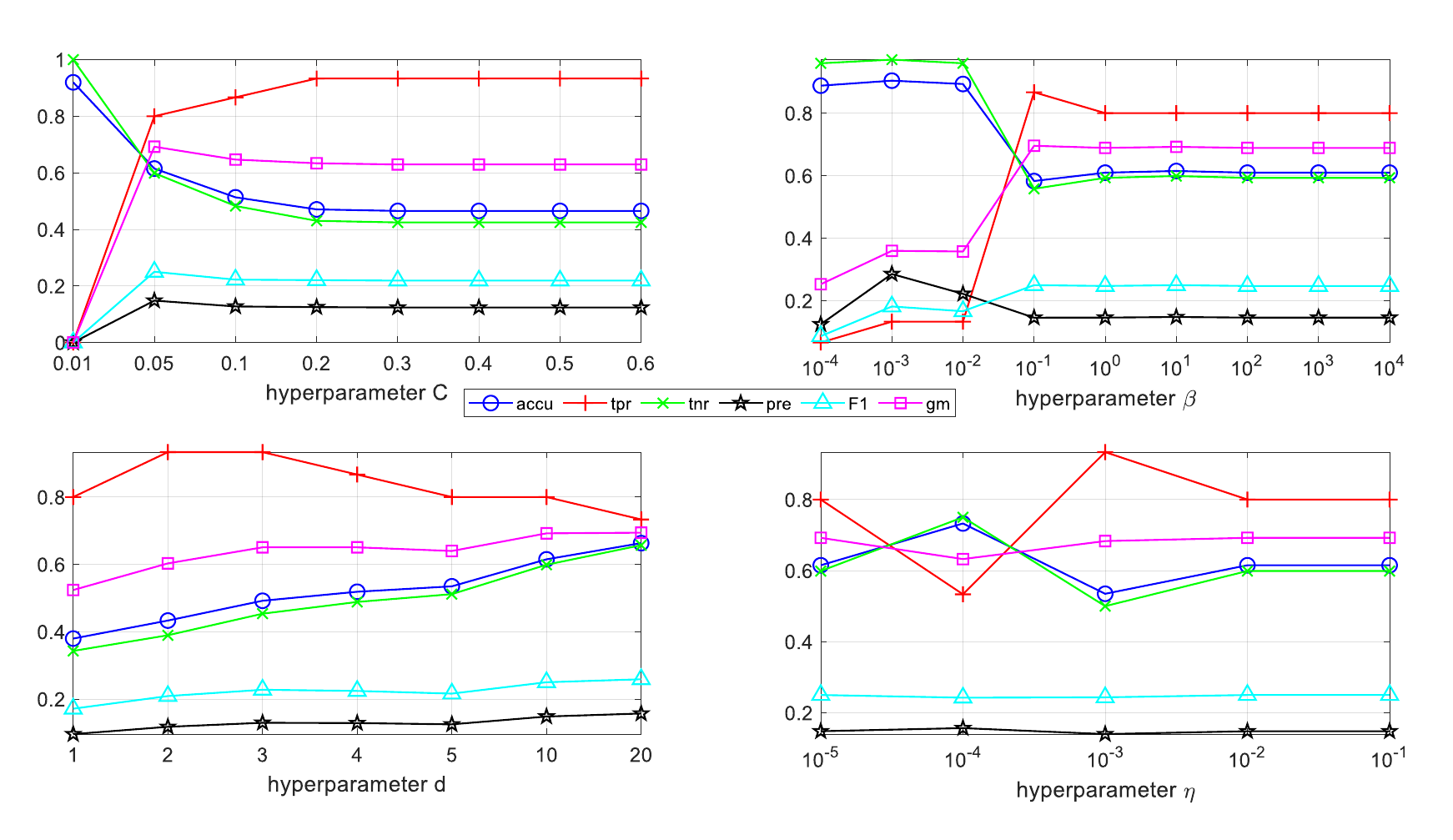}
	\caption{Hyperparameter sensitivity analysis for linear MS-SVDD $\omega_1ds3$ on SPECTF heart dataset}
	\label{w1d3}
\end{figure*}
 \begin{figure*}[ht]
	\centering
	\includegraphics[width=\textwidth]{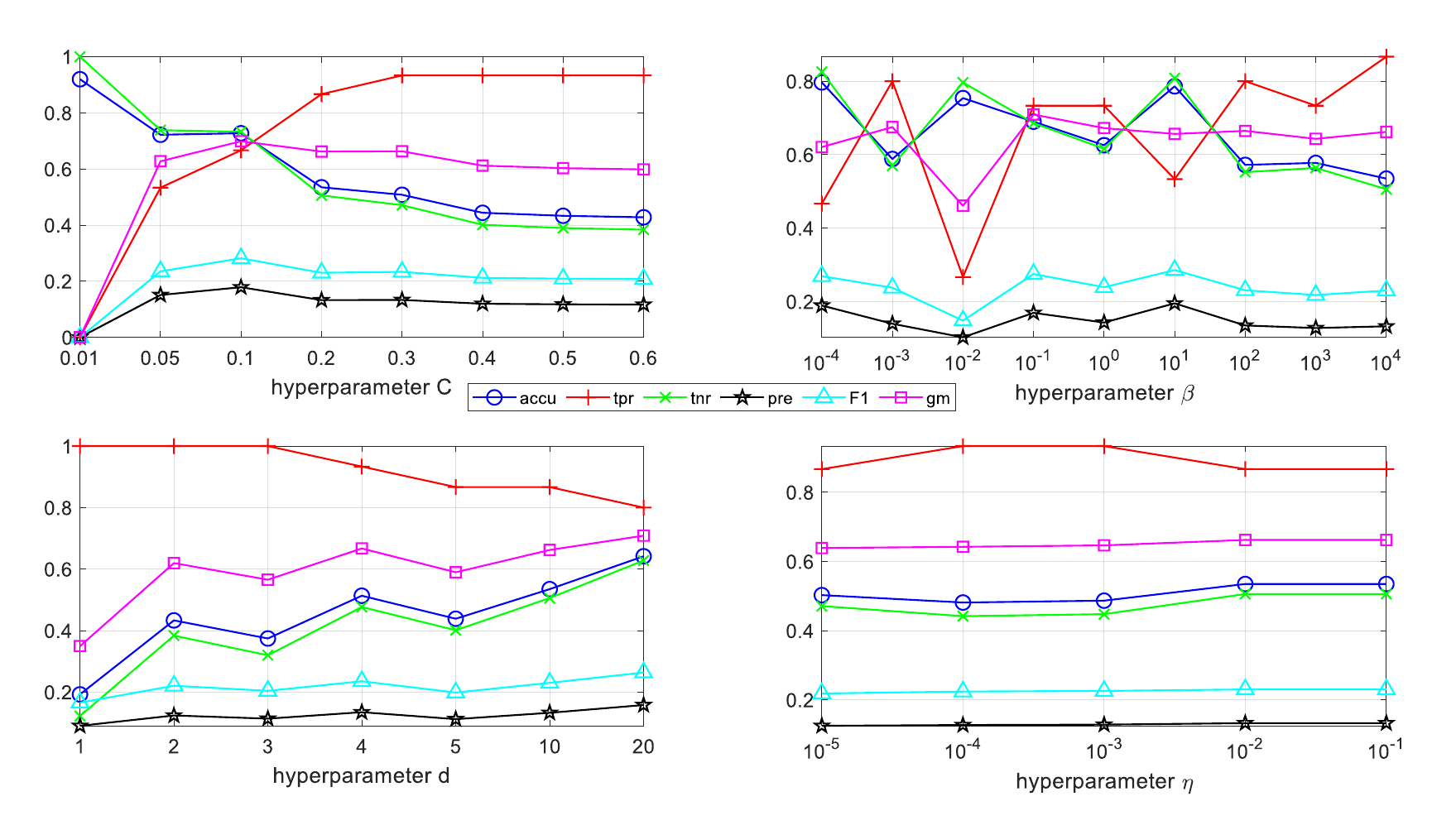}
	\caption{Hyperparameter sensitivity analysis for linear MS-SVDD $\omega_2ds3$ on SPECTF heart dataset}
	\label{w2d3}
\end{figure*}
 \begin{figure*}[ht]
	\centering
	\includegraphics[width=\textwidth]{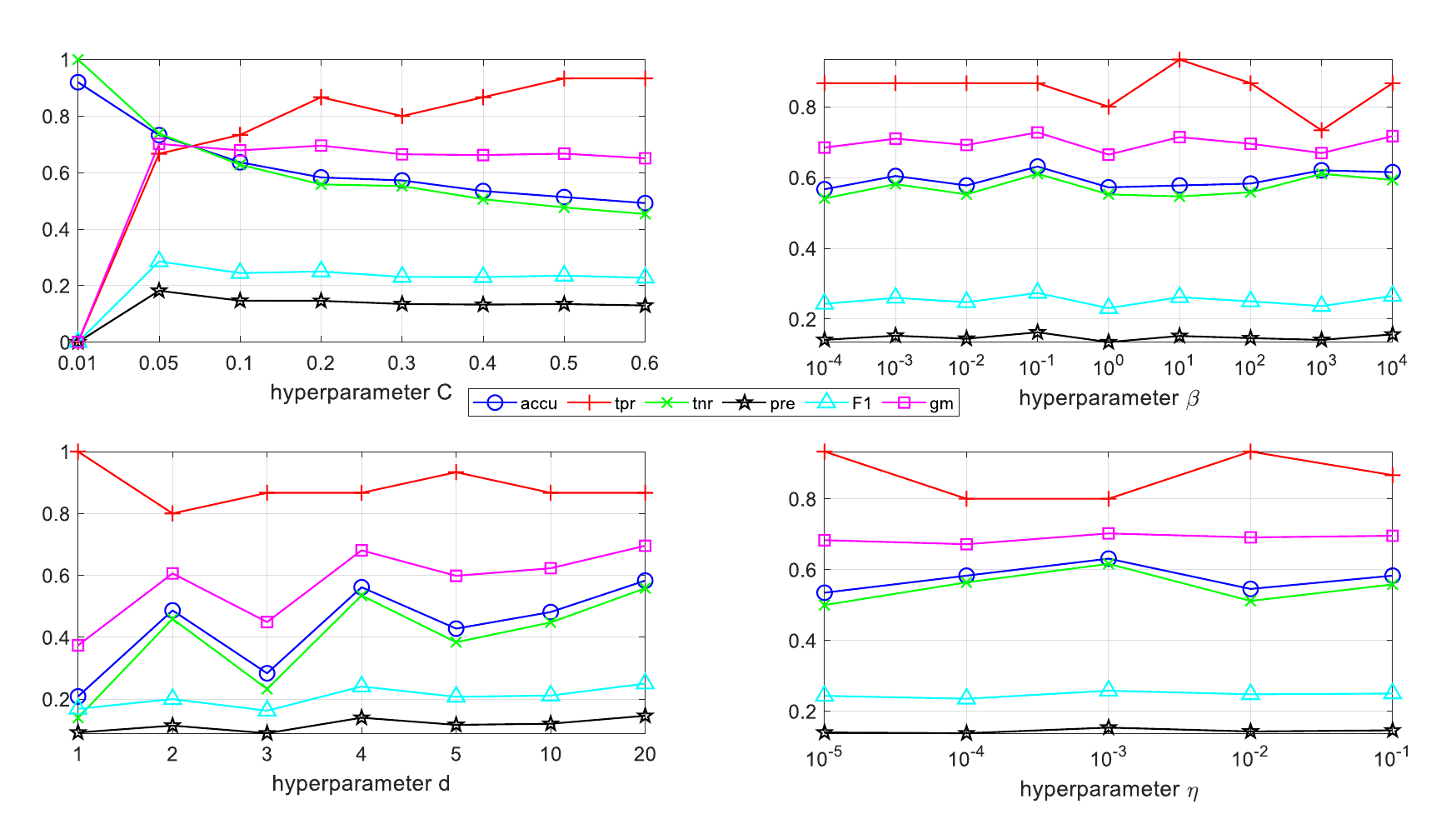}
	\caption{Hyperparameter sensitivity analysis for linear MS-SVDD $\omega_3ds3$ on SPECTF heart dataset}
	\label{w3d3}
\end{figure*}
 \begin{figure*}[ht]
	\centering
	\includegraphics[width=\textwidth]{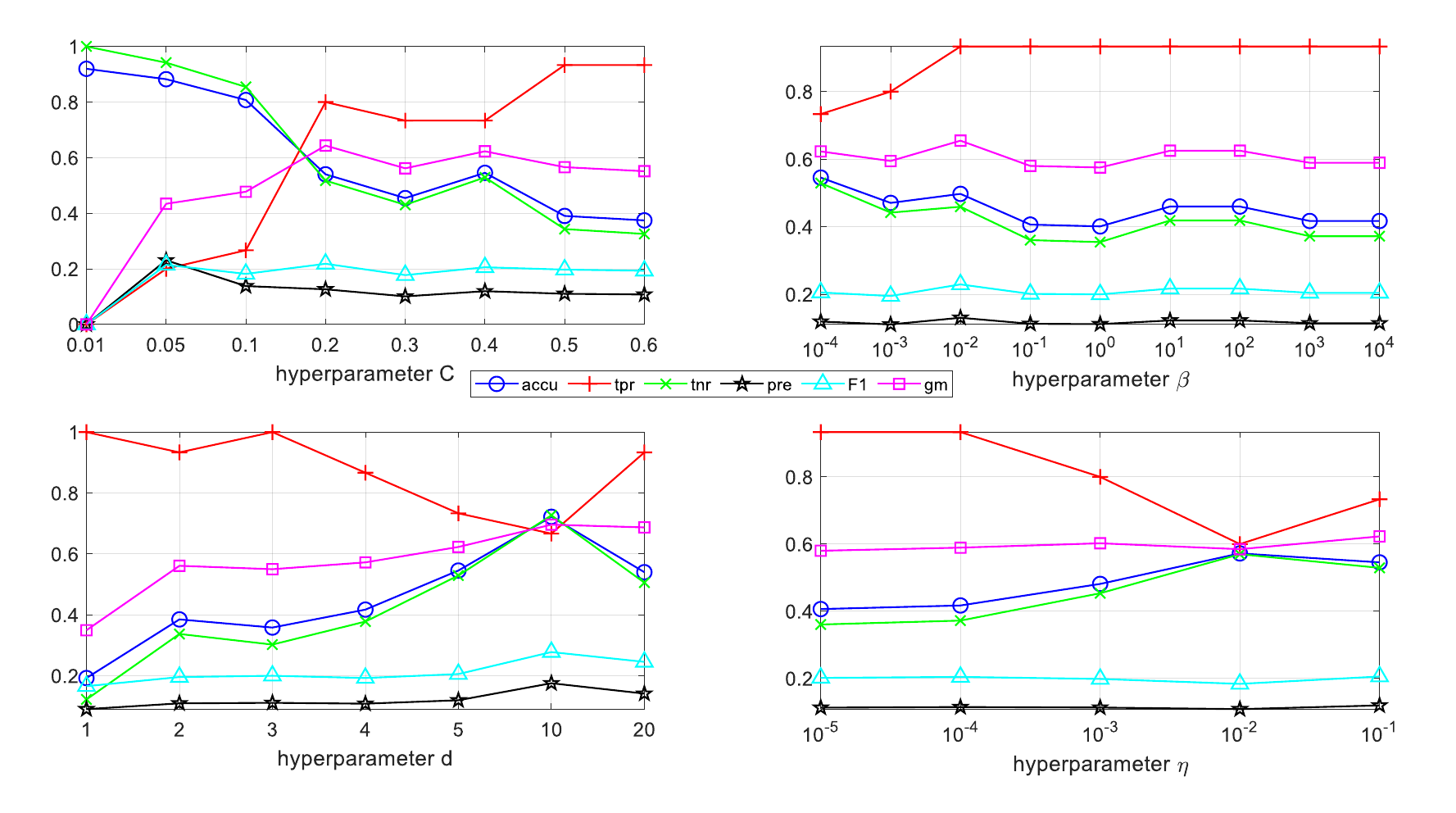}
	\caption{Hyperparameter sensitivity analysis for linear MS-SVDD $\omega_4ds3$ on SPECTF heart dataset}
	\label{w4d3}
\end{figure*}
 \begin{figure*}[ht]
	\centering
	\includegraphics[width=\textwidth]{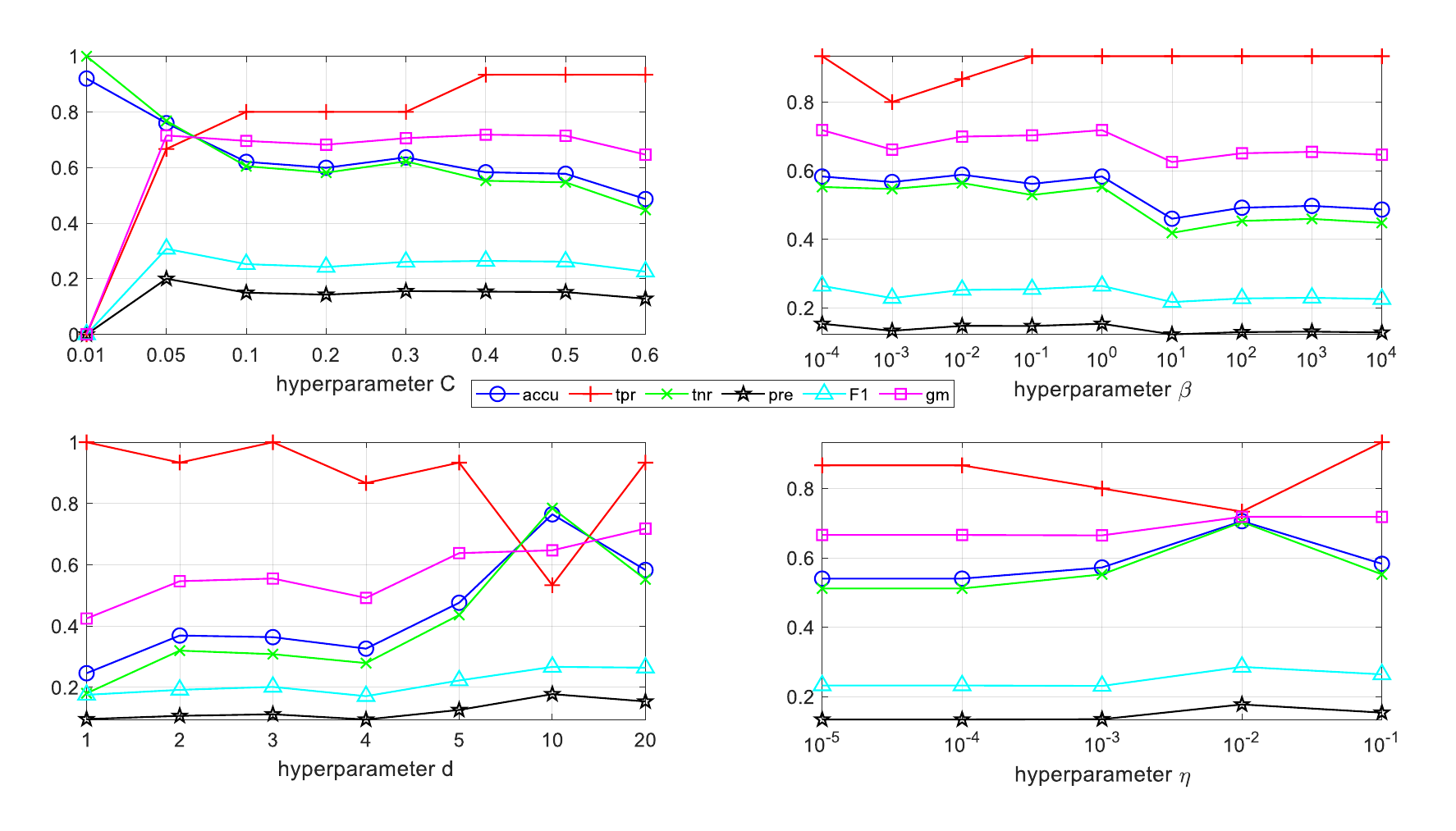}
	\caption{Hyperparameter sensitivity analysis for linear MS-SVDD $\omega_5ds3$ on SPECTF heart dataset}
	\label{w5d3}
\end{figure*}
 \begin{figure*}[ht]
	\centering
	\includegraphics[width=\textwidth]{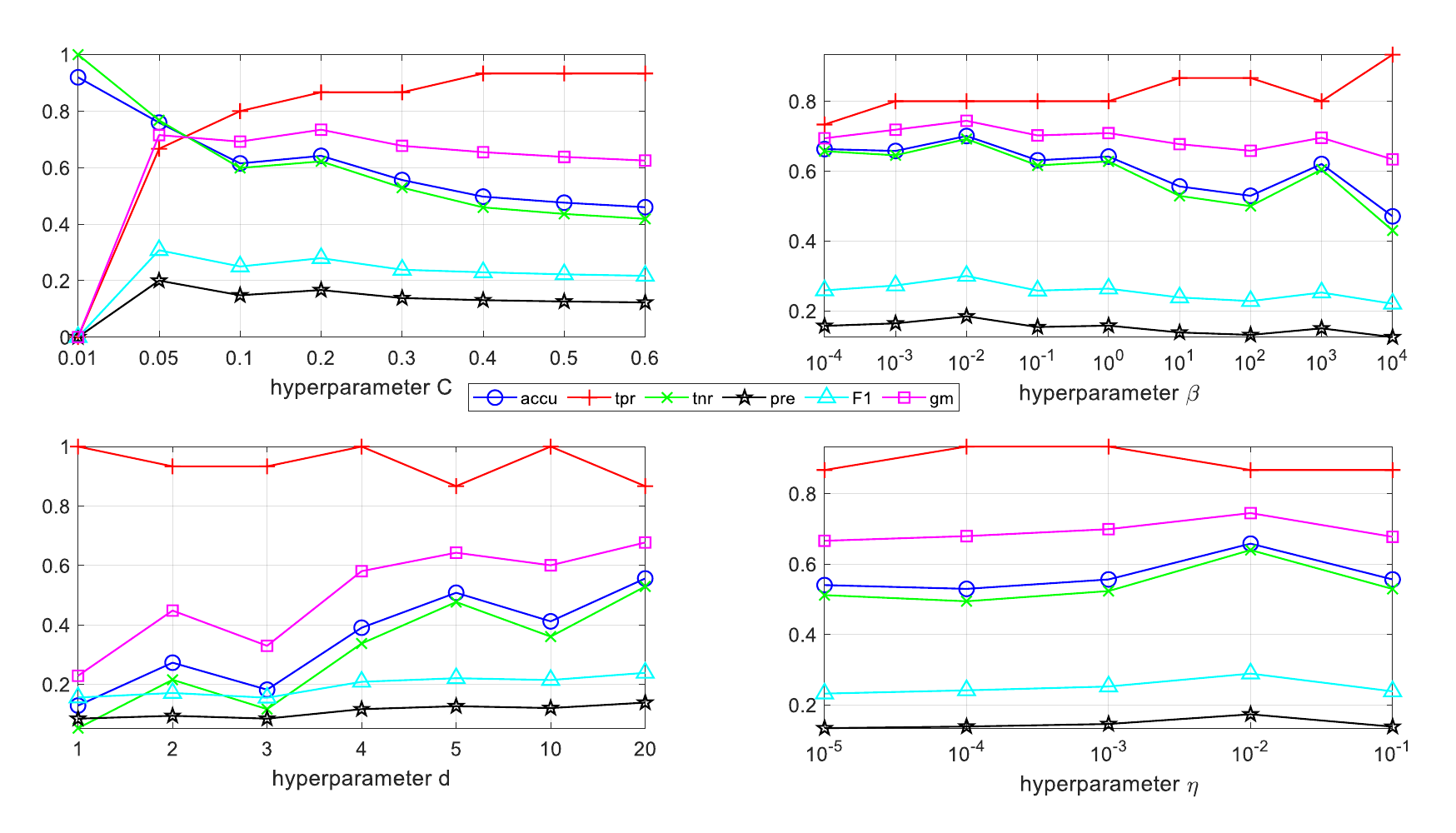}
	\caption{Hyperparameter sensitivity analysis for linear MS-SVDD $\omega_6ds3$ on SPECTF heart dataset}
	\label{w6d3}
\end{figure*}
 \begin{figure*}[ht]
	\centering
	\includegraphics[width=\textwidth]{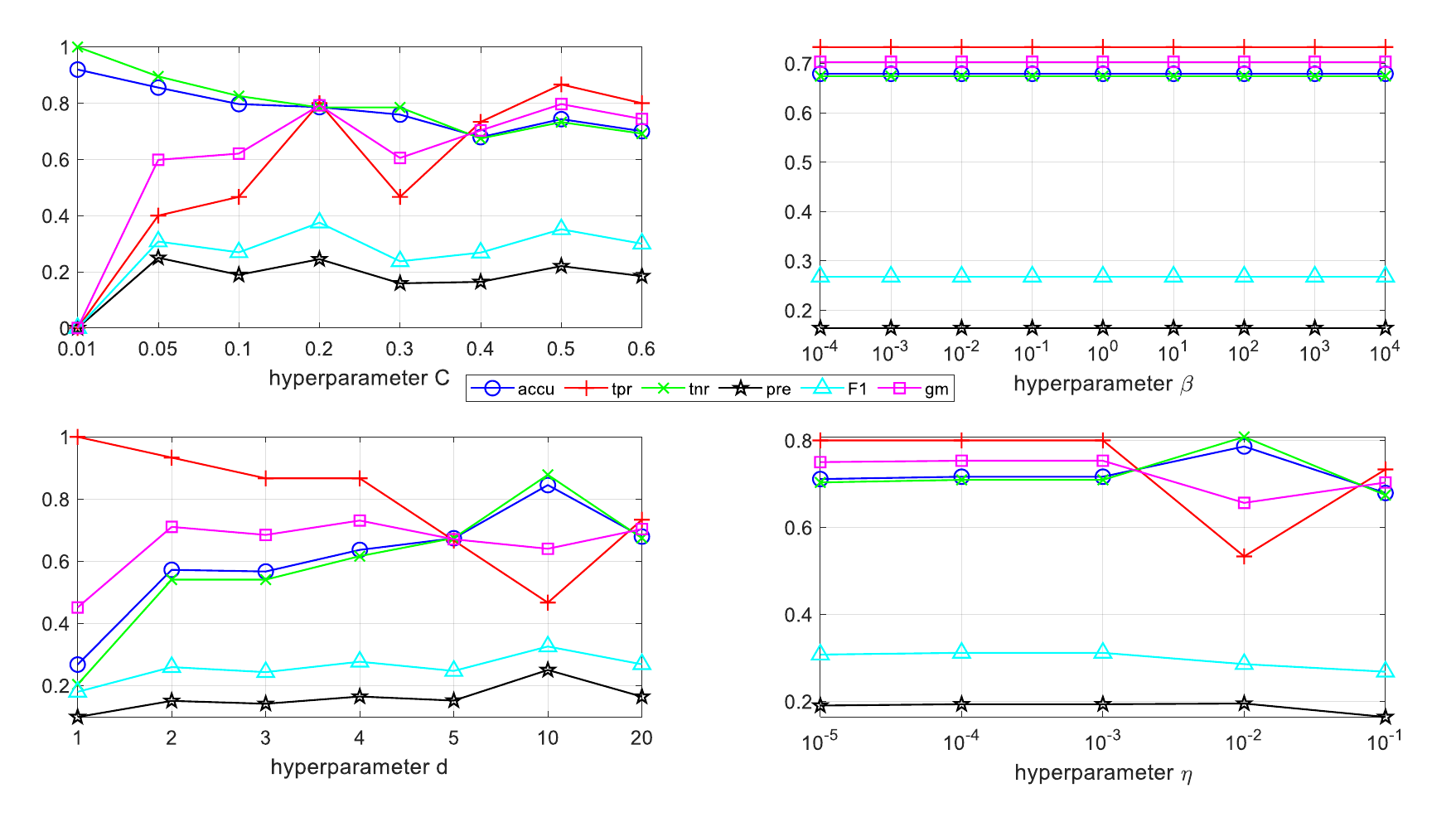}
	\caption{Hyperparameter sensitivity analysis for linear MS-SVDD $\omega_0ds4$ on SPECTF heart dataset}
	\label{w0d4}
\end{figure*}
 \begin{figure*}[ht]
	\centering
	\includegraphics[width=\textwidth]{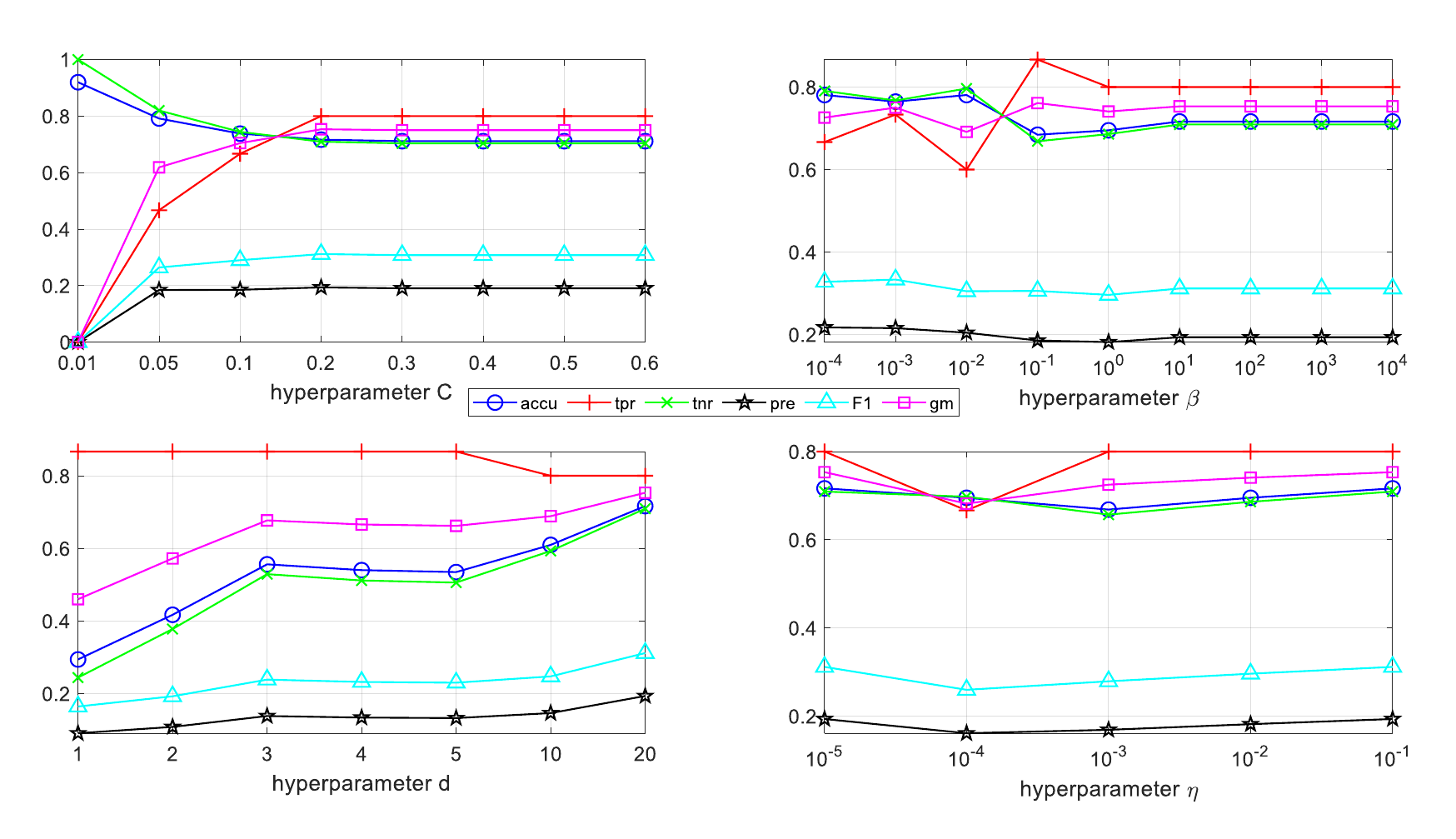}
	\caption{Hyperparameter sensitivity analysis for linear MS-SVDD $\omega_1ds4$ on SPECTF heart dataset}
	\label{w1d4}
\end{figure*}
 \begin{figure*}[ht]
	\centering
	\includegraphics[width=\textwidth]{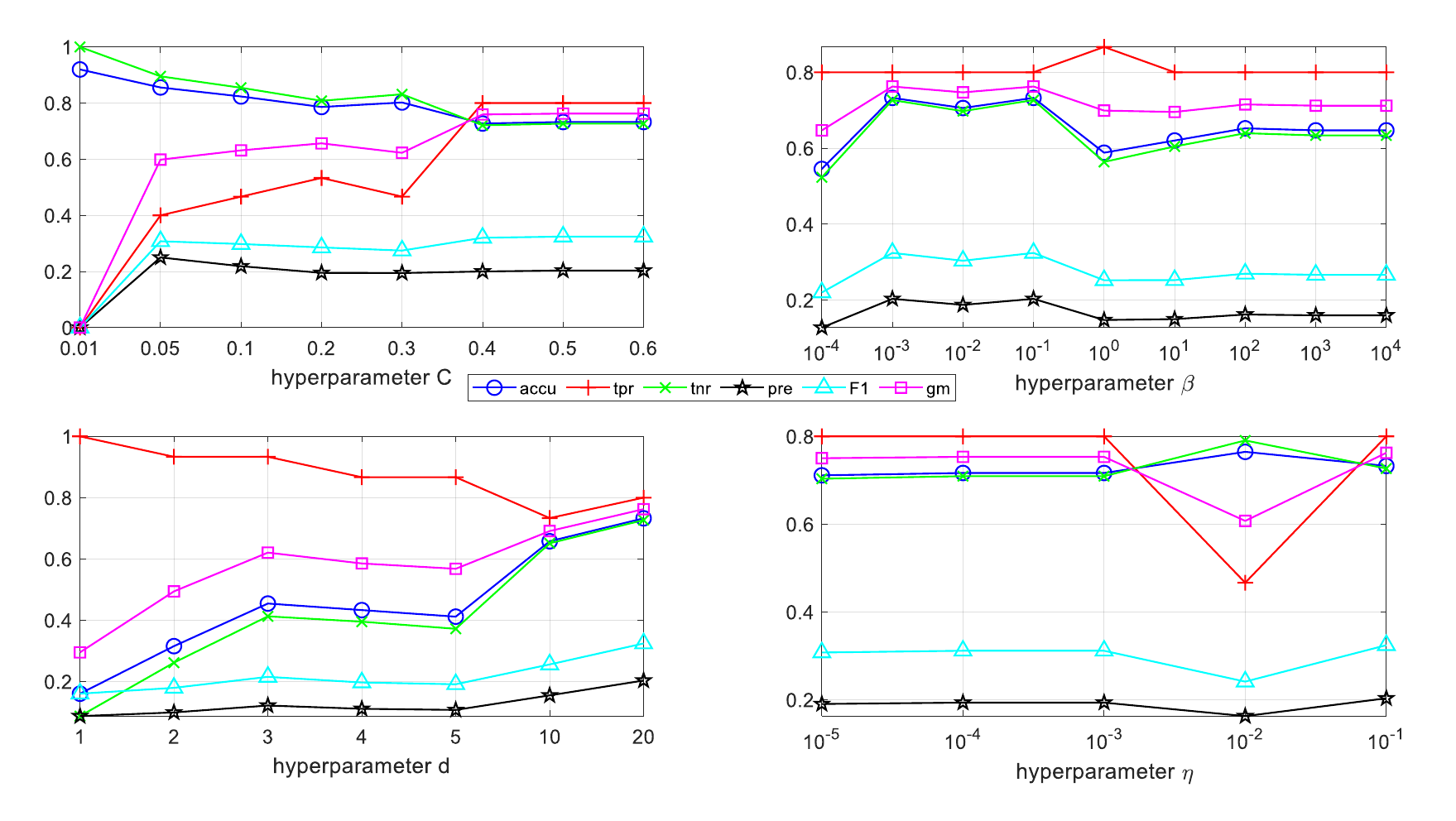}
	\caption{Hyperparameter sensitivity analysis for linear MS-SVDD $\omega_2ds4$ on SPECTF heart dataset}
	\label{w2d4}
\end{figure*}
 \begin{figure*}[ht]
	\centering
	\includegraphics[width=\textwidth]{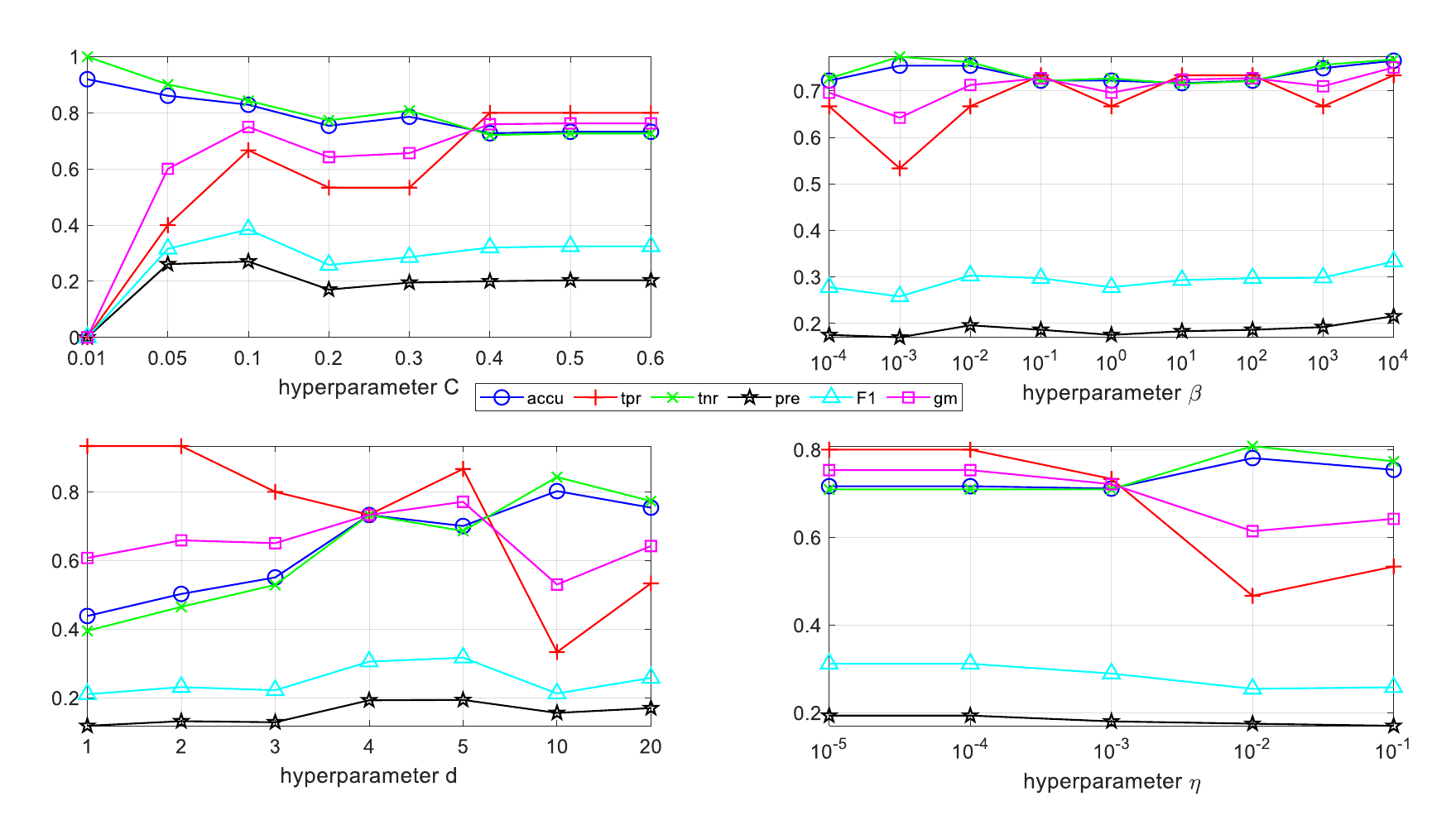}
	\caption{Hyperparameter sensitivity analysis for linear MS-SVDD $\omega_3ds4$ on SPECTF heart dataset}
	\label{w3d4}
\end{figure*}
 \begin{figure*}[ht]
	\centering
	\includegraphics[width=\textwidth]{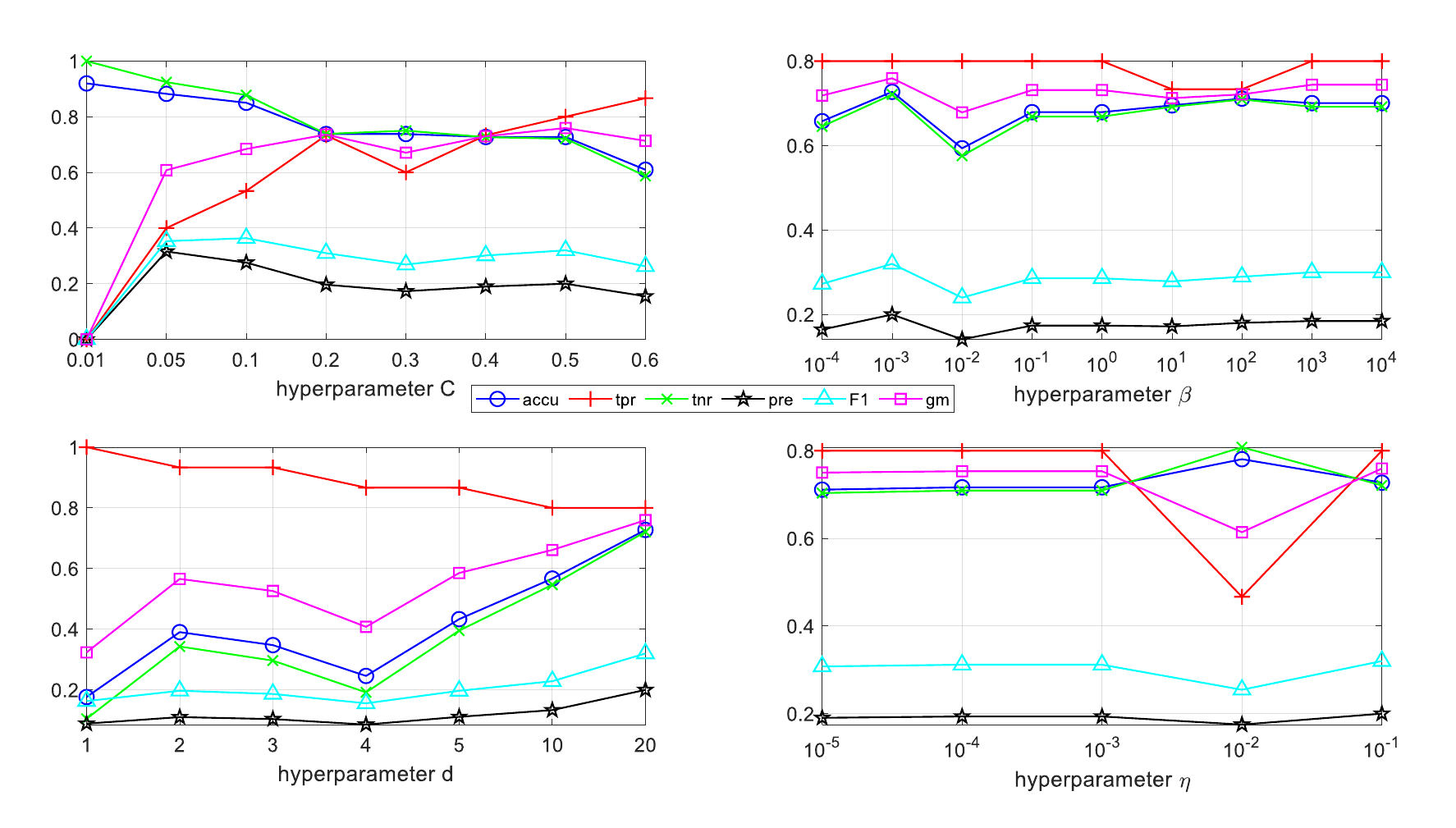}
	\caption{Hyperparameter sensitivity analysis for linear MS-SVDD $\omega_4ds4$ on SPECTF heart dataset}
	\label{w4d4}
\end{figure*}
 \begin{figure*}[ht]
	\centering
	\includegraphics[width=\textwidth]{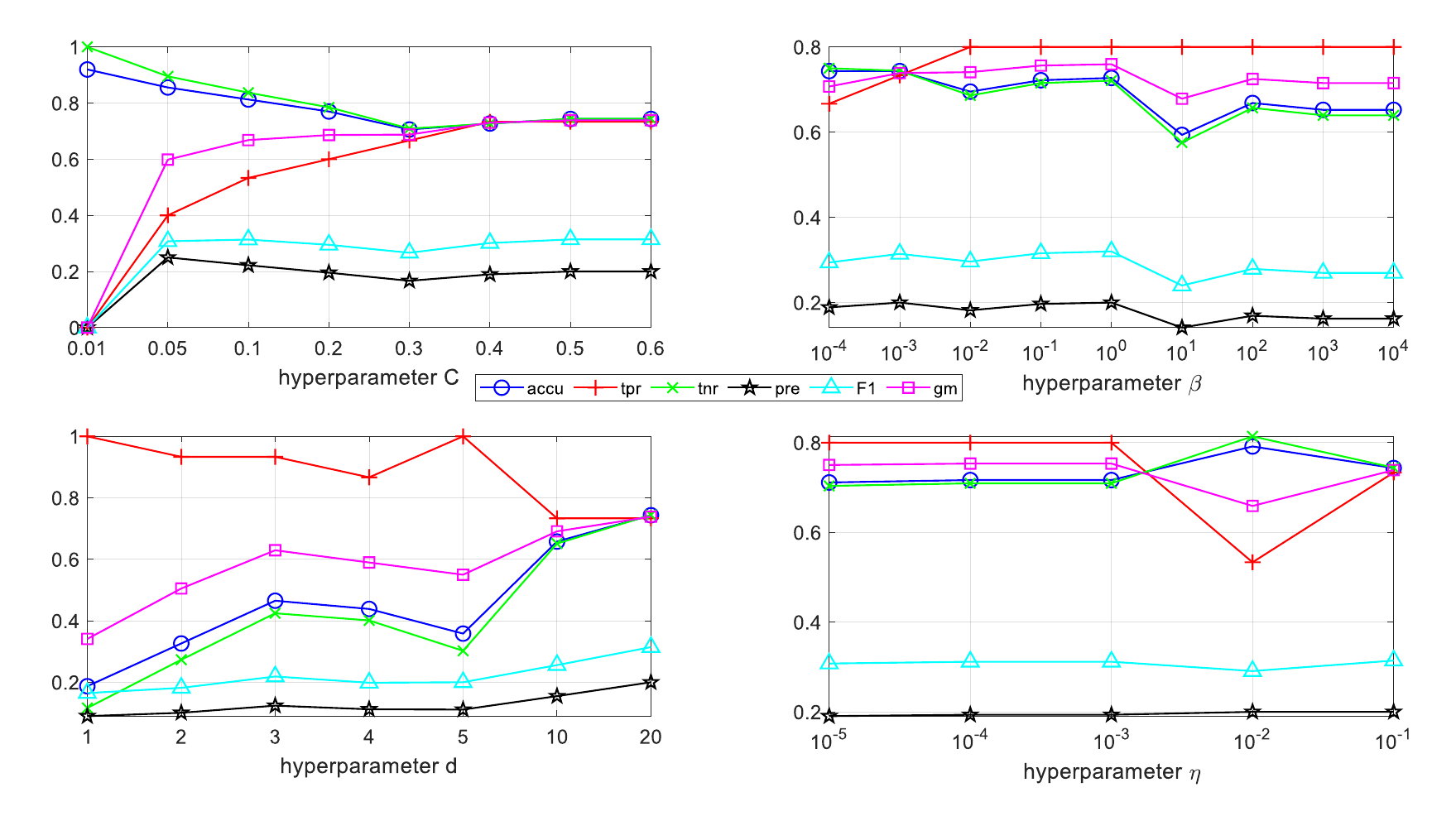}
	\caption{Hyperparameter sensitivity analysis for linear MS-SVDD $\omega_5ds4$ on SPECTF heart dataset}
	\label{w5d4}
\end{figure*}
 \begin{figure*}[ht]
	\centering
	\includegraphics[width=\textwidth]{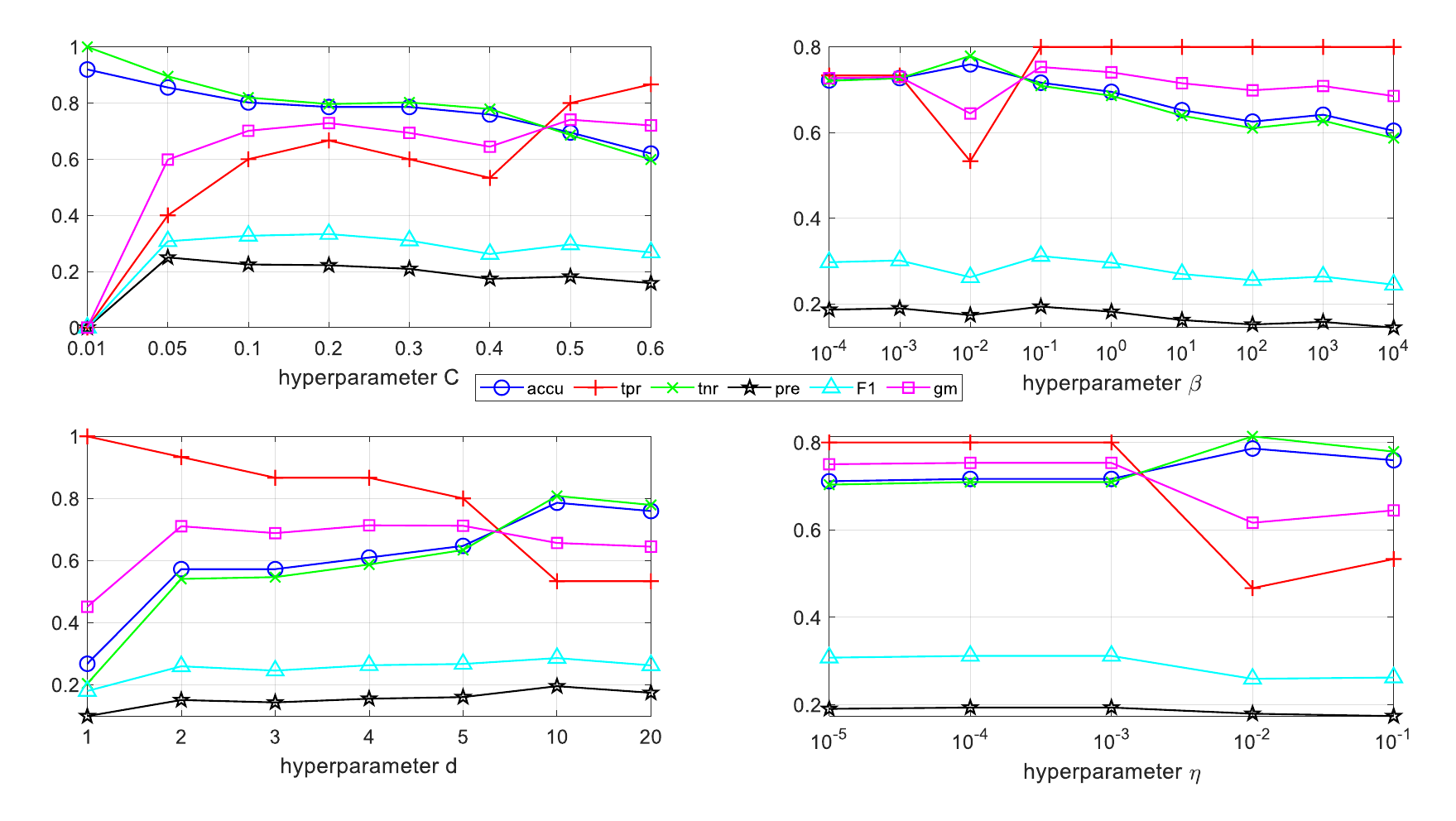}
	\caption{Hyperparameter sensitivity analysis for linear MS-SVDD $\omega_6ds4$ on SPECTF heart dataset}
	\label{w6d4}
\end{figure*}
\end{document}